\documentclass[english]{article}

\pdfoutput=1

\usepackage{times}
\usepackage{epsfig}
\usepackage{graphicx}
\usepackage{amsmath}
\usepackage{amssymb}
\usepackage{multirow}

\makeatletter

\newtheorem{thm}{\protect\theoremname}

\usepackage{algorithm}
\usepackage{algorithmic}
\usepackage{epstopdf}

\providecommand{\theoremname}{Theorem}

\usepackage[pagebackref=true,breaklinks=true,letterpaper=true,colorlinks,bookmarks=false]{hyperref}

\begin{document}

%%%%%%%%% TITLE
\title{A Grassmannian Graph Approach to Affine Invariant Feature Matching}

\author{Mark Moyou, John Corring, Adrian Peter, Anand Rangarajan\\
{\tt\small mmoyou@my.fit.edu, corring@cise.ufl.edu, apeter@fit.edu}\\
{\tt\small anand@cise.ufl.edu}
%% omit the following lines up until the closing ``}''.
%% Additional authors and addresses can be added with ``\and'',
%% just like the second author.
%% To save space, use either the email address or home page, not both
%\and \\
%%Institution2\\
%%First line of institution2 address\\
}

\maketitle
%\thispagestyle{empty}

%%%%%%%%% ABSTRACT
\begin{abstract}
		 In this work, we present a novel and practical approach to address
	one of the longstanding problems in computer vision: 2D and 3D affine
	invariant feature matching. Our Grassmannian Graph (GrassGraph)
	framework employs a two stage procedure that is capable of robustly
	recovering correspondences between two unorganized, affinely related
	feature (point) sets. The first stage maps the
	feature sets to an affine invariant Grassmannian representation, where
	the features are mapped into the same subspace. It turns out that
	coordinate representations extracted from the Grassmannian differ
	by an arbitrary orthonormal matrix. In the second stage, by approximating
	the Laplace-Beltrami operator (LBO) on these coordinates, this extra
	orthonormal factor is nullified, providing true affine-invariant coordinates
	which we then utilize to recover correspondences via simple nearest
	neighbor relations. The resulting GrassGraph algorithm is empirically
	shown to work well in non-ideal scenarios with noise, outliers, and
	occlusions. Our validation benchmarks use an unprecedented \textbf{440,000}+
	experimental trials performed on 2D and 3D datasets, with a variety
	of parameter settings and competing methods. State-of-the-art performance
	in the majority of these extensive evaluations confirm the utility
	of our method.
\end{abstract}

%%%%%%%%% BODY TEXT
\section{Introduction}

Feature matching has been a vital component of computer vision
since Fischler and Elschlager\textquoteright s seminal work in 1973
\cite{Fischler73}.  Since that beginning, there have been torrents of work in
this area making it almost impossible to characterize or synthesize. Given
this voluminous previous work, it\textquoteright s understandable if one
adopts a perspective that the bar is too high for new ideas. Our goal in the
present work is to belie that opinion, clearly demonstrating a novel approach
which is simultaneously new (to the best of our knowledge) and easy to
implement. Graph representations abound in computer vision.  The very first
matching work featured a relational graph representation \cite{Fischler73},
invariant under rigid shape transformations. Within this subfield, there again
exist innumerable works with extensions, new formulations, algorithms and the
like. Graph representations lead to graph matching\textemdash which despite
being NP-hard \cite{Zhou12}\textemdash has attracted a huge amount of
attention over the decades.

\begin{figure*}[t]
\centering{}%
\begin{tabular}{ccc}
  \includegraphics[width=2.9cm,height=2.5cm]{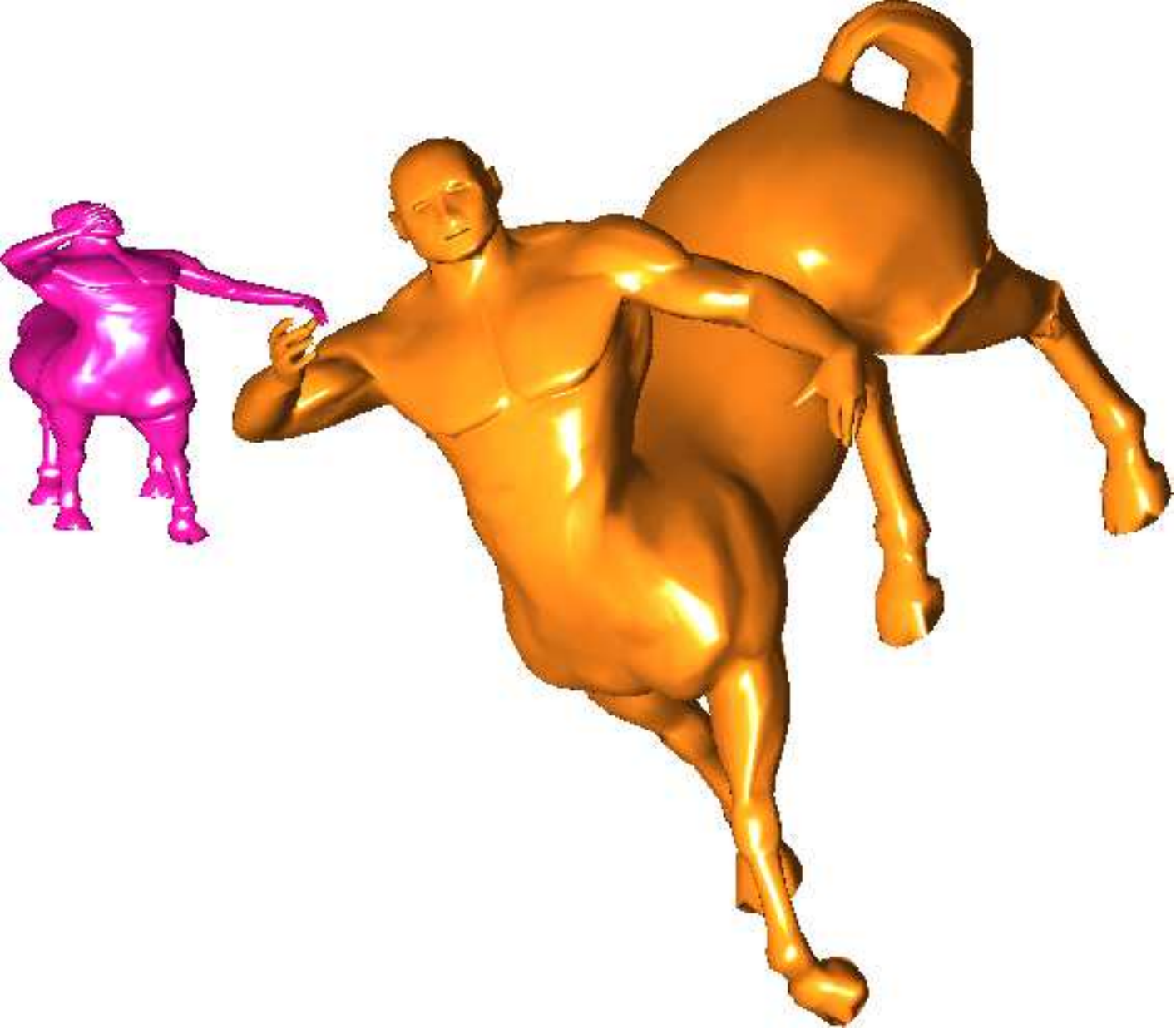}
  & \includegraphics[width=2.9cm,height=2.5cm]{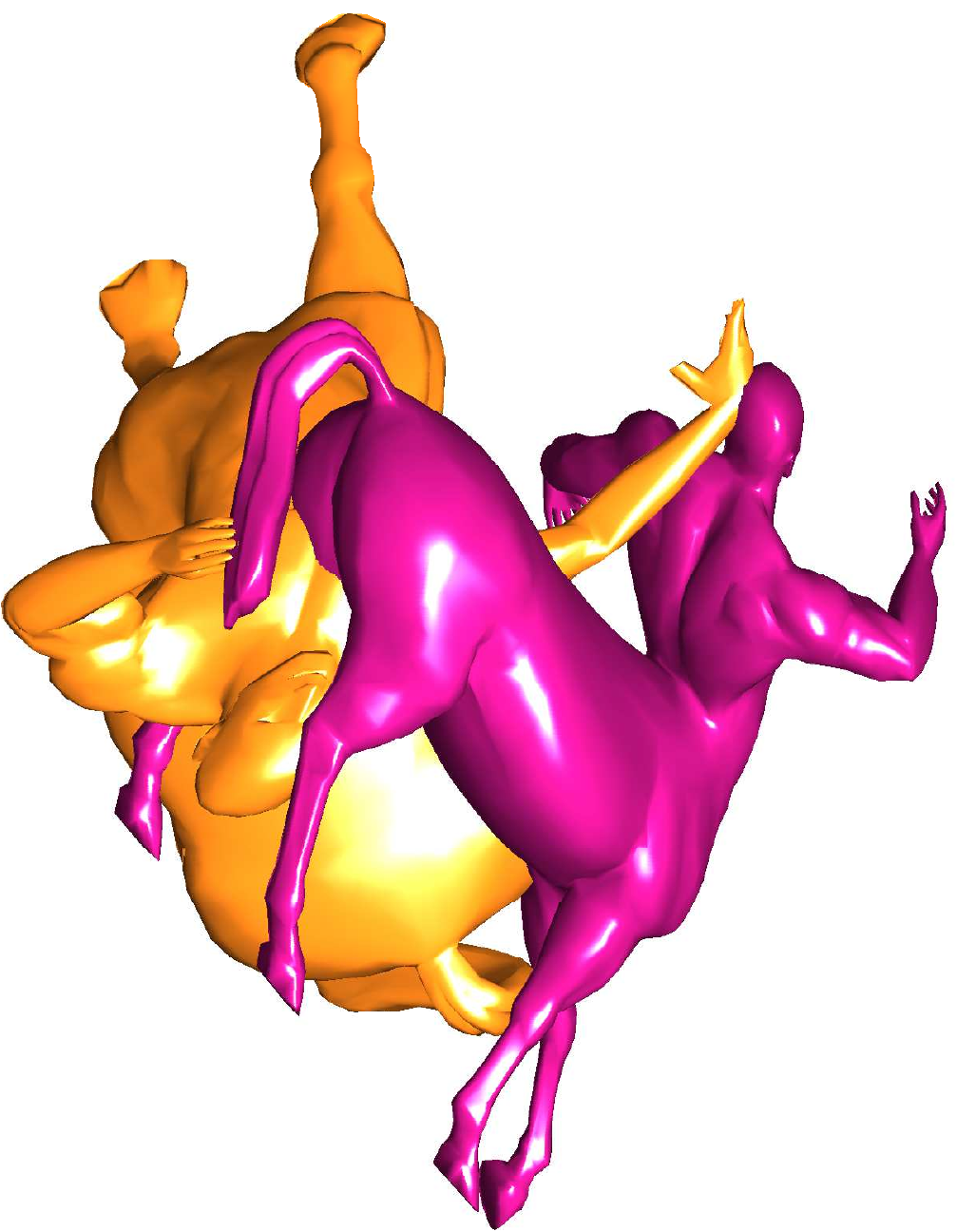}
  & \includegraphics[width=5.2cm,height=2.5cm]{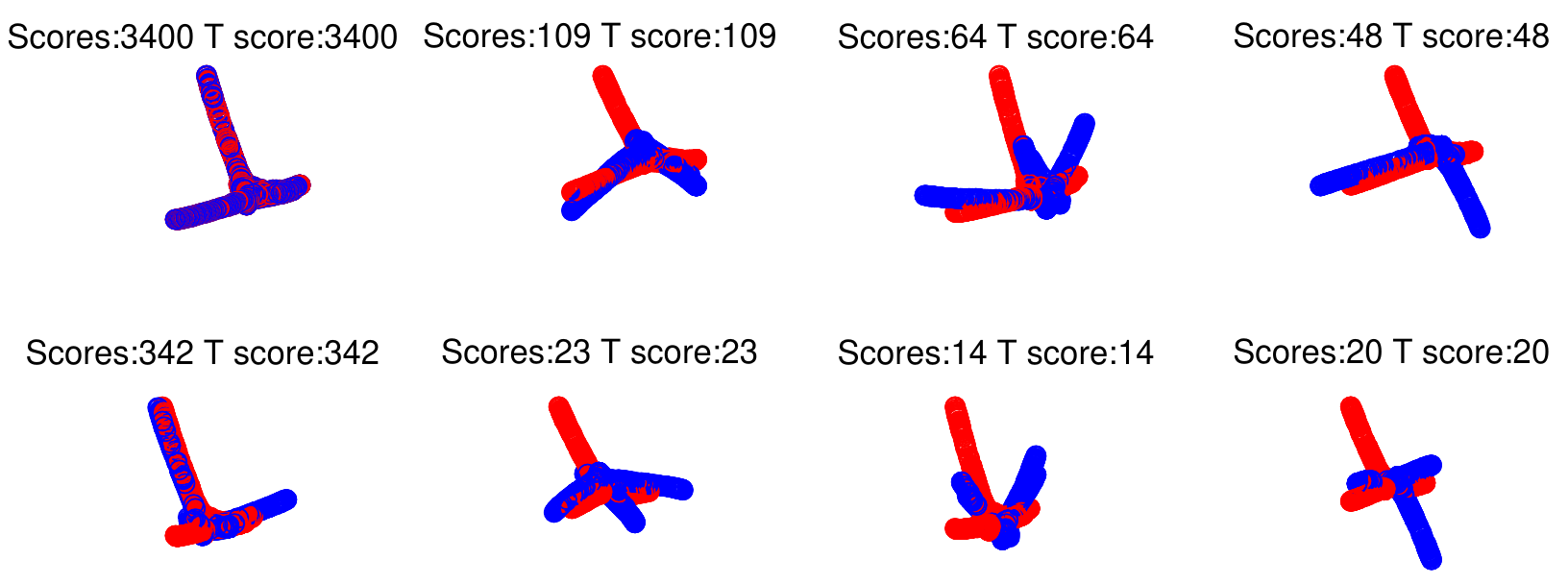}\vspace{-1pt}\tabularnewline   
{\scriptsize{}Original Affine Shapes} & {\scriptsize{}Grassmannian Representation } & {\scriptsize{}LBO Eigenvector Representation with Sign-flips}\tabularnewline
\end{tabular}\protect\caption{Visual overview of the GrassGraph algorithm. On the far left, we begin with the original shapes 
  which differ by an affine transform.  Moving to the right, the Grassmannian
  representation stage---constructed via a SVD on the coordinates---removes
  the affine transformation, leaving only a rotation between the shapes. Next,
  in our second stage, we approximate the Laplace-Beltrami operator (LBO) on
  these invariant coordinates and compute its eigenvectors to obtain a set of
  coordinates that are invariant to isometries (which includes rotations).
  Finally, we correct for the sign ambiguties that arise from the eigenvector
  computations. To choose the correct sign flips, a number of mutual
  nearest neighbor correspondences 
  are computed and the flip combination with the highest score is chosen.
  The proposed GrassGraph approach is able to recover dense correspondences
  with the same identical steps applicable to either 2D or 3D data. 
	\label{fig: GGAlgorithm}}
	\end{figure*}

	Graph representations have also been extended to include invariances
	under similarity transformations \cite{Siddiqi99} and lately have
	surfaced in non-rigid matching situations as well \cite{Zhou2013}.
	However, in the present work, our focus is on affine invariance. In
	particular, we introduce both a new \emph{Grassmannian Graph} representation
	and an attendant set of affine invariant feature coordinates. We believe
	this combination to be both novel and useful in feature matching and
	indexing applications. 

	Invariants in computer vision have seen better days. While they were
        the flavor \emph{du jour} in the early \textquoteright 90s, with work
        ranging from geometric hashing \cite{Lamdan90} to generalized Hough
        transforms \cite{Ball81}, recently this work has not seen much
        development.  Invariants were not robust to missing features and they
        could not easily be extended to non-rigid matching situations. We
        speculate, however, that an important additional reason why invariants
        did not see wide adoption was due to the absence of invariant feature
        coordinate systems. For example, there were relatively few attempts at
        creating similarity transformation invariant coordinates from a set of
        features leading to simpler correspondence finding algorithms (such as
        nearest neighbor). It is not surprising that it is exactly this aspect
        of invariance which has seen a resurgence in the past few years
        \cite{Reuter06}. 

	While the idea of constructing new similarity transformation-invariant
	coordinates dates back earlier, it is only in recent years that it
	has become commonplace to see new coordinates built out of discrete
	Laplace-Beltrami Operator (LBO) eigenvectors \cite{Jones08,rustamov2007laplace}.
	The basic idea is extremely straightforward: construct a weighted
	graph representation from a set of features and then use the principal
	component eigenvectors of the graph (interpreted as a linear operator)
	to complete the construction. These new coordinates can then be pressed
	into service in matching and indexing applications. However, note
	the fundamental limitation to similarity transformations: can this
	approach be extended to affine invariance while remaining somewhat
	robust in the face of noise and outliers?

	We answer in the affirmative. In this work, we begin with the original
	feature set and first construct the Grassmannian. The Grassmannian
	\cite{Boothby02} is a geometric object with the property that all
	linear combinations of the feature coordinates remain in the same
	Grassmannian subspace. Therefore a \emph{single} element of the Grassmannian
	can be considered to be an affine invariant equivalence class of feature
	sets. Further, Grassmannians are homeomorphic to orthogonal projection
	matrices with the latter available for the construction of new affine
	invariant coordinates. Consequently, the first stage in our construction
	of affine invariant coordinates is the Grassmannian representation
	(GR) utilizing orthogonal projection matrices computed from each feature
	set. Unfortunately, it turns out that two factorizations of projection
	matrices can differ by an unknown orthonormal matrix. To circumvent
	this problem, in a second stage, we construct a weighted graph which
	is invariant to this additional orthonormal factor and then use LBO
	principal components (as outlined above) to obtain new affine invariant
	coordinates. Given two sets of features, finding good correspondences
	is considerably easier in this representation since the affine invariant
	coordinates lead to efficient nearest neighbor correspondences. The
	twin strands of research married in our approach are therefore (i)
	affine invariant Grassmannian representations and (ii) LBO-based weighted
	graphs resulting in the \emph{GrassGraph algorithm} for affine invariant
	feature matching. We believe this combination to be novel and useful
	and a visual representation of the algorithm is shown in Figure \ref{fig: GGAlgorithm}. 

	More than 440,000 experiments conducted in the present work buttress
	our claims. We take the outlier problems faced by invariant representations
	very seriously since they derailed previous work (from the early \textquoteright 90s).
	We also conduct realistic experiments in the presence of noise, missing
	points and outliers\textemdash on a greater scale than we have seen
	in comparable related work. Comparisons are conducted against
	state-of-the-art feature matching algorithms. In this way, we hope
	to have made the case for the affine invariant representation and
	new shape coordinates. The ease of implementation and use should pave
	the way for the Grassmannian Graph representation to be widely used
	in feature matching and indexing applications.

	%\begin{figure*}[t]
%\centering{}%
%\begin{tabular}{cccc}
%\includegraphics[width=5cm,height=3cm]{FinalCVPRImages/Other2/GorillaAffineCorres2}\hspace{-13pt} & \includegraphics[width=4cm,height=3cm]{FinalCVPRImages/Other2/seahorseAffine}\hspace{-13pt} & \includegraphics[width=2.5cm,height=3cm]{FinalCVPRImages/Other2/Lady3CorrAffine} & \includegraphics[width=5cm,height=3cm]{FinalCVPRImages/Other2/articulatedCorr1_Best}

\begin{figure*}
\centering{}%
\begin{tabular}{cc}
\includegraphics[width=5cm,height=3cm]{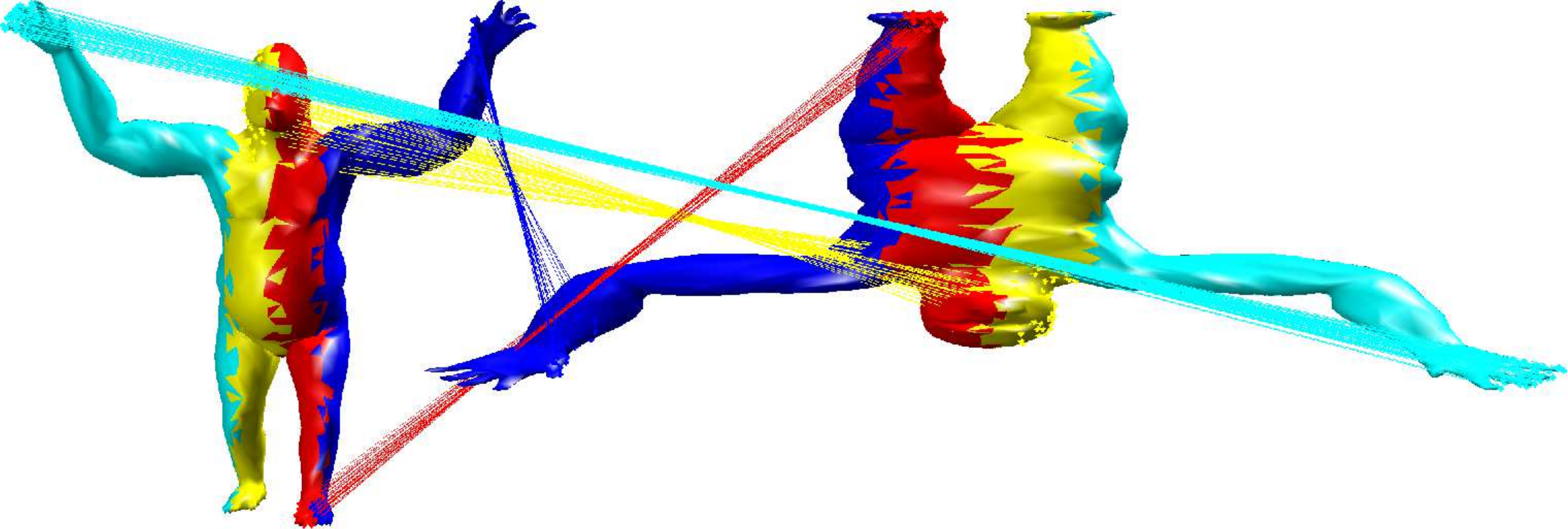} & 
\includegraphics[width=4cm,height=3cm]{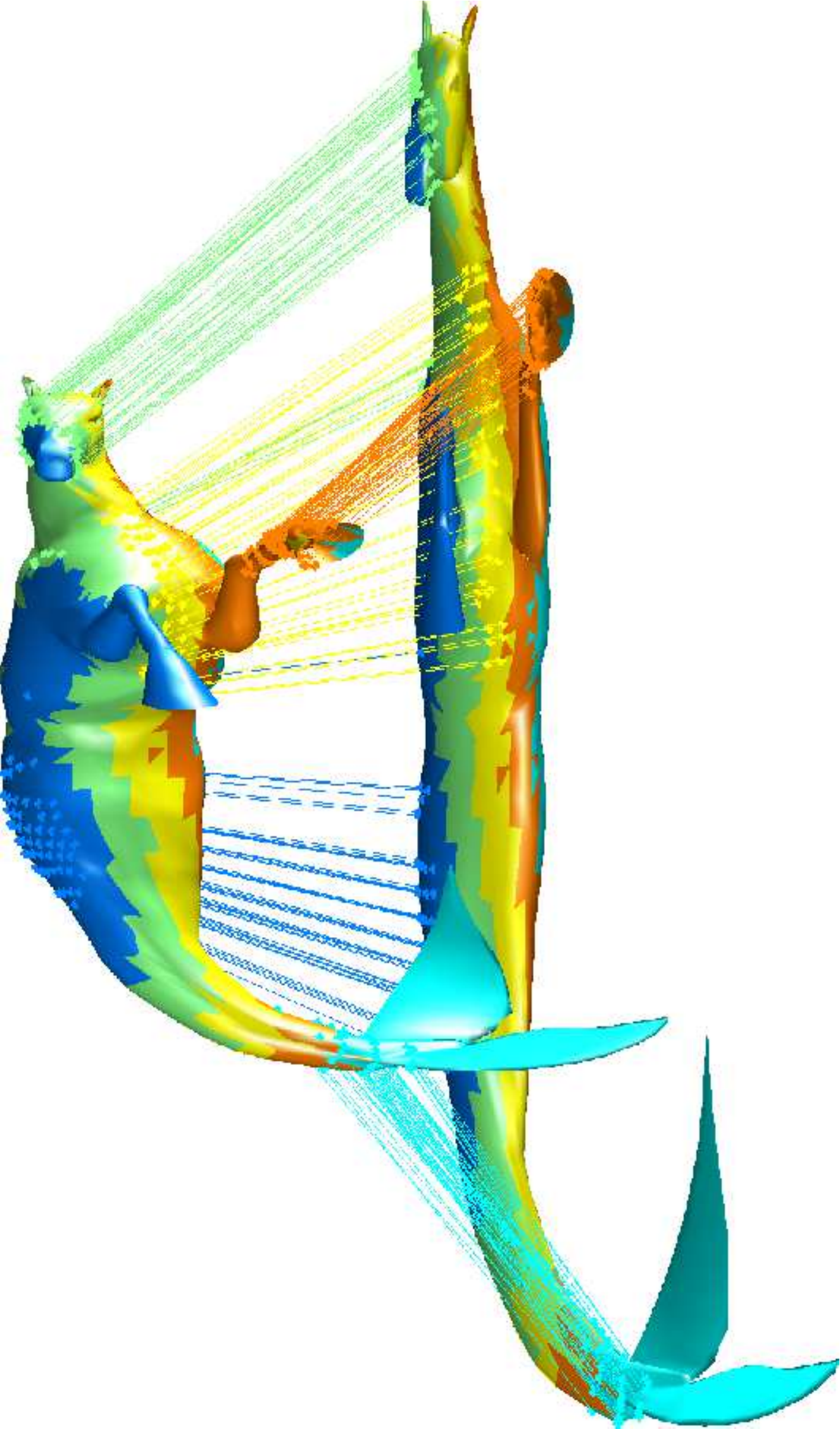}\vspace{0.5cm}
\tabularnewline 
\includegraphics[width=2.5cm,height=3cm]{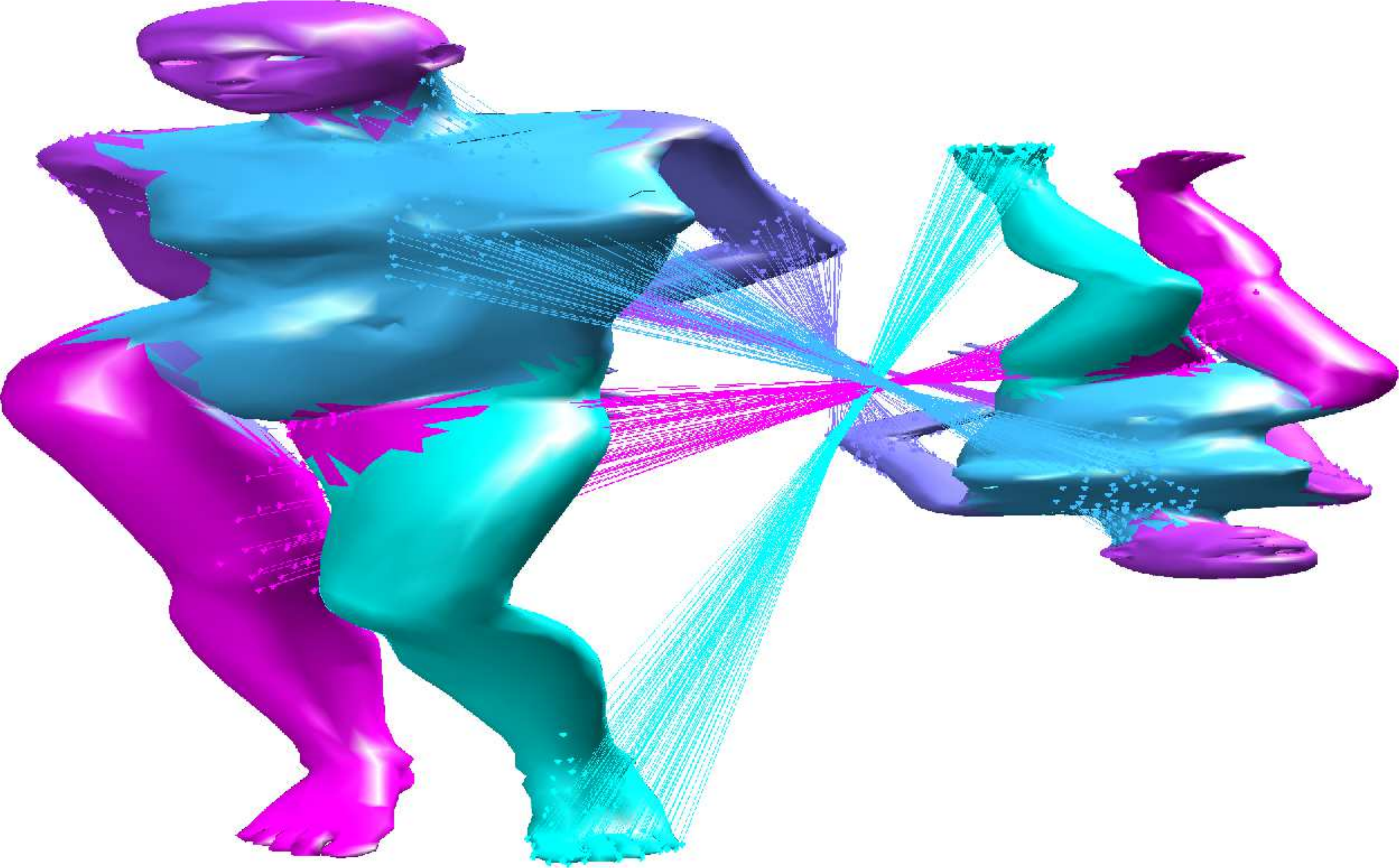} & 
\includegraphics[width=5cm,height=3cm]{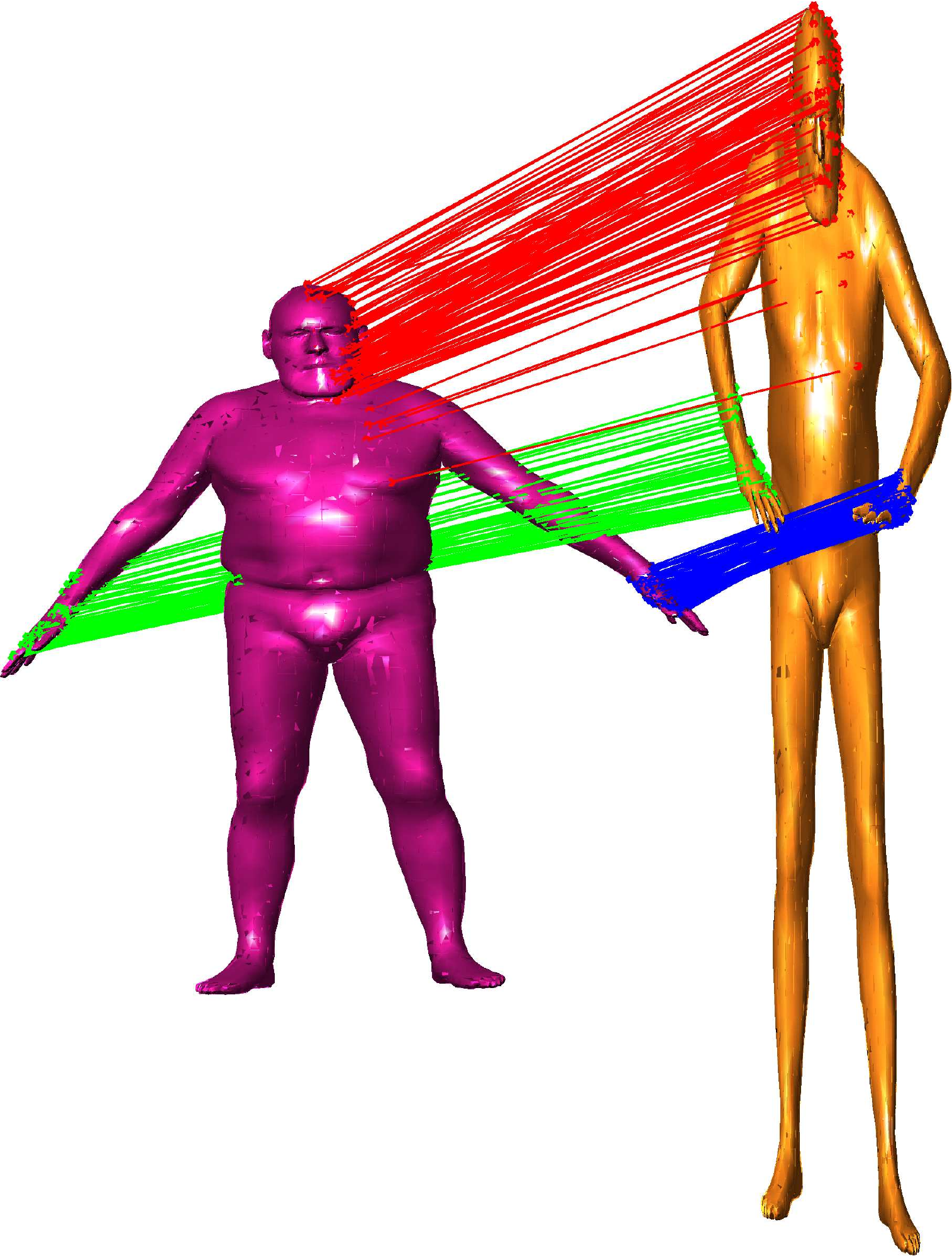}

\tabularnewline
	\end{tabular}\protect\caption{Correspondence recovery on 3D affine shapes and one articulated trial using the proposed GrassGraph algorithm.
	Note, that we recover dense correspondences but only a subset of the
	matching points are shown for visualization purposes. \label{fig: Dense Correspondences in 2D and 3D}}
	\end{figure*}

\section{Related Work}

As mentioned previously the corpus related to affine invariance and graph
matching is quite vast, here we highlight the most relevant and early
pioneering works that have paved the way for the current method.  Umeyama
\cite{Umey88} pioneered spectral approaches for the weighted graph matching
problem, with several other spectral methods proposed in
\cite{Caelli04,Carcassoni03,Luo00,Mateus08,Shapiro92}. After Umeyama, Scott
and Longuet-Higgins \cite{Scott91} designed an inter-image proximity matrix
between landmarks using Gaussian functions, then solve for correspondences via
the singular value decomposition (SVD) to match sets of rigid points. This
work was extended by Shapiro \cite{Shapiro92} to find correspondences on the
eigenvectors themselves and overcome the invariance to rotation limitation of
the work in \cite{Scott91}.  Carcassoni \cite{Carcassoni03b,Carcassoni03}
extended the Gaussian kernel to other robust weighting functions and proposed
a probabilistic framework for point-to-point matching. In \cite{Jain06}, the
Laplacian embedding is used to embed 3-D meshes using a global geodesic
distance where these embeddings are matched using the ICP algorithm
\cite{Zhang94}.  Mateus \cite{Mateus07} used a subset of Laplacian
eigenfunctions to perform dense registration between articulated point
sets. Once in the eigenspace, the registration was solved using unsupervised
clustering and the EM algorithm. All of these previous methods used an
eigenvector decomposition to solve the point matching problem under different
transformations: translation, rotation, scale and shear, but none are truly
affine invariant nor do they produce affine invariant coordinates. In addition,
establishing correspondence in the related feature spaces require more
complicated methods versus our simple nearest neighbor matching to recover
correspondences (even in the presence of noise and outliers).

Another way to address the affine invariance problem is through multi-step
approaches which provide invariance to particular transformations
at each step. Ha and Moura \cite{Ha05} recovered an intrinsic shape
which is invariant to affine-permutation distortions, and use the
steps of centering, reshaping, reorientation and sorting. Dalal et
al. \cite{Dalal07} constructed a rough initial correspondence between
two 3D shape surfaces by removing translation, scale, and rotation.
They then used a landmark-sliding algorithm based on 3D thin-plate
splines (TPS) to model the shape correspondence errors and achieve
affine invariance. The methods \cite{Dalal07,Ha05} stepwise construct
their affine invariance by targeting individual transformations whereas
our subspace method is a true invariant to the entire class of affine
transformations. Ho et al. \cite{Ho07} proposed an elegant noniterative
algorithm for 2D affine registration by searching for the roots of
the associated polynomials. Unfortunately, this method does not generalize
to higher dimensions. 

Begelfor and Werman \cite{begelfor2006affine} popularized the use of
Grassmannian subspaces as an affine invariant. Their work focused on
developing clustering algorithms on the Grassmann manifold, but they did not
utilize this invariance to solve the correspondence problem.  A method that
did try to address this correspondence problem through subspace invariance was
\cite{Wang09b}. The robustness of their method was never evaluated, and,
moreover, their approach of using QR factorizations of rank-deficient
orthogonal projection matrices is quite different from our proposed two-stage
approach. Finally, Chellappa et al. have shown the effectiveness of
Grassmannian representations for object recognition \cite{Chellappa08a} and their use
has expanded to other vision domains \cite{Li14b}.

Algorithms that address non-rigid transformations inherently have
an advantage over strictly affine methods, here we highlight some
notable non-rigid methods that can be used to address the affine correspondence
problem. Raviv et al. \cite{Raviv14} form an equi-affine (volume
preserving) invariant
Laplacian for surfaces, with applications in shape matching. Their method, however, requires
explicit metric tensor calculations on mesh surfaces which can lead to further complications of singular points 
on the surfaces.  We avoid surfaces parameterization issues, offering an
approach that works directly on point sets. Popular non-rigid matching
algorithms include CPD \cite{Myronenko06}, {\sf gmmreg} \cite{jian2011robust}
and TPS-RPM \cite{Chui03b}. Graph matching methods have also been quite
popular recently. Zhou and de la Torre \cite{Zhou12,Zhou2013} presented the factorized graph matching
(FGM) algorithm in which the affinity matrix is factorized as a Kronecker
product of smaller matrices. Although the factorization of affinity
matrices makes large graph matching problems more tractable, the method is
still computationally expensive.

\section{Grassmannians and Affine Invariance\label{sec:Grassmannians-and-Affine}}

The principal contributions of this work are (i) a Grassmannian representation
(GR) of feature vectors and (ii) new affine invariant coordinates
in which feature correspondences can be sought. Below, we describe
both the intuitive and formal aspects of the new representation.

\subsection{Formulation\label{sub:Formulation}}

Let $X\in\mathbb{R}^{N\times D}$ denote a set of $N$ features living
in a $D$ dimensional space (with $D=2$ or 3). 
The application of an affine transformation on the feature
set $X$ can be written as 
\begin{equation}
\tilde{X}=XA+\mathbf{1}T\label{eq:Xtilde}
\end{equation}
where $A\in\mathbb{R}^{D\times D}$ and $T\in\mathbb{R}^{D}$ are
the multiplicative and additive aspects of the affine transformation.
The matrix $A$ comprises global rotation, scale and shear factors
and the vector $T$ contains the global translation factors (while
the vector $\mathbf{1}\in\mathbb{R}^{N}$ is the vector of all ones).
The new feature set $\tilde{X}$ lives in the same $\mathbb{R}^{N\times D}$
space as $X$. The above notation can be considerably simplified by
moving to a homogeneous coordinate representation. 
$\tilde{X}=XA$
where $X\in\mathbb{R}^{N\times(D+1)}$ has an additional (last) column
set to $\mathbf{1}$ and $A\in\mathbb{R}^{(D+1)\times(D+1)}$ now
subsumes the translation factors. (The operator $A$ is also constrained
to have only $(D+1)\times D$ free parameters. 
If two feature sets $X$ and $\tilde{X}$
(in the homogeneous representation and of the same cardinality) differ
by an affine transformation $A$, the best least-squares estimate
of $A$ is
\begin{equation}
\hat{A}=\left(X^{T}X\right)^{-1}X^{T}\tilde{X}.\label{eq:Ahat}
\end{equation}
Two sets of features need not be linked by just an affine transformation.
The relations can include noise, occlusion, spurious features, unknown
correspondence and non-rigid transformations. In this paper, we are
mainly concerned with the group of affine transformations and in using
equivalence classes of feature sets (under affine transformations)
to construct new invariant coordinate representations. Since unknown
correspondence is often the key confounding factor in vision applications,
we model the relationship between two feature sets with the inclusion
of this factor as $Y=PXA$. Here, $P$ is a permutation matrix (a square binary
valued $N\times N$ matrix with rows and columns summing to one) included
to model the loss of correspondence between two sets of features $X$
and $Y$. While the inclusion of a permutation matrix does not account
for occlusion and spurious features, we show in (numerous) experiments
that affine transformation recovery is not adversely hampered provided
\emph{strong} correspondences persist. 

\begin{figure}
\begin{centering}
\begin{tabular}{ccccc}
\includegraphics[width=1.4cm,height=1.7cm]{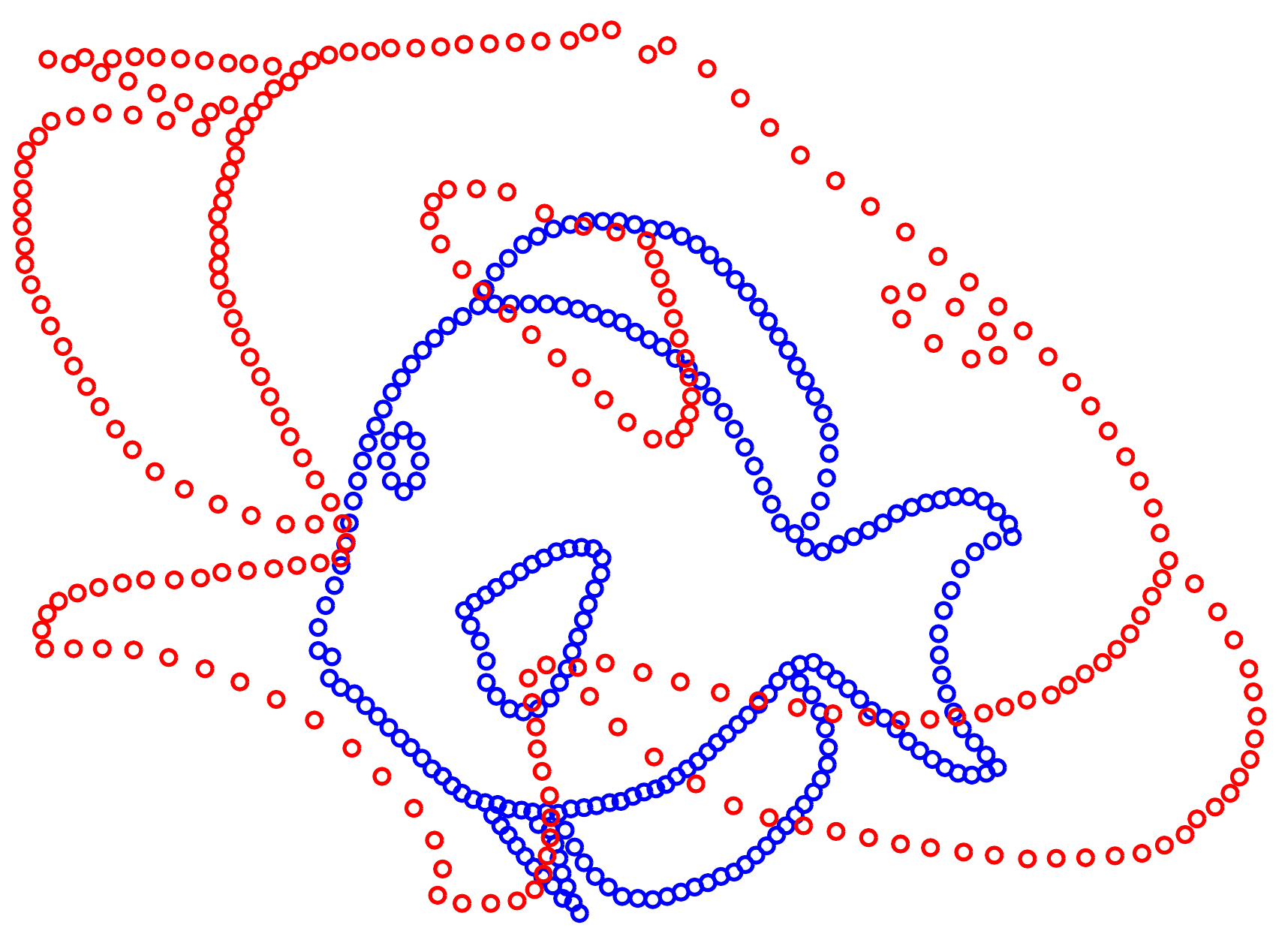} & {\tiny{}{\tiny{}}}\includegraphics[width=1.4cm,height=1.7cm]{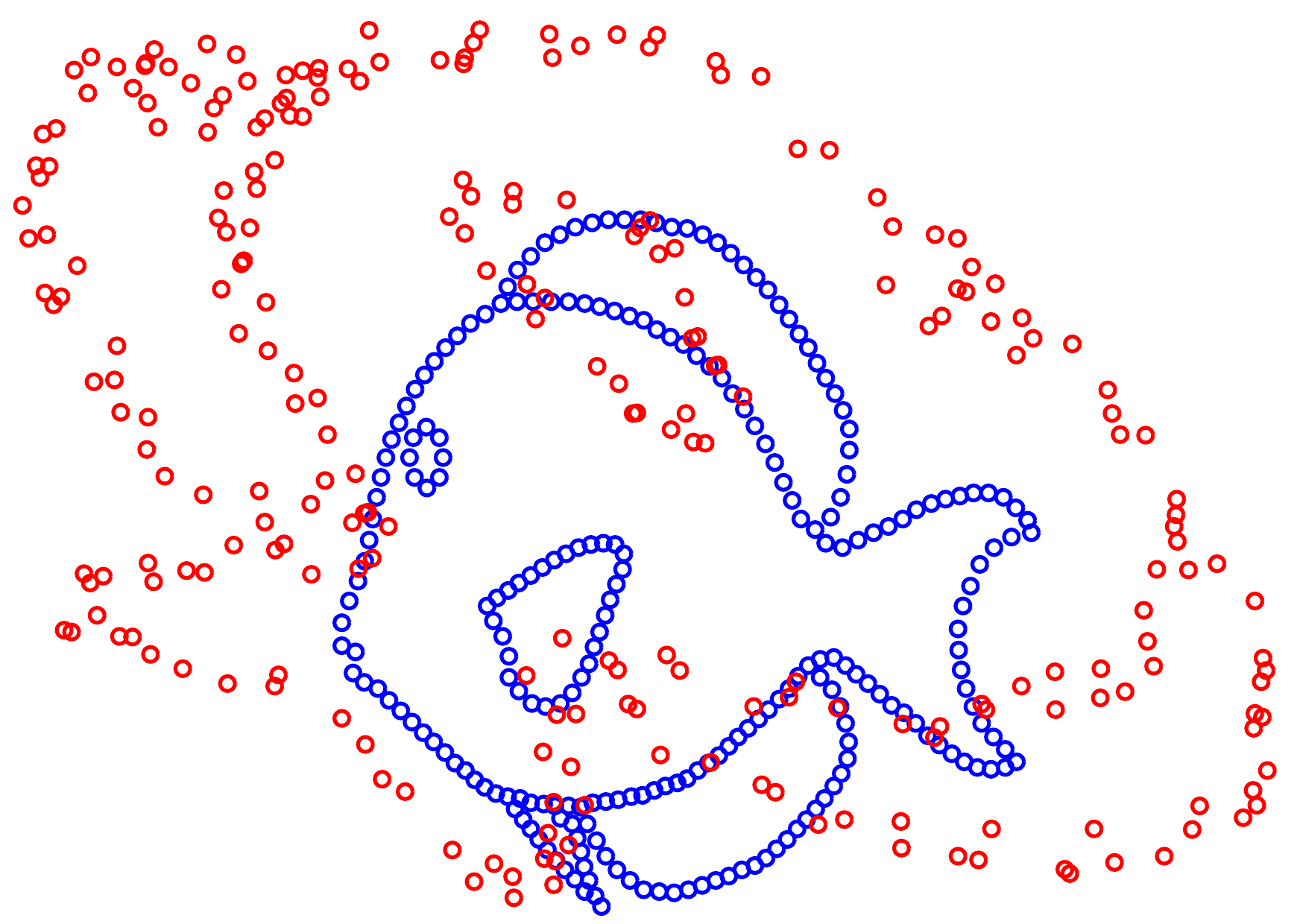} & {\tiny{}{\tiny{}}}\includegraphics[width=1.4cm,height=1.7cm]{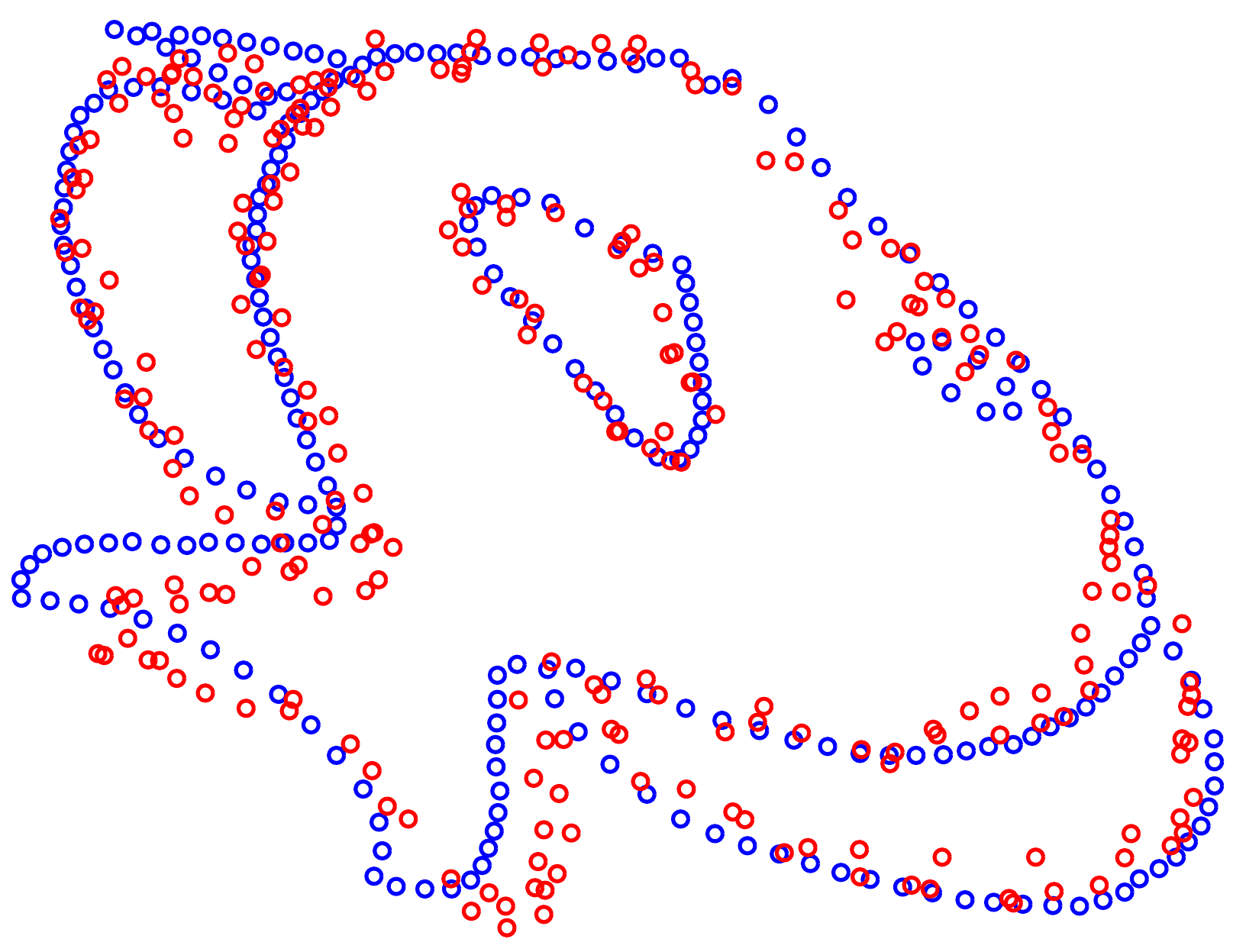} & {\tiny{}{\tiny{}}}\includegraphics[width=1.4cm,height=1.7cm]{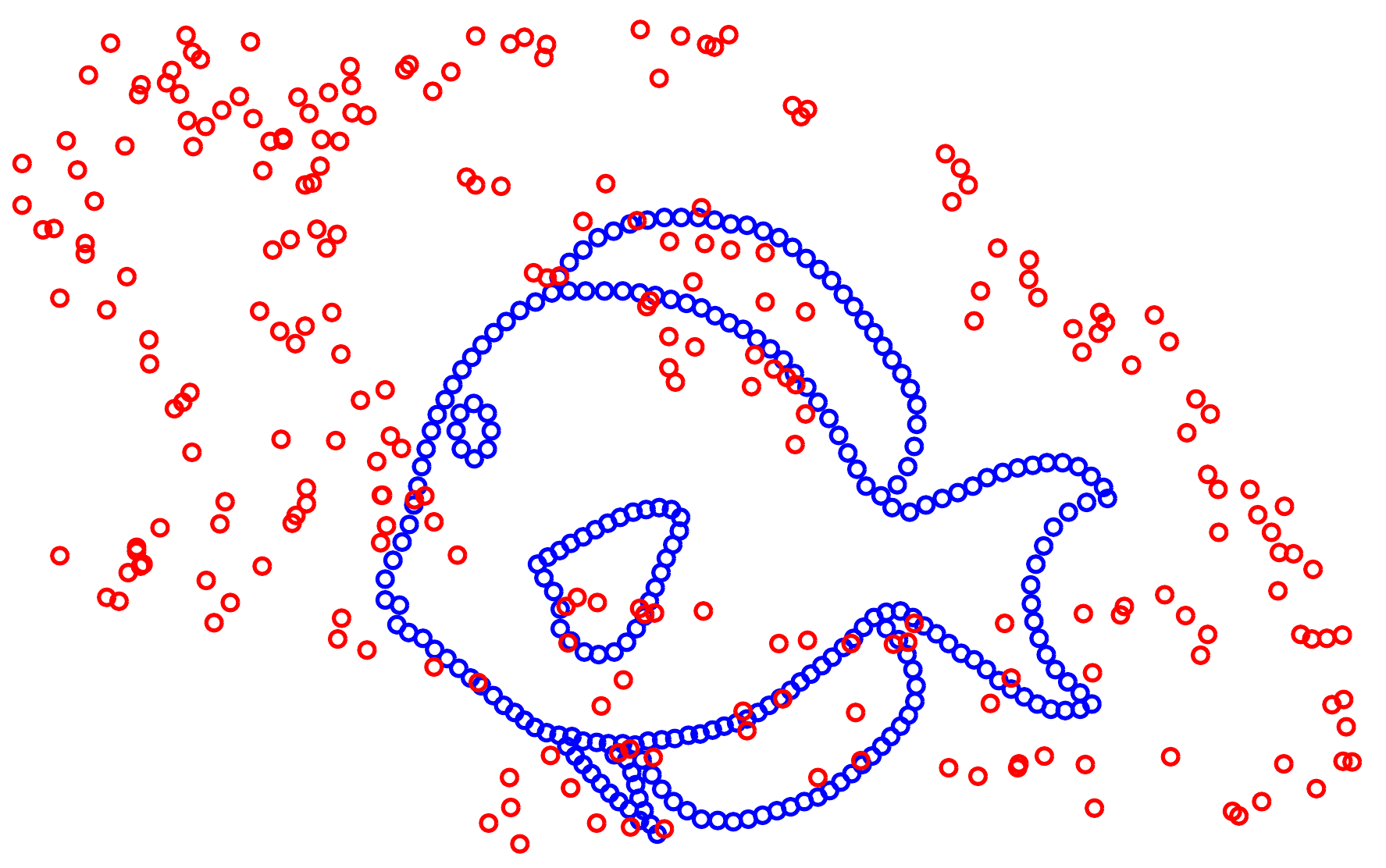} & {\tiny{}{\tiny{}}}\includegraphics[width=1.4cm,height=1.7cm]{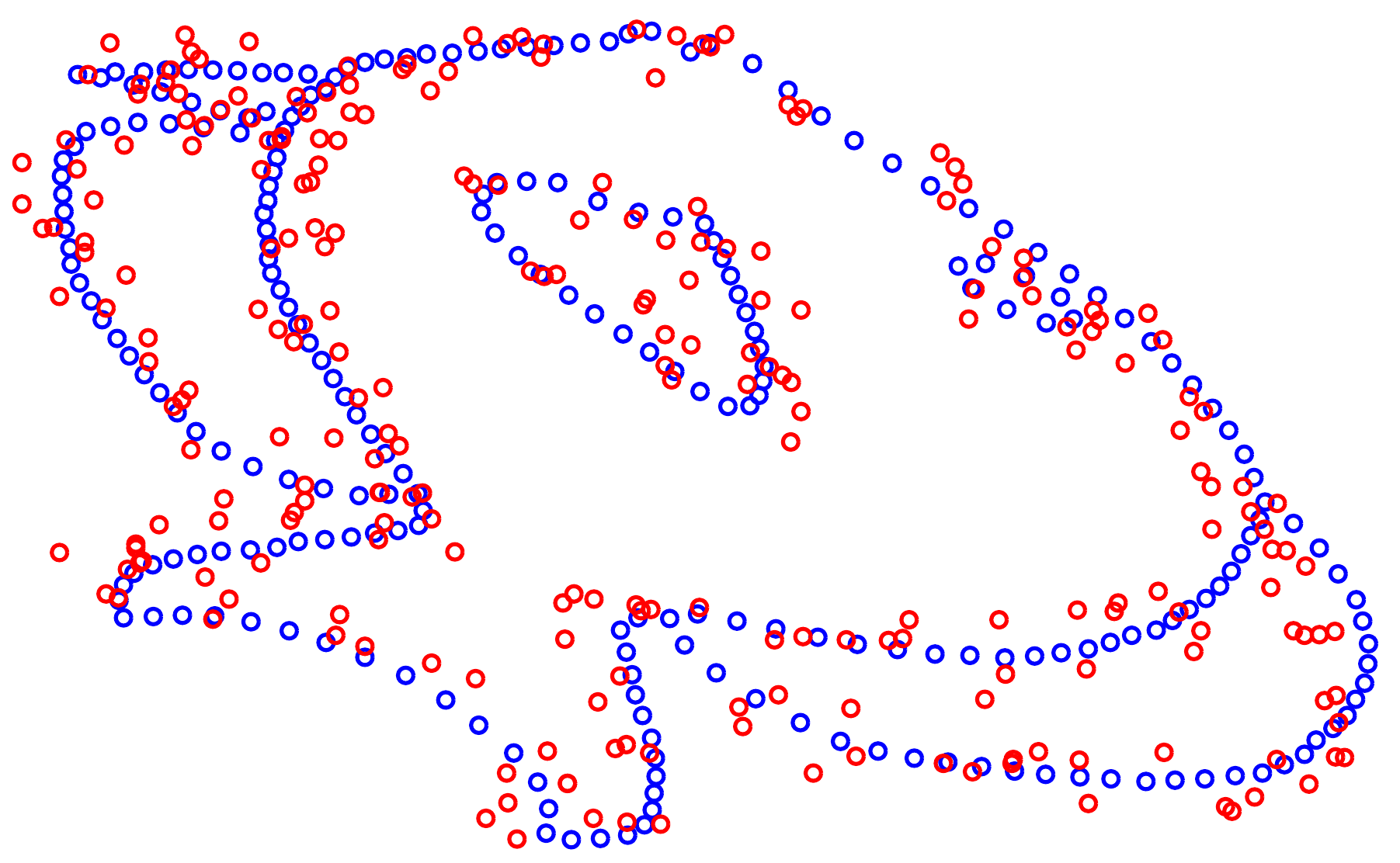}\tabularnewline
\includegraphics[width=1.4cm,height=1.7cm]{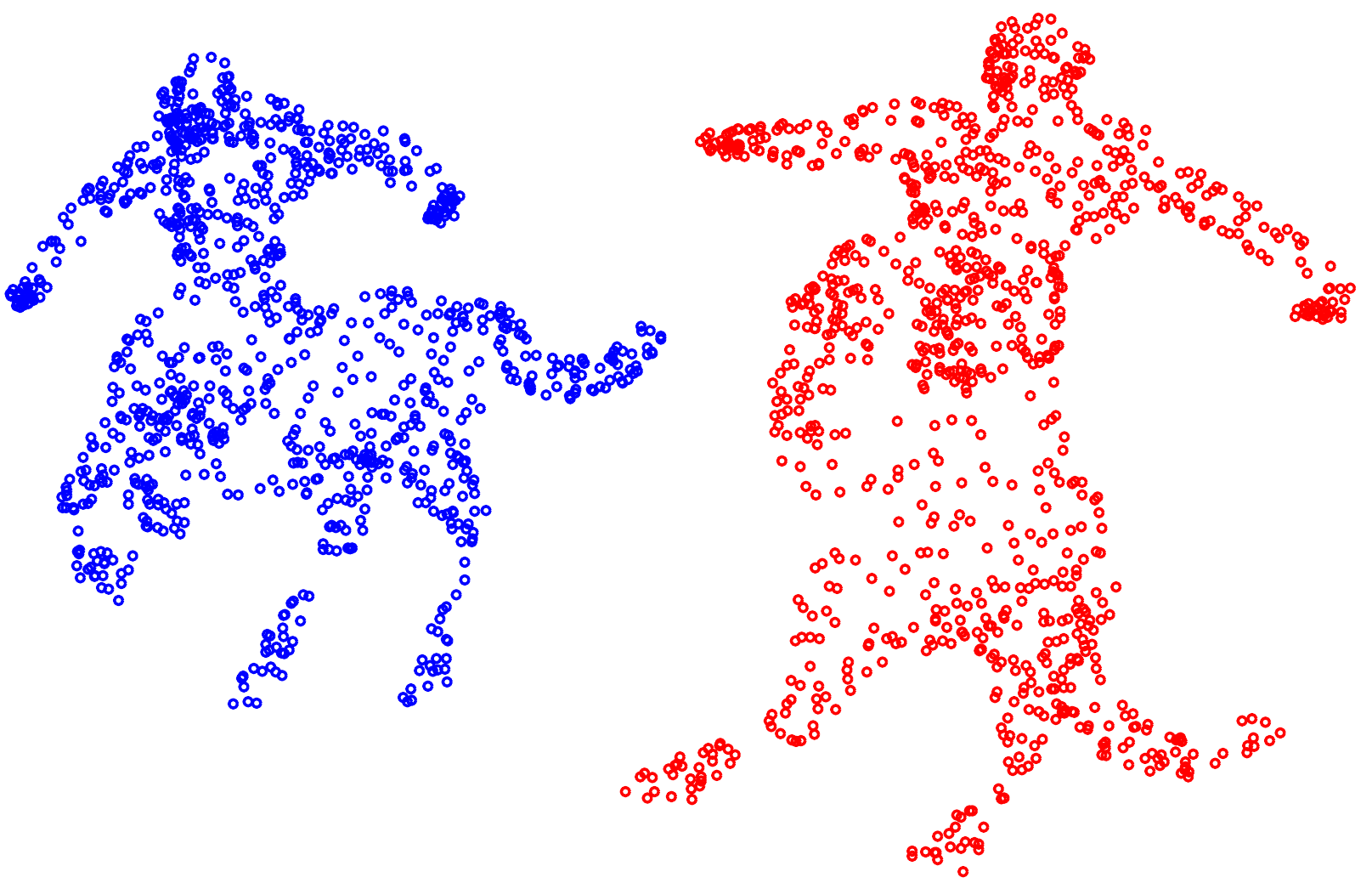} & {\tiny{}{\tiny{}}}\includegraphics[width=1.4cm,height=1.7cm]{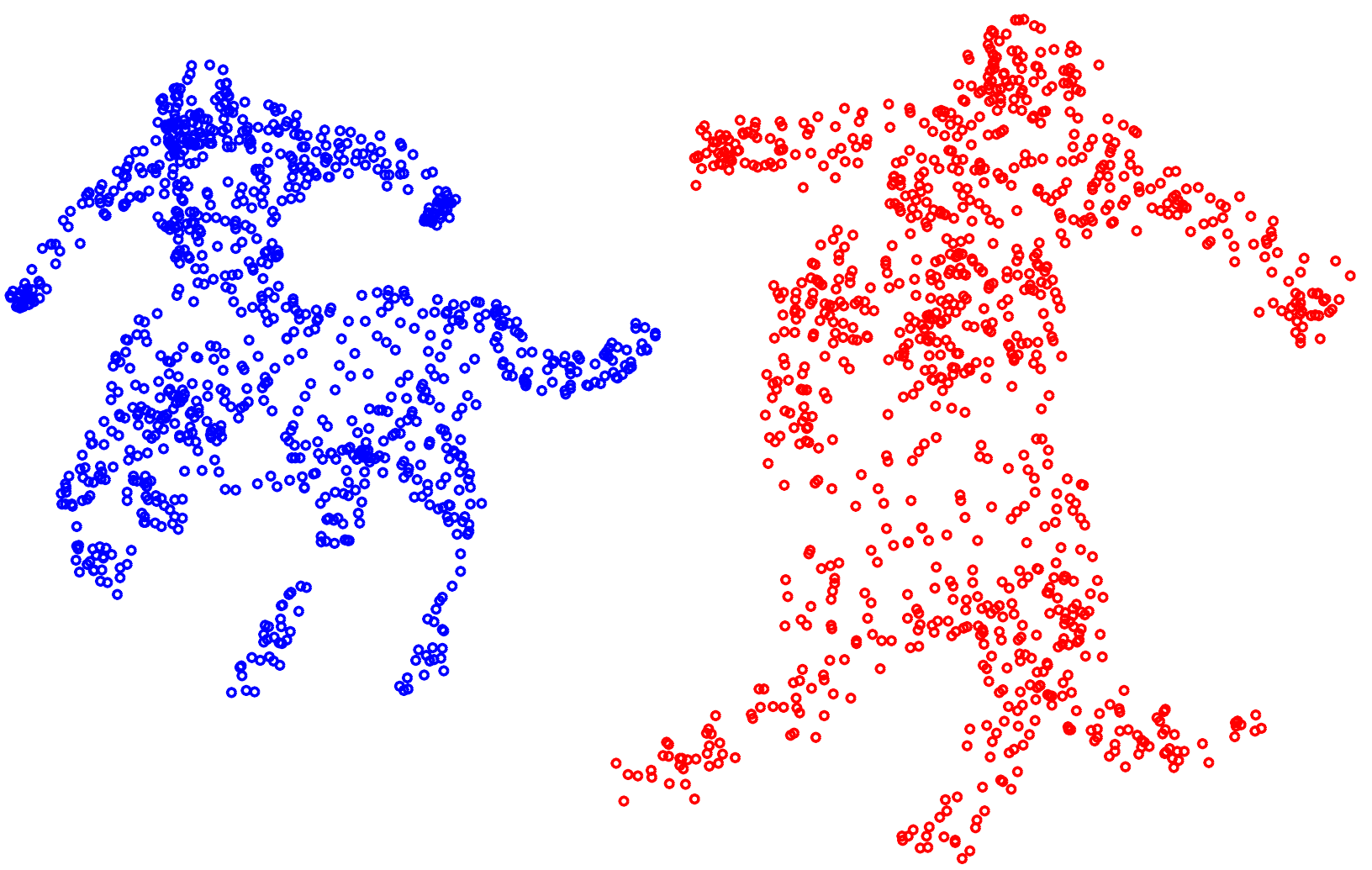} & {\tiny{}{\tiny{}}}\includegraphics[width=1.4cm,height=1.7cm]{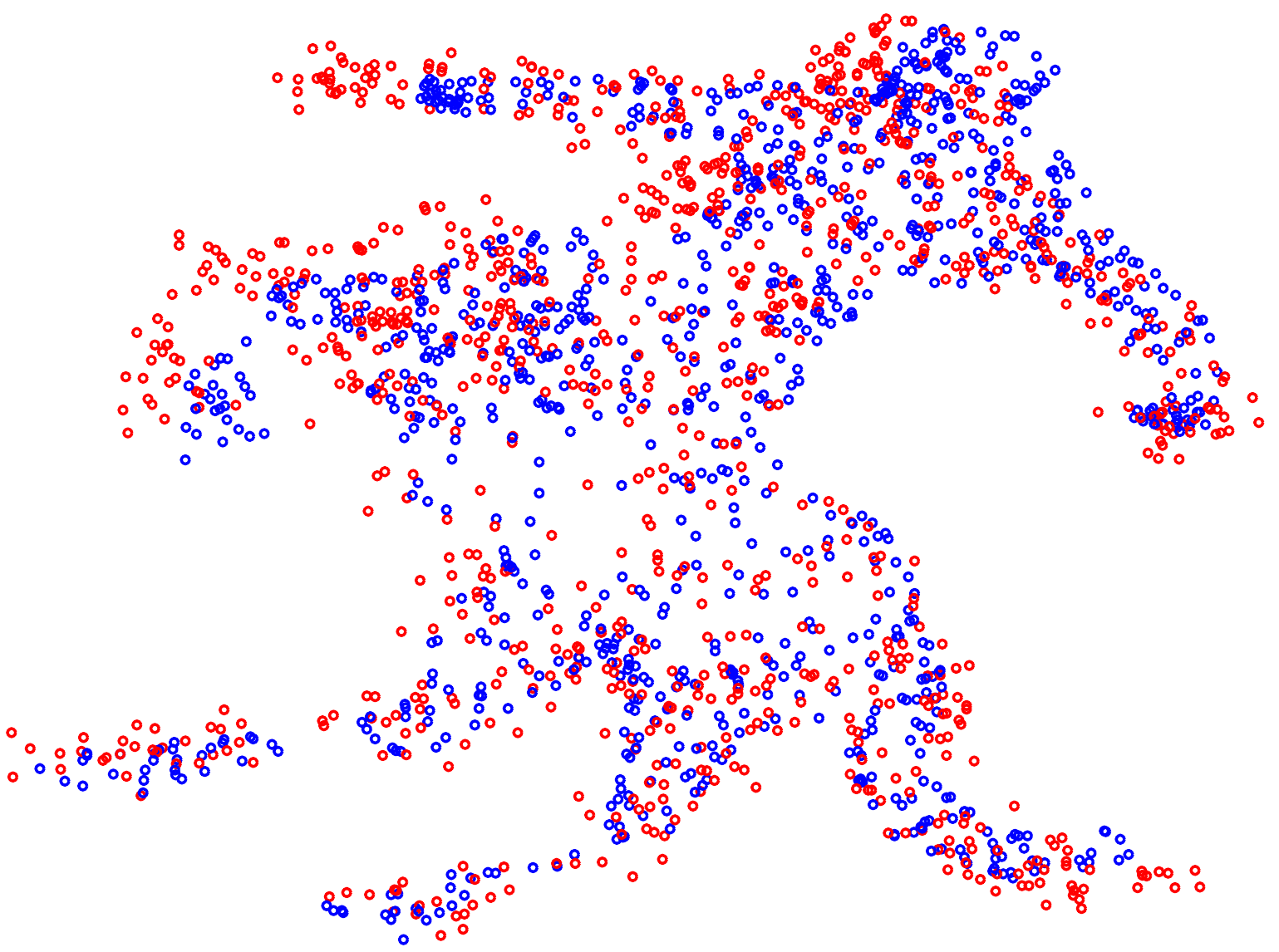} & {\tiny{}{\tiny{}}}\includegraphics[width=1.4cm,height=1.7cm]{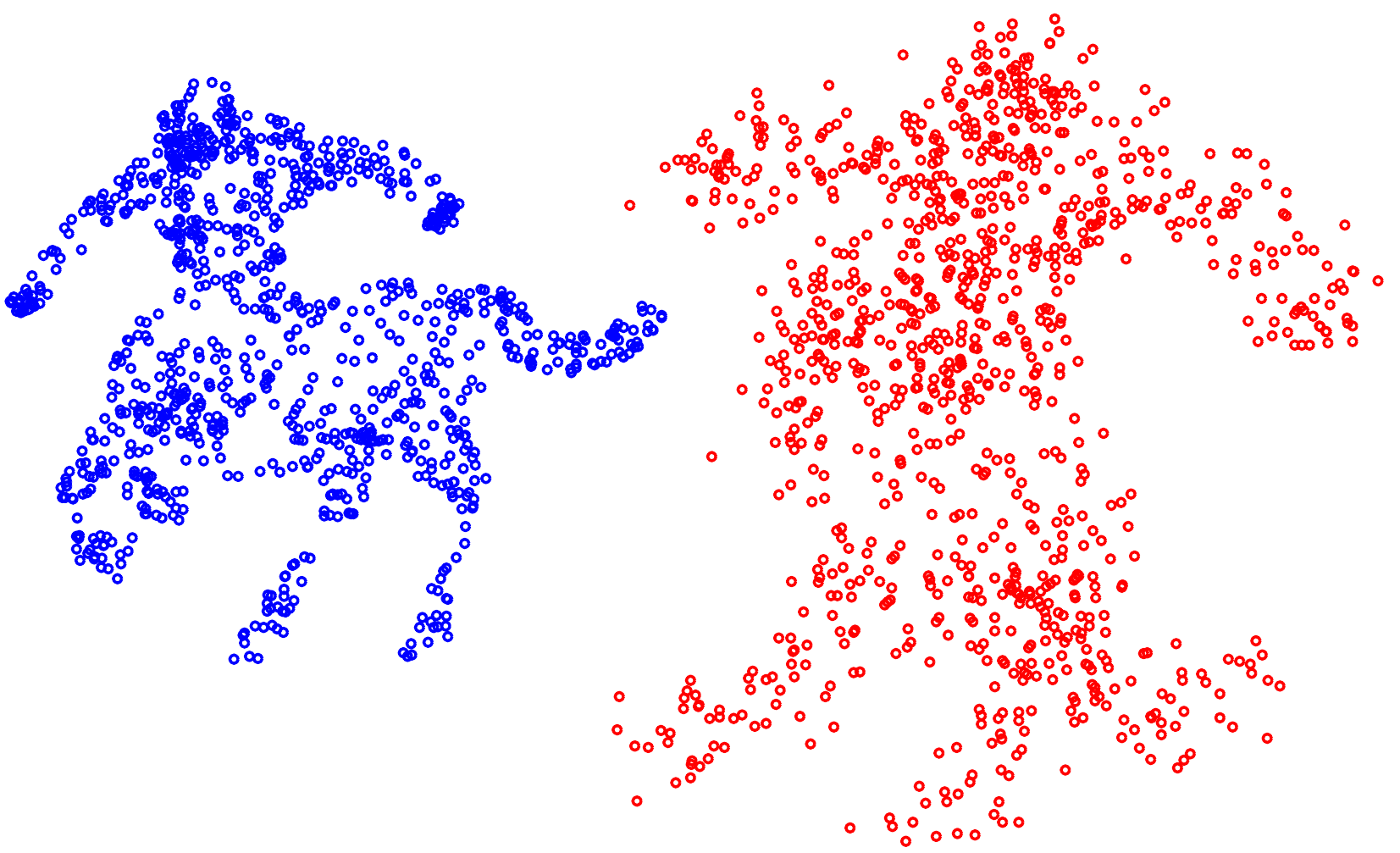} & {\tiny{}{\tiny{}}}\includegraphics[width=1.4cm,height=1.7cm]{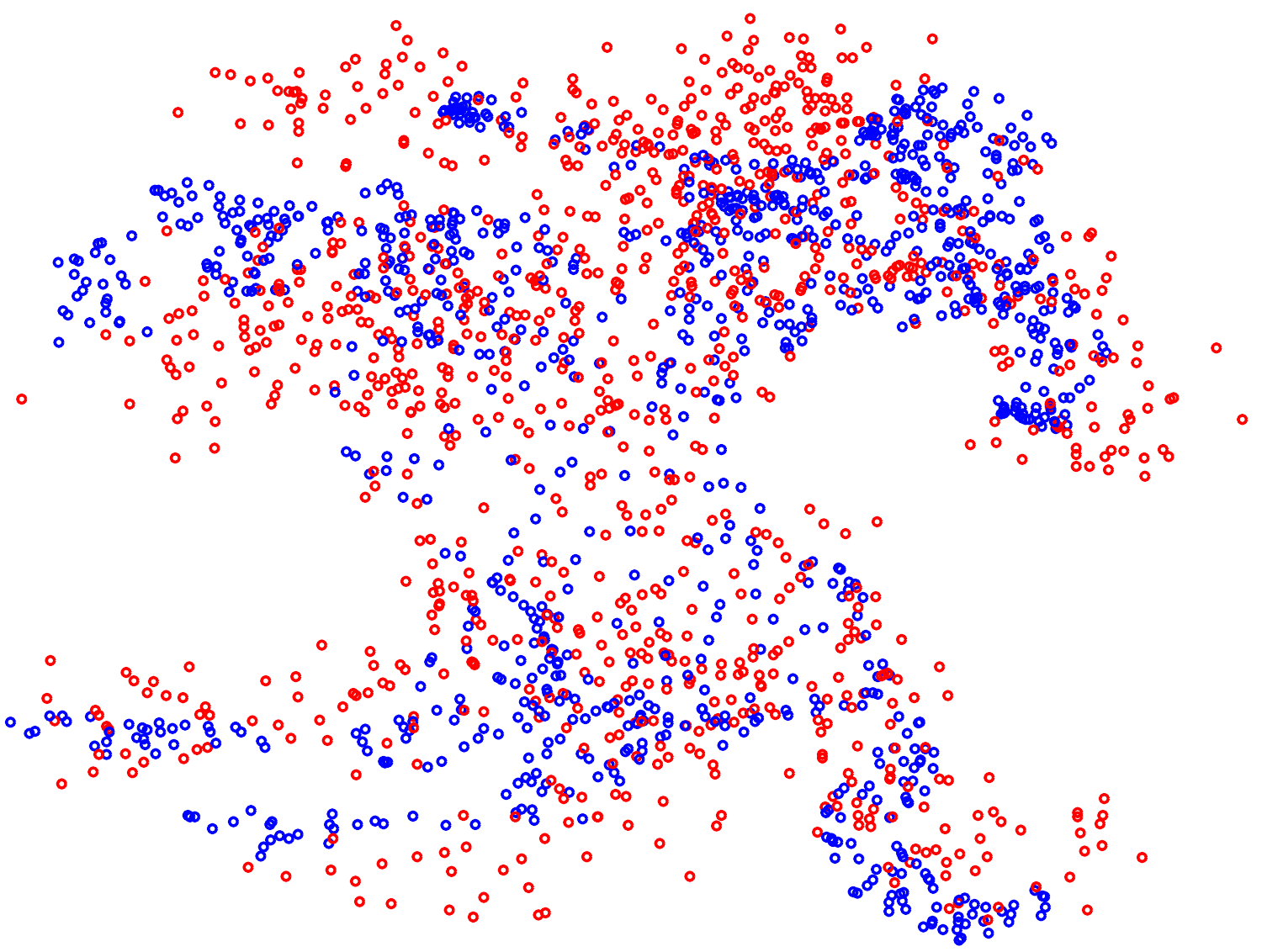}\vspace{-7pt}\tabularnewline
{\tiny{}Original Shapes } & {\tiny{}{\tiny{}\hspace{-8pt}}Small Noise} & {\tiny{}{\tiny{}\hspace{-8pt}}Registered Small Noise} & {\tiny{}{\tiny{}\hspace{-8pt}}Large Noise} & {\tiny{}{\tiny{}\hspace{-8pt}}Registered Large Noise}\tabularnewline
\end{tabular}
\par\end{centering}

\centering{}\protect\caption{Sample registration results for 2D and 3D affine related shapes in the presence
of small and large levels of noise. Even with
large amounts of noise, the GrassGraph method is able to recover the
transformation accurately with just a simple nearest neighbor correspondence. \label{fig: Registered Noisy shapes}}
\end{figure}

We now establish the connection to Grassmannians. This connection
is first intuitively shown followed by a formal treatment. 
The action of the affine transformation clearly
results in a new feature set whose columns are linear combinations
of the columns in $X$. In other words, if the columns of $X$ span
a (two or three dimensional) subspace of $\mathbb{R}^{N}$, then $XA$,
by virtue of being a linear combination of the columns, continues
to live in the \emph{same two or three dimensional subspace} as $X$.
Therefore, the particular subspace carved out by $X$ can be considered
to be an \emph{affine invariant equivalence class} of feature sets.
If we have multiple feature sets which differ from each other by affine
transformations, then all the feature sets inhabit the same two or
three dimensional subspace of $\mathbb{R}^{N}$. Since Grassmannians
are the set of all subspaces of $\mathbb{R}^{N}$, the equivalence
class merely picks out one element of the Grassmannian. Or in other
words, quotienting out affine transformations from a set of features
is tantamount to moving to a linear subspace representation whose
dimensionality depends only on the original feature dimensions. 
We now state the
following well known theorem\footnote{Quoted almost verbatim from Wikipedia
  \url{https://en.wikipedia.org/wiki/Grassmannian}} which allows us to move from Grassmannians
to orthogonal projection operators.
\begin{thm}
\label{thm:Grassmannian} \cite{Boothby02} Let $\mathrm{Gr}(d,\mathbb{R}^{N})$ denote
the Grassmannian of d-dimensional subspaces of $\mathbb{R}^{N}$.
Let $M(N,\mathbb{R})$ denote the space of real $N\times N$ matrices.
Consider the set of matrices $G(d,N)\subset M(N,\mathbb{R})$ defined
by $Q\in G(d,N)$ if and only if the three conditions are satisfied:
(i) $Q$ is a projection operator with $Q^{2}=Q$. (ii) $Q$ is symmetric
with $Q^{T}=Q$. (iii) $Q$ has a trace with $\mathrm{trace}(Q)=d$.
Then $G(d,N)$ and $Gr(d,\mathbb{R}^{N})$ are homeomorphic, with
a correspondence established (since each $Q$ is unique) between each
element of the Grassmannian and a corresponding $Q$.
\end{thm} 
Theorem~\ref{thm:Grassmannian} establishes the equivalence between
each element of $\mathrm{Gr}(d,\mathbb{R}^{N})$ and a corresponding
orthogonal projection matrix $Q$. Given a feature set $X$, the theorem
implies that we construct an orthogonal projector $Q_{X}$ which projects
vectors into the $d=D+1$ dimensional subspace spanned by the columns
of $X$. This can be readily constructed via $Q_{X}=X(X^{T}X)^{-1}X^{T}$
for $X$ and likewise for a feature set $Y$. Provided the relevant
matrix inverses exist, two feature sets $X$ and $Y$ with both in
$\mathbb{R}^{N\times(D+1)}$ project to the same element $G\in\mathrm{Gr}(d,\mathbb{R}^{N})$
(of the Grassmannian) if and only if $Q_{X}=Q_{Y}$. Orthogonal projectors
have a drawback which we now address.

\subsection{Grassmannian Graphs\label{sub:Grassmannian-Graphs}}

Orthogonal projectors can be represented using the singular value
decomposition of $X$. If $X=U_{X}S_{X}V_{X}^{T}$, then
$Q_{X}=U_{X}U_{X}^{T}$. If feature sets $X$ and $Y$ project to identical
elements of the Grassmannian, then
$Q_{X}=U_{X}U_{X}^{T}=U_{Y}U_{Y}^{T}=Q_{Y}$.
This suggests that we look for new affine invariant coordinates of
$X$ via its SVD decomposition matrix $U$. Unfortunately, this is
not straightforward since 
\begin{equation}
Q_{X}=Q_{Y}\implies U_{X}U_{X}^{T}=U_{Y}U_{Y}^{T}\implies U_{X}R=U_{Y}\label{eq:QxQyimplication}
\end{equation}
where $R$ is an \emph{unknown} orthonormal matrix in $\mathbb{O}(D+1)$.
Or in intuitive terms, orthonormal matrices $U_{X}$ and $U_{Y}$
differ by an arbitrary rotation (and reflection). If we seek affine
invariant \emph{coordinate} representations, as opposed to affine
invariant Grassmannian elements, then we must overcome this rotation
problem. 

The unknown rotation problem can be overcome by introducing the \emph{Grassmannian
graph representation}. In a nutshell, we (after computing the SVD
of $X$), build a rotation invariant weighted graph from the rows
of $U$ (treated as points in $\mathbb{R}^{D+1}$). The Euclidean
distances between rows of $U$ are invariant under the action of an
arbitrary orthonormal matrix $R$. Consequently, weighted graphs constructed
from $U$ with each entry depending on the Euclidean distance between
rows is an affine invariant of $X$. This Grassmannian graph representation,
which we now introduce via the popular Laplace-Beltrami operator approach,
is therefore central to the goals of this paper.

The \emph{Laplace-Beltrami operator} (LBO) generalizes the Laplacian
of Euclidean spaces to Riemannian manifolds. For computational applications,
one has to discretize the LBO which results in a finite dimensional
operator. Though several discretization schemes exist, probably the
most widely used is the graph Laplacian \cite{Chung97}. The Laplacian
matrix of a graph is a symmetric positive semidefinite matrix given
as $L=D-K$, where $K$ is the adjacency matrix and $D$ is the diagonal
matrix of vertex degrees. The spectral decomposition of the graph
Laplacian is given as $Lv=\lambda v$ where $\lambda$ is an eigenvalue
of $L$ with a corresponding eigenvector $v$. The eigenvalues of
the graph Laplacian are non-negative and constitute a discrete set.
The spectral properties of $L$ are used to embed the feature points
into a lower dimensional space, and gain insight into the geometry
of the point configurations \cite{Isaacs11,Moyou14}. (Note: For the
LBO, $\lambda=0$ is always an eigenvalue for which its corresponding
eigenvector is constant and hence discarded in most applications.)

We can utilize the LBO to realize the Grassmannian graph's goal to
eliminate the arbitrary orthonormal matrix $R$ present in the relationship
between $U_{X}$ and $U_{Y}$. To achieve this, we leverage
the graph Laplacian approximation described above to construct new
coordinates from the Grassmannian graph's $N\times N$ Laplacian matrix
by taking its top $(D+1)$ eigenvectors (with the rows of the eigenvectors
serving as coordinates). Since the Grassmannian graph is affine invariant,
so are its eigenvectors. We can now conduct feature comparisons in
this eigenspace to obtain correspondences, clusters and the like.
For our present application of affine invariant matching, we recover
the correspondences between point configurations $X$ and $Y$ by
representing each in the LBO eigenspace, and then use nearest neighbor
(\emph{kNN}) selection to recover the permutation matrix $P$. The ability
to simply use \emph{kNN} arises from the fact that the affine transformation
has been rendered moot in LBO coordinates. Algorithm~\ref{alg:GrassGraph}
details the steps in our GrassGraph approach.

\begin{algorithm}[t]
\begin{raggedright}
\protect\caption{The GrassGraph Algorithm\label{alg:GrassGraph}}

\par\end{raggedright}

\medskip{}

\begin{algorithmic}

\STATE \textbf{Input: }$X,Y\in\mathbb{R}^{N\times(D+1)}$

\STATE \textbf{Output:} Estimated correspondences $\hat{P}$ and
affine transformation $\hat{A}$

\STATE 1.\textbf{~SVD: }$X=U_{X}S_{X}V_{X}^{T}$ and $Y=U_{Y}S_{Y}V_{Y}^{T}$

\STATE 2.\textbf{~Graph Laplacian (GL): }

\qquad{}Retain the top $D+1$ columns of $U_{X}$ and $U_{Y}\rightarrow\hat{U}_{X}$
and $\hat{U}_{Y}$, respectively.$^{*}$

\qquad{}Build weighted graph $L_{X}$ from the rows of $\hat{U}_{X}$ 

\qquad{}Build weighted graph $L_{Y}$ from the rows of $\hat{U}_{Y}$\textbf{ }

\STATE 3.~\textbf{GL Eigenvectors:}

\qquad{}Take top 3 eigenvectors$^{\star}$ of $L_{X}$ $(E_{X})$
and $L_{Y}$ $(E_{Y})$ 

\STATE 4.~\textbf{Estimate }$P$\textbf{ and $A$:} 

\qquad{}$\hat{P}$: Correspondence using rows of $E_{X}$ and $E_{Y}$$^{\dagger}$ 

\qquad{}$\hat{A}:$ Apply $P$ to $X$, then $\hat{A}=(X^{T}X)^{-1}X^{T}Y$ 

$^{*}$Treated as points in $\mathbb{R}^{D+1}$.

$^{\star}$Can use any $k$ combinations where $2\leq k\leq N$. 

$^{\dagger}$Simple mutual nearest neighbor assignment works well.

\end{algorithmic}
\end{algorithm} 
More specifically, the correspondence algorithm used is a simple mutual
nearest neighbor search. Consider a point set $X$ and its target
$Y$ in $\mathbb{R}^{D}$. First, $X$ is held fixed and the nearest
neighbors in $Y$ are found through the minimum Euclidean distance.
Next, $Y$ is held fixed and the nearest neighbors in $X$ are found
using the same distance measure. For a pair of points to be in correspondence,
they must both be each other's nearest neighbors. This reduces the
chances of assigning a single point in $Y$ to many points in $X$.
Although we are afforded a simple nearest neighbor search, we pay
a small price due to sign ambiguities of the eigenvectors
resulting from the eigendecomposition step.

\subsection{Eigenanalysis Sign Ambiguities\label{sub:Eigenanalysis-Sign-Ambiguities}}

As formulated, the GrassGraph algorithm requires two eigendecompositions---one
from the SVD to obtain the orthogonal projector and the other to get
the eigenvectors of the graph Laplacian. It is well known that numerical
procedures for eigenanalysis can introduce arbitrary sign flips on
the eigenvectors. Though there have been previous attempts at addressing
the sign ambiguity issue \cite{Caelli04}, they are commonly considered
as application specific or highly unreliable. Hence, the only solution
remains to evaluate all possible sign flips, i.e. for $k$ eigenvectors
we have $2^{k}$ possibilities. In GrassGraph, we have two such decompositions,
so one may construe that we require evaluation of $2^{k_{1}+k_{2}}$
sign flips, where $k_{1}$ is the number of eigenvectors selected
($k_{1}=3$ for 2D point sets and $k_{1}=4$ for 3D point sets), and
$k_{2}$ is the number of graph Laplacian eigenvectors (typically
$k_{2}=3$ for 2D and 3D). It turns out, however, that the graph Laplacian
eigendecomposition is invariant to any sign flips induced by the initial
SVD. This is due to the fact that in forming the graph we use nearest-neighbor
relationships which are determined using the standard Euclidean distance.
The proposition below details how the distance metric nullifies the
sign ambiguity.

Suppose $u=(u_{1},u_{2},\ldots,u_{k_{1}})$ are coordinates obtained
via a numerical eigendecomposition procedure. This introduces the
possibility that any coordinate $u_{i}$ may be sign flipped, i.e.
$\pm u_{i}$. The calculation of pairwise distances between
any two different points $u^{(1)}$ and $u^{(2)}$ \emph{in the same
coordinate space} under the presence of an sign ambiguity is simplified due to
the fact that 
\begin{equation}
\sum_{i=1}^{k_{1}}\left(\pm u_{i}^{(1)}-\pm u_{i}^{(2)}\right)^{2}
= \sum_{i=1}^{k_{1}}\left(u_{i}^{(1)}-u_{i}^{(2)}\right)^{2}\label{eq:distVar2}
\end{equation}
Hence, when we are forming the graph using the GR coordinates, we
are invariant to sign flips introduced by the SVD and subsequently
only have to resolve the sign ambiguity in the eigenvectors of the
graph Laplacian. Since GrassGraph only uses three eigenvectors for
the spectral coordinates, this is a low order search space that allows
us to easily determine the best eigenvector orientation from the set
of eight possibilities. Having addressed the eigenvector sign flipping
problem, we now give a comprehensive account of our benchmarking process
for GrassGraph.

\section{Experimental Results\label{sec:Experimental-Results}}

In this section we detail the 441,000 experiments (2D and 3D combined)
performed to benchmark our framework. The goal was to rigorously evaluate
the capabilities of the GrassGraph (GG) approach against other well
known registration methods: Coherent Point Drift (CPD) \cite{Myronenko06},
Registration using Mixtures of Gaussians (GMM) \cite{jian2011robust} (note
that the affine versions of CPD and GMM were used in the experiments)
and Algebraic Affine (AA) \cite{Ho07}. This was done by testing the
accuracy of the affine transformation matrix recovered in the presence
of simulated artifacts: noise and missing points with outliers (MPO). For
2D experiments, the three previous competing methods were used but
only CPD and GMM were used for 3D since AA is strictly for 2D. Open source implementations
by the authors of the competing methods were used in the experiments.

In each trial, the target shape was created by applying an affine
transformation to the source shape (an ``affine shape'')
with additional artifacts (noise and MPO) added depending on the experiment.
The breakdown of the number of experiments conducted for all cases is given
in Table \ref{tab:Breakdown-of-experiements}. 
Each experiment measured the ability of the various methods to recover
the true affine transformation that generated the target shape---the
error metric was the Frobenius norm between the true affine matrix
and the recovered affine matrix.  

For the numerous affine registration trials, we used 20 2D and 3D
shapes from the established GatorBait 100 \cite{GatorBait100}
and SHREC'12 \cite{SHREC12a} datasets. The GatorBait
100 database consisted of 100 images of individual fishes from
27 different fish families. The images contain unordered contours
of the fish including the body, fins, eyes and other interior parts.
The SHREC'12 3D shape 
dataset contains 13 categories with 10 shapes
per category in its basic version.  

\begin{figure}
\centering{}%
\begin{tabular}{ccc}
\includegraphics[scale=0.15]{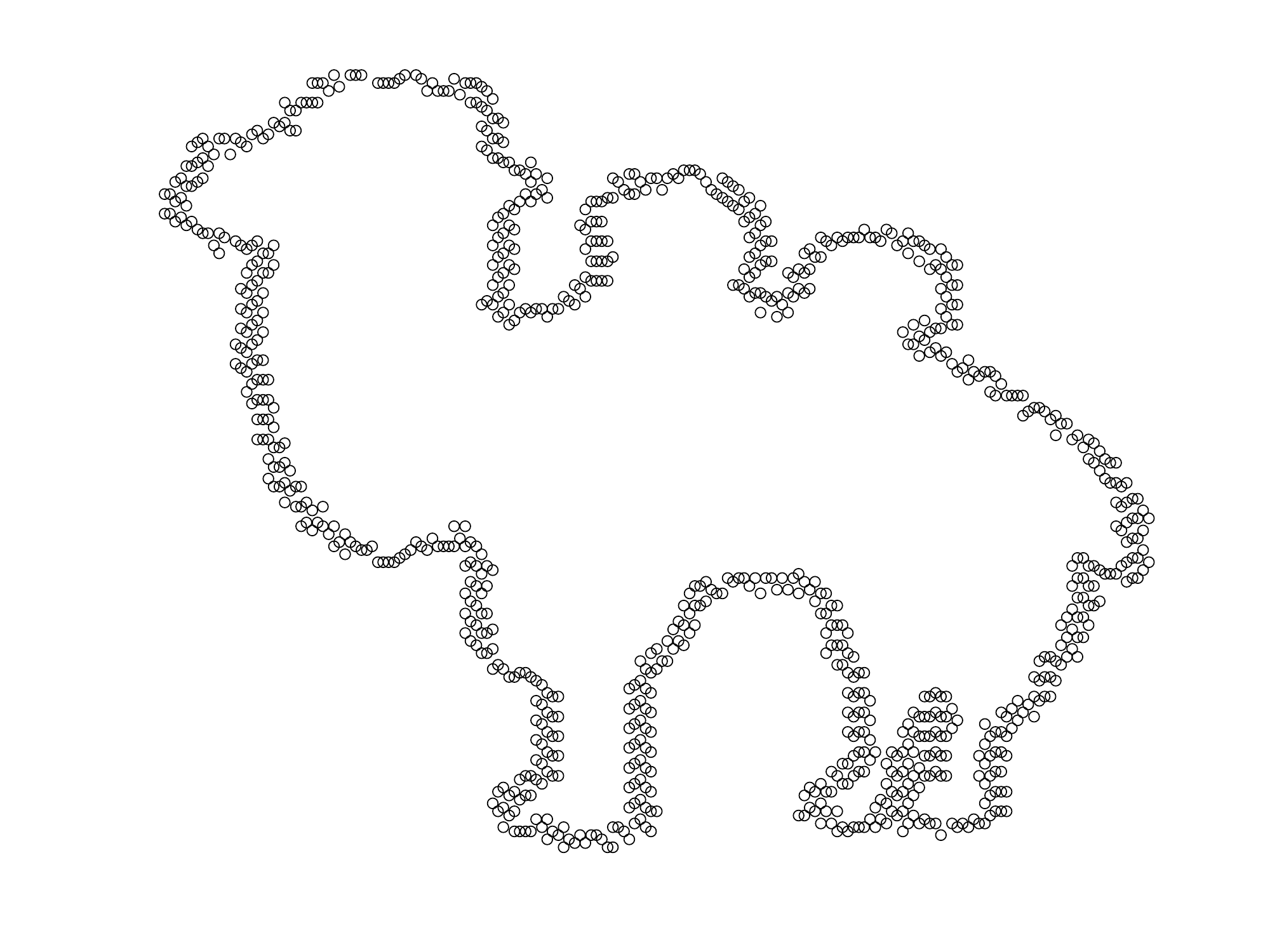} & \includegraphics[scale=0.15]{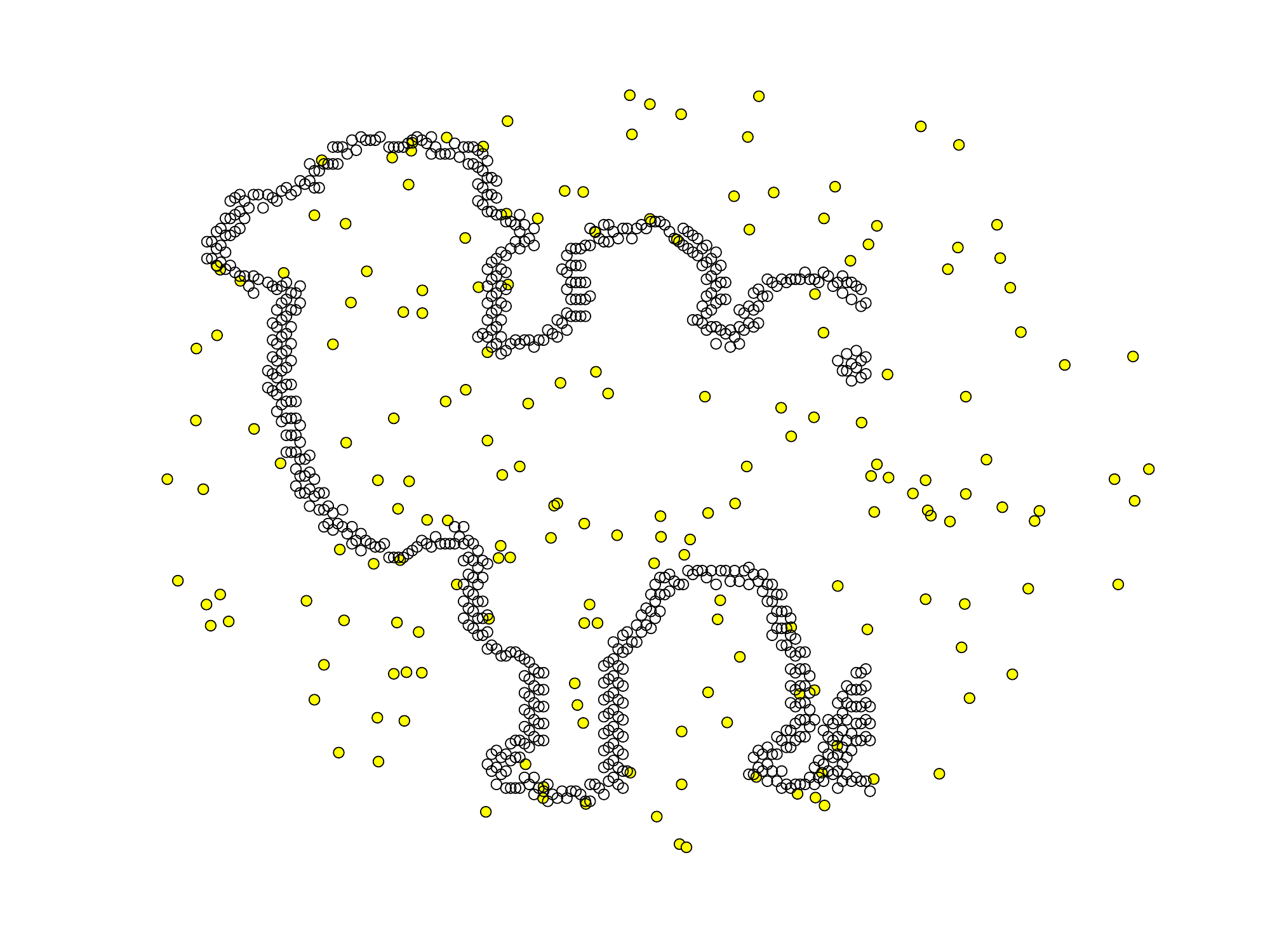} & \includegraphics[scale=0.15]{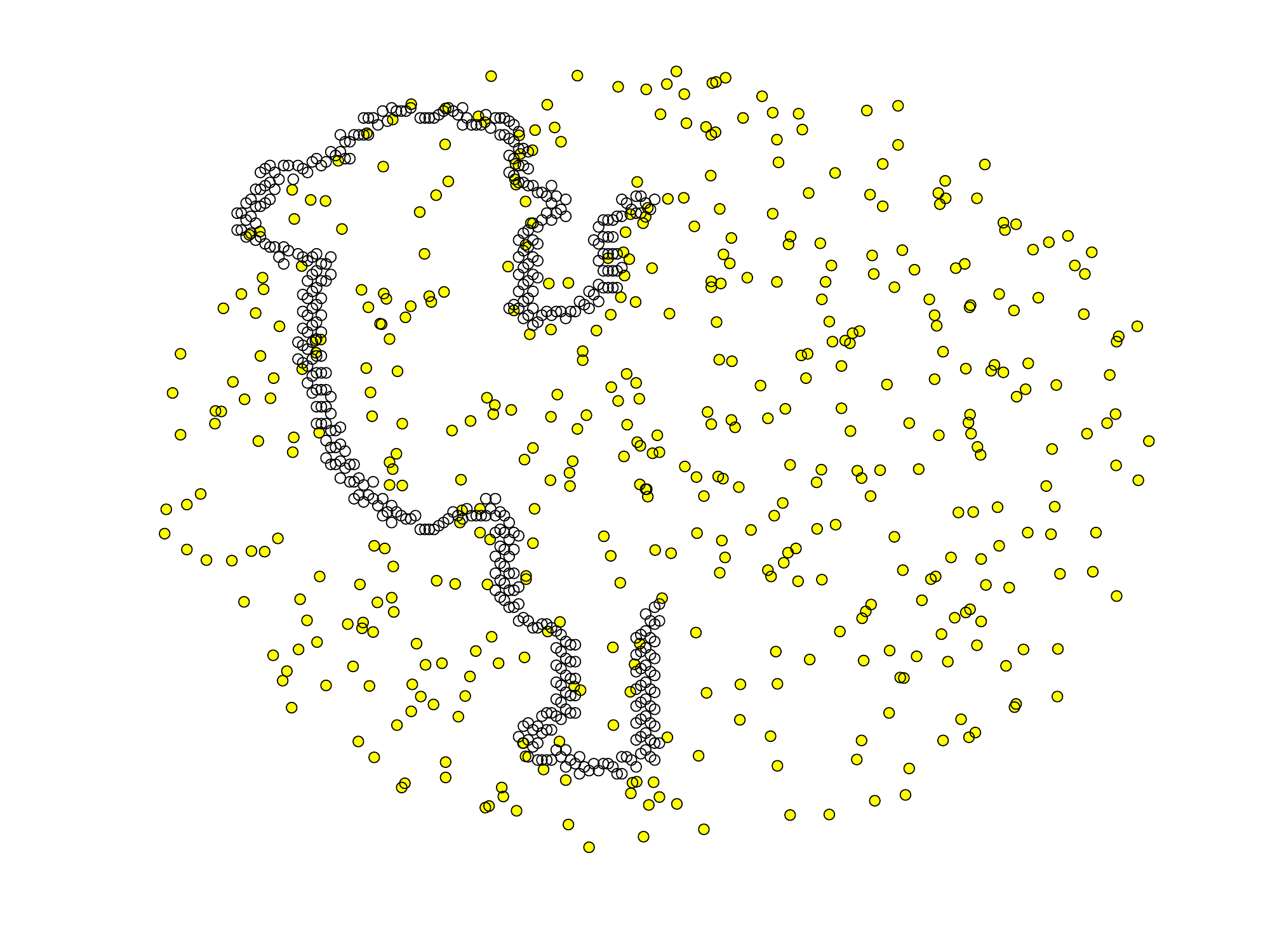}\vspace{-10pt}\tabularnewline
\includegraphics[scale=0.15]{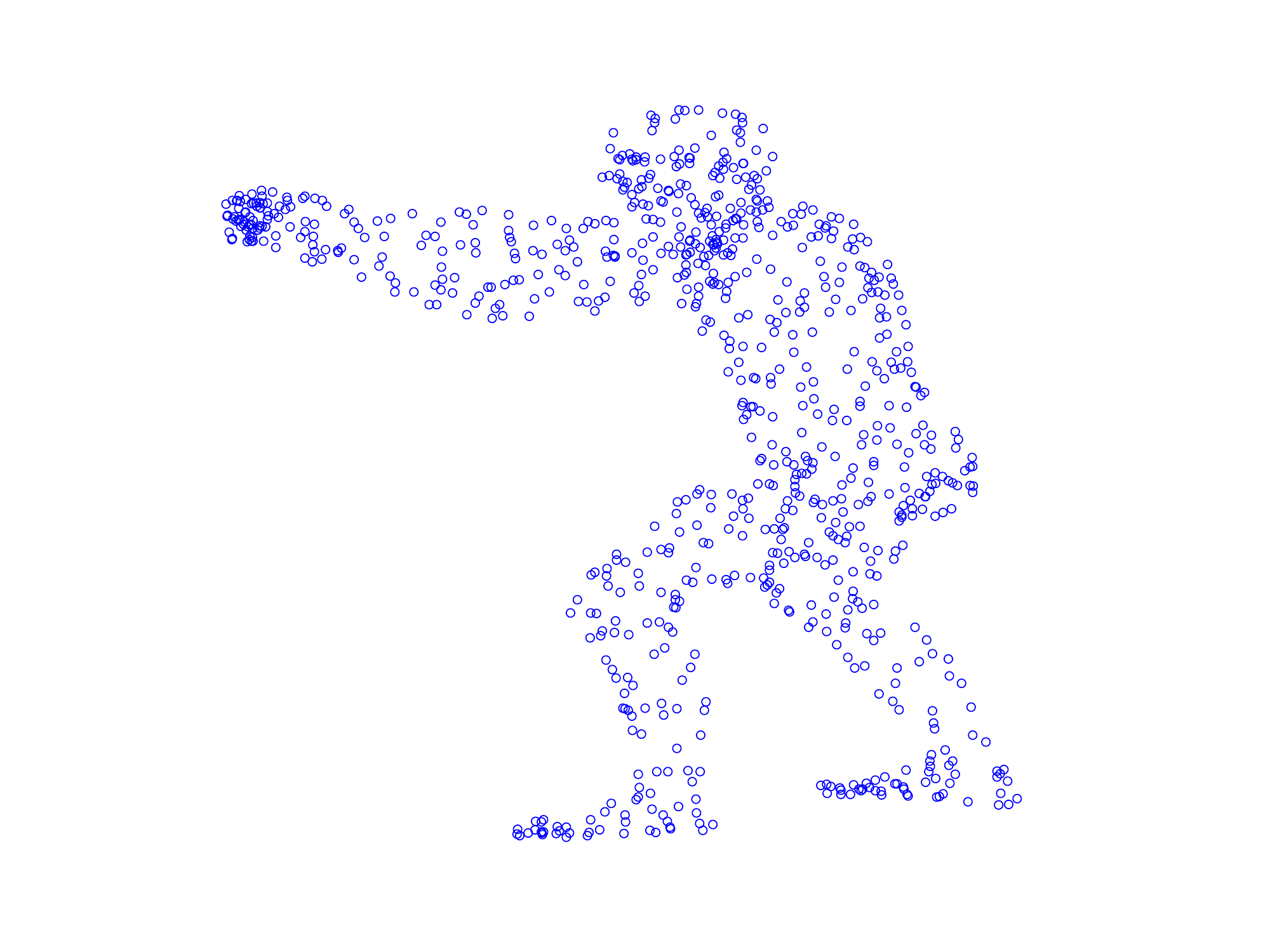} & \includegraphics[scale=0.15]{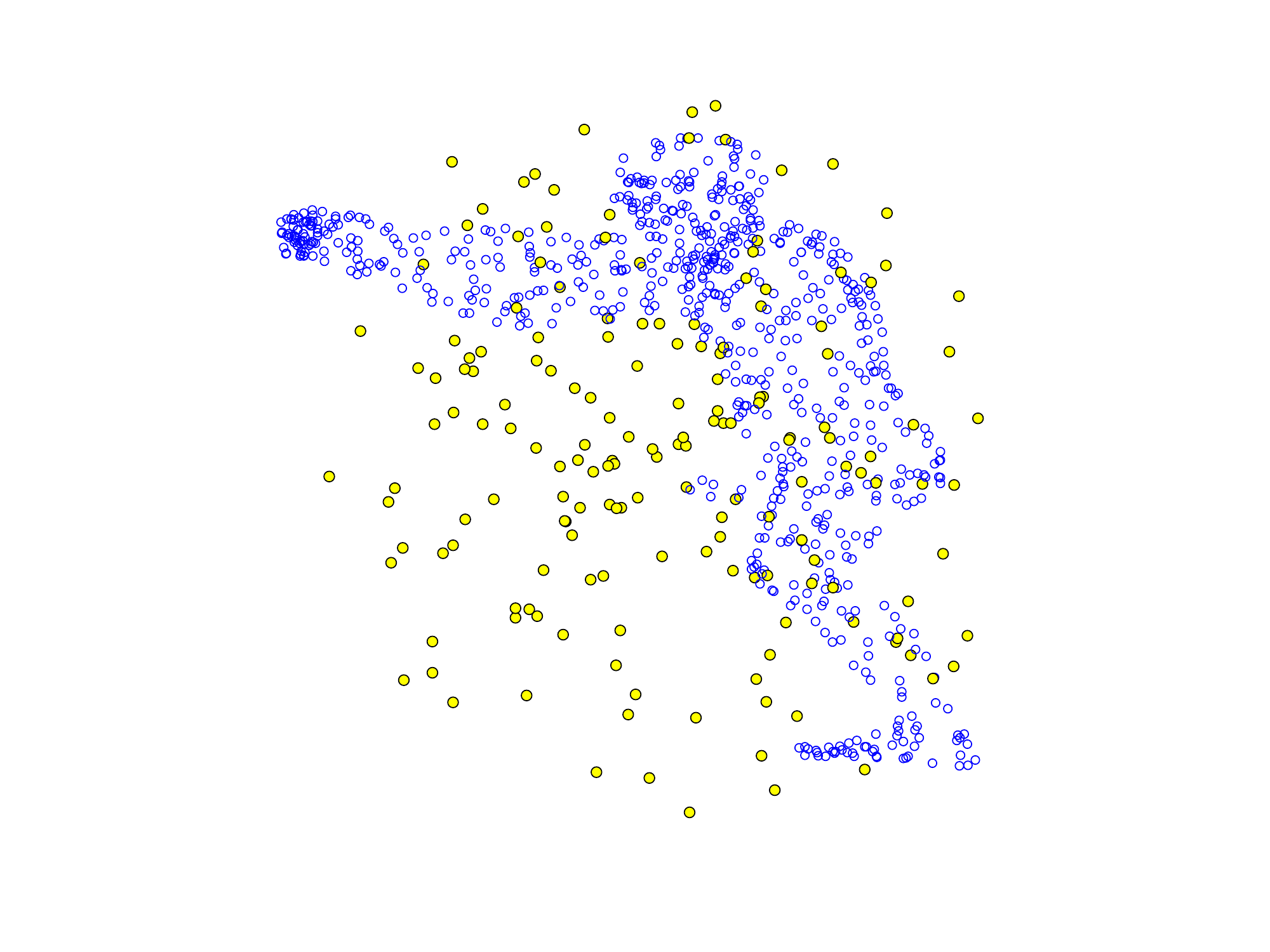} & \includegraphics[scale=0.15]{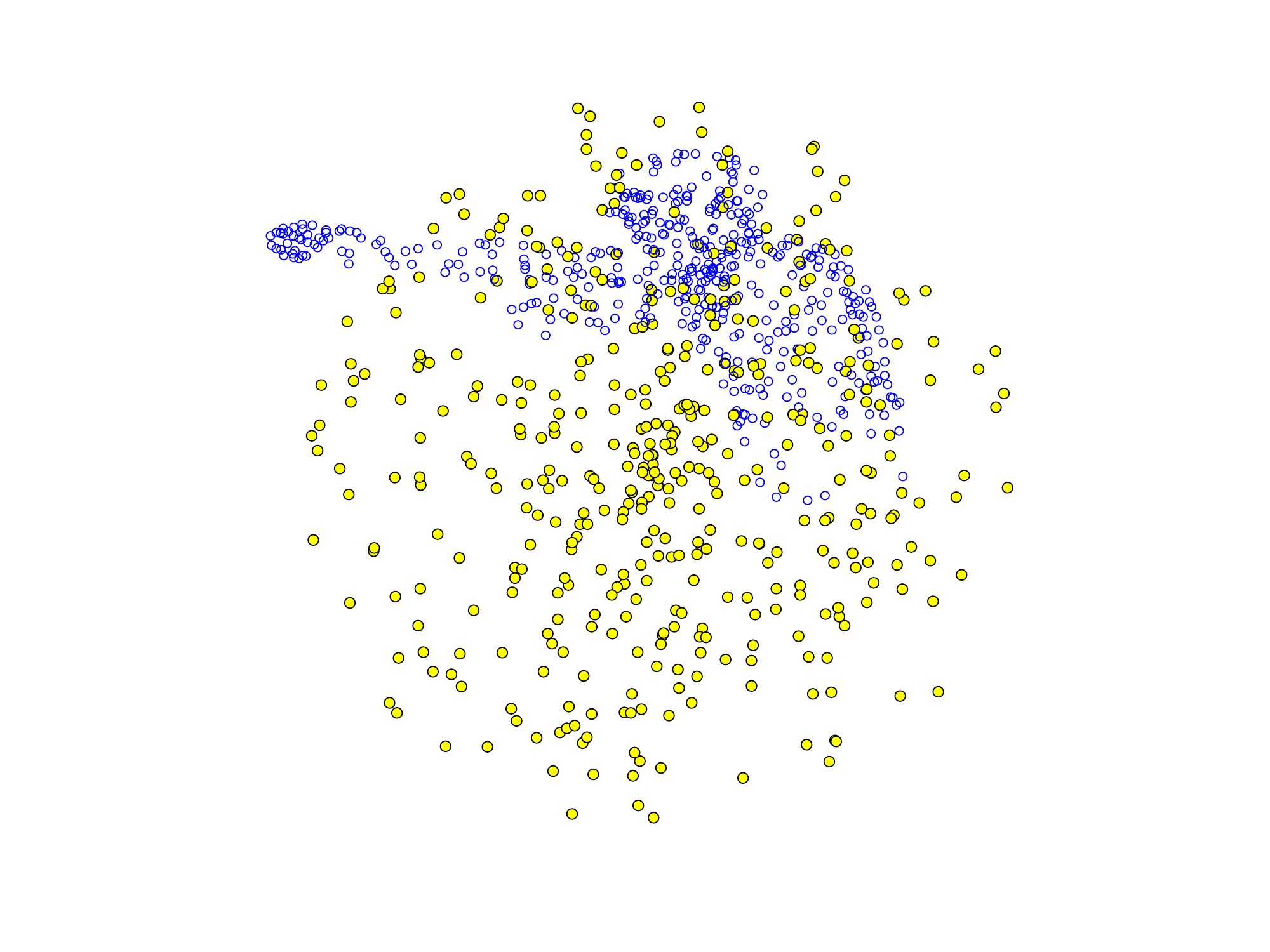}\vspace{-10pt}\tabularnewline
{\footnotesize{}MPO = 0} & {\footnotesize{}MPO = 0.2} & {\footnotesize{}MPO = 0.5}\tabularnewline
\end{tabular}\protect\caption{Examples of 2D and 3D shapes with increasing values of missing point/outliers
(mpo). The MPO values used in the experiment ranged from 0 to 0.5.\label{fig: MPO examples}}
\end{figure}

The number of points in the shapes collected varied between 250 to
10,000 points. This version of the GrassGraph approach focused
on equal point-set cardinalities between the source and target shapes.
To obtain equal numbers of points, all of the shapes
were clustered using the k-means algorithm (using the desired number clusters) where
the closest point from the shape to a cluster center was used as the
new point. Given the base shape, we now explain how to generate the
various noise and MPO artifacts on the shapes.

As mentioned previously, the two standard artifacts that we applied
to the clustered shapes were noise and missing points with outliers (MPO).
The process of adding noise and MPO artifacts will be referred to
as ``noise protocol'' and ``MPO protocol'', respectively. To generate
noise in 2D, we uniformly sampled a new point from a circle of radius
$r$ around each point. The uniformly sampled point replaced the original
point (center of circle) in the shape---the larger the radius of the
circle the more noisy the shape became.\textbf{ }The same principle
applied to 3D shapes, now instead of a circle, we uniformly sampled
from a sphere or radius $r$. To generate a noisy shape for experiments,
an affine shape was first created, the points are randomly shuffled to remove
the correspondence and then the noise protocol is applied.

In the GrassGraph framework, we combine the missing points and outliers
into a single artifact. The number of points removed corresponding
to the missing point (MP) percentage ($0\leq mp\leq0.5$) became the
same number of points added as outliers. To create outliers, we uniformly
draw samples from a circle in 2D and a sphere in 3D of radius $r$
(independent of $r$ for noise generation). To get $r$, the max spread
of the points across all coordinate directions was divided by two
and multiplied by 1.2. This ensured that the circle or sphere fully
encompassed the shape volume, leading to outliers as shown in Figure
\ref{fig: MPO examples}. To generate shapes with MPO for experiments,
the artifact was applied to the source shape followed by affine transformation application.

\begin{table}[t]
\centering{}{\scriptsize{}}%
\begin{tabular}{|c|c|c|c|c|c|c|c|c|}
\hline 
\textbf{\tiny{}Exp} & \textbf{\tiny{}Dim} & \textbf{\tiny{}\#Sh} & \textbf{\tiny{}\#NL} & \textbf{\tiny{}\#Cases} & \textbf{\tiny{}\#Meth} & \textbf{\tiny{}\#Aff} & \textbf{\tiny{}\#MPO} & \textbf{\tiny{}\#Trials}\tabularnewline
\hline 
\multirow{2}{*}{\textbf{\tiny{}N\hspace{-5pt}}} & \multirow{1}{*}{{\scriptsize{}2D}} & {\scriptsize{}20} & {\scriptsize{}11} & {\scriptsize{}5} & {\scriptsize{}4} & {\scriptsize{}30} & {\scriptsize{}-} & {\scriptsize{}132,000}\tabularnewline
\cline{2-9} 
 & {\scriptsize{}3D} & {\scriptsize{}20} & {\scriptsize{}11} & {\scriptsize{}5} & {\scriptsize{}3} & {\scriptsize{}30} & {\scriptsize{}-} & {\scriptsize{}99,000}\tabularnewline
\hline 
\multirow{2}{*}{\textbf{\tiny{}MPO\hspace{-5pt}}} & \multirow{1}{*}{{\scriptsize{}2D}} & {\scriptsize{}20} & {\scriptsize{}-} & {\scriptsize{}5} & {\scriptsize{}4} & {\scriptsize{}30} & {\scriptsize{}10} & {\scriptsize{}120,000}\tabularnewline
\cline{2-9} 
 & {\scriptsize{}3D} & {\scriptsize{}20} & {\scriptsize{}-} & {\scriptsize{}5} & {\scriptsize{}3} & {\scriptsize{}30} & {\scriptsize{}10} & {\scriptsize{}90,000}\tabularnewline
\hline 
\multicolumn{4}{|c||}{{\scriptsize{}Total Number of Trials }} & \multicolumn{4}{c||}{} & \textbf{\scriptsize{}441,000}\tabularnewline
\hline 
\end{tabular}{\scriptsize{}\vspace{2pt}}\protect\caption{Breakdown of 441,000 experiments used to benchmark the GrassGraph
framework. N: Noise; MPO: Missing points with outliers; Dim: Shape dimension;
\#Sh: No. of base shapes; \#NL: No. of noise levels;
\#Meth: No. of methods; \#Aff: No. of affine transformations; \#MPO: No. of MPO levels.\label{tab:Breakdown-of-experiements}}
\end{table}

To generate affine transformations for 2D and 3D we follow a similar approach
taken in \cite{Zhang04}. Each transformation is decomposed into a rotation,
scale and rotation (in that order) followed by a translation. In 2D, the
translation parameters used for the experiments were $\{-30 \leq
t_{x},t_{y}\leq 30\}$, where $t_{x}$ is the translation in the $x$ direction
and $t_{y}$ is the translation in the $y$ direction. The rotation and scale
parameters were varied in the following ranges:
${-\pi/2\leq\theta,\sigma\leq\pi/2}$ ,and ${-3\leq s_{x},s_{y}\leq3}$.  Here
$\sigma$ is angle of the first 2D rotation matrix, followed by the scaling
matrix with $x-$ and $y-$direction parameters $s_{x}$ and $s_{y}$,
respectively. The 3D affine transformation matrix is similarly decomposed.

\begin{figure}[t]
\begin{centering}
\begin{tabular}{ccc}
  \hspace{-8pt}\includegraphics[width=3.5cm,height=2.8cm]
  {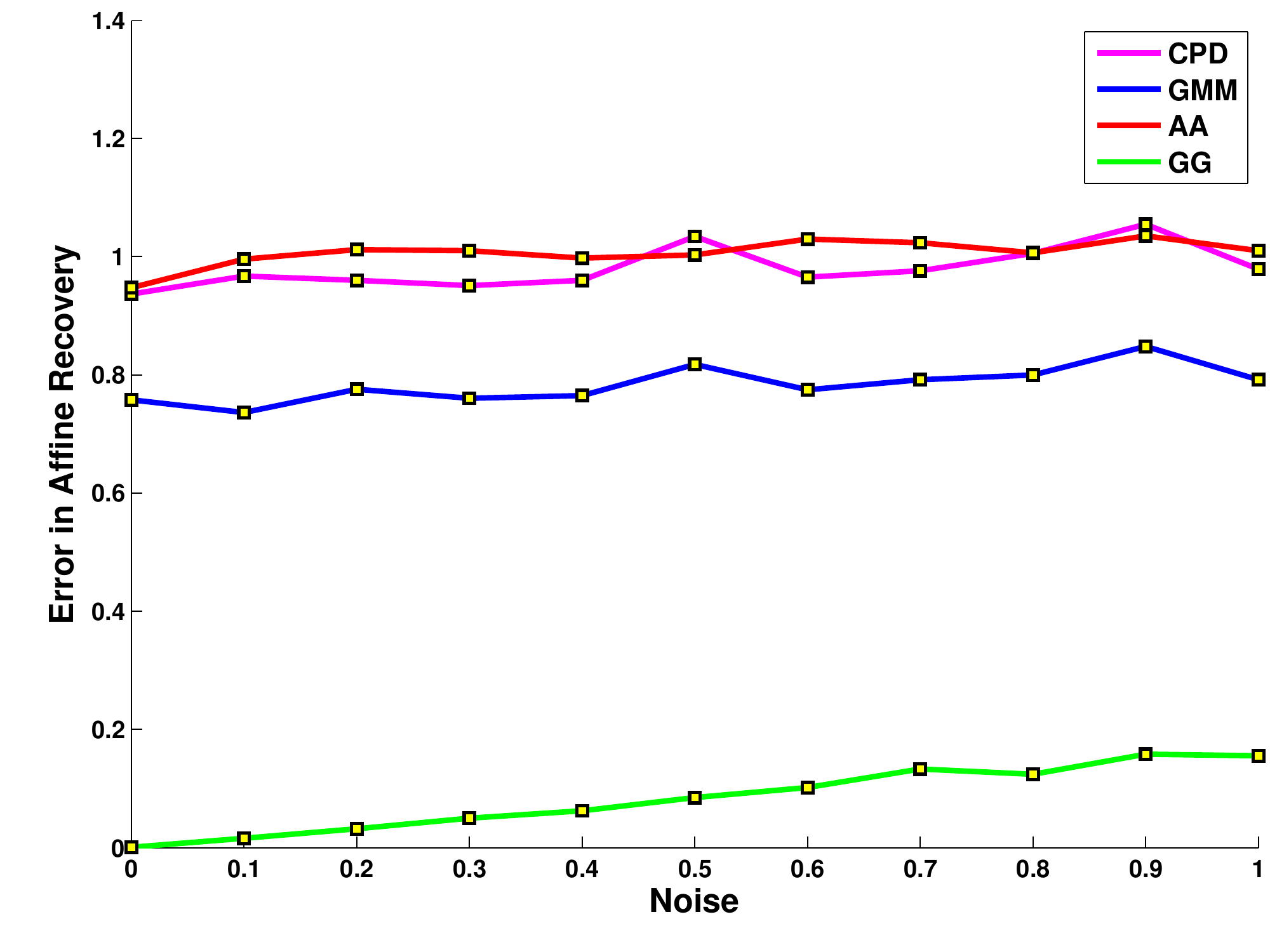} &
  \hspace{-12pt}
	\includegraphics[width=3.5cm,height=2.8cm]{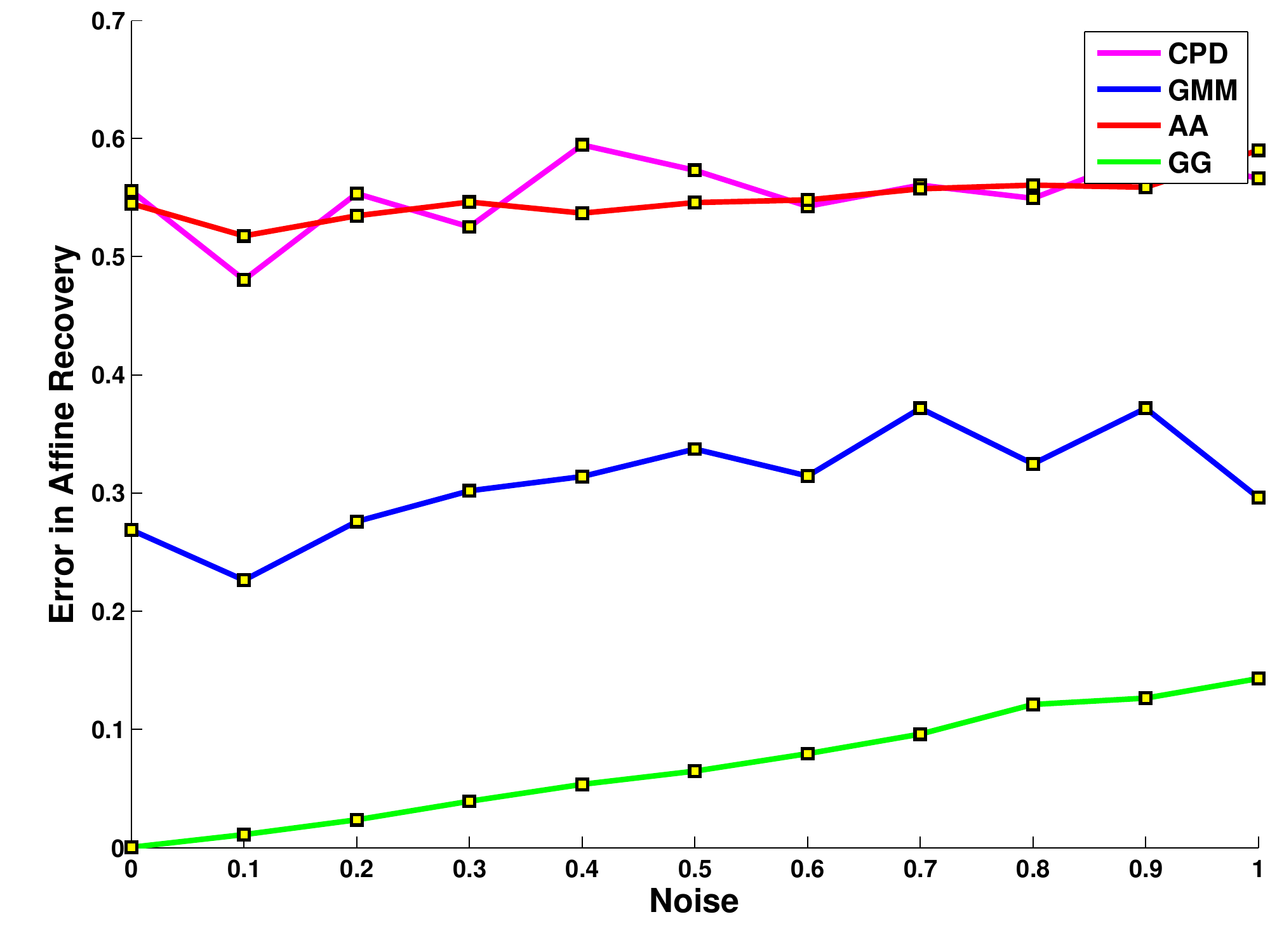}\vspace{-5pt}
  &
  {\hspace{-12pt}
	\includegraphics[width=3.5cm,height=2.8cm]{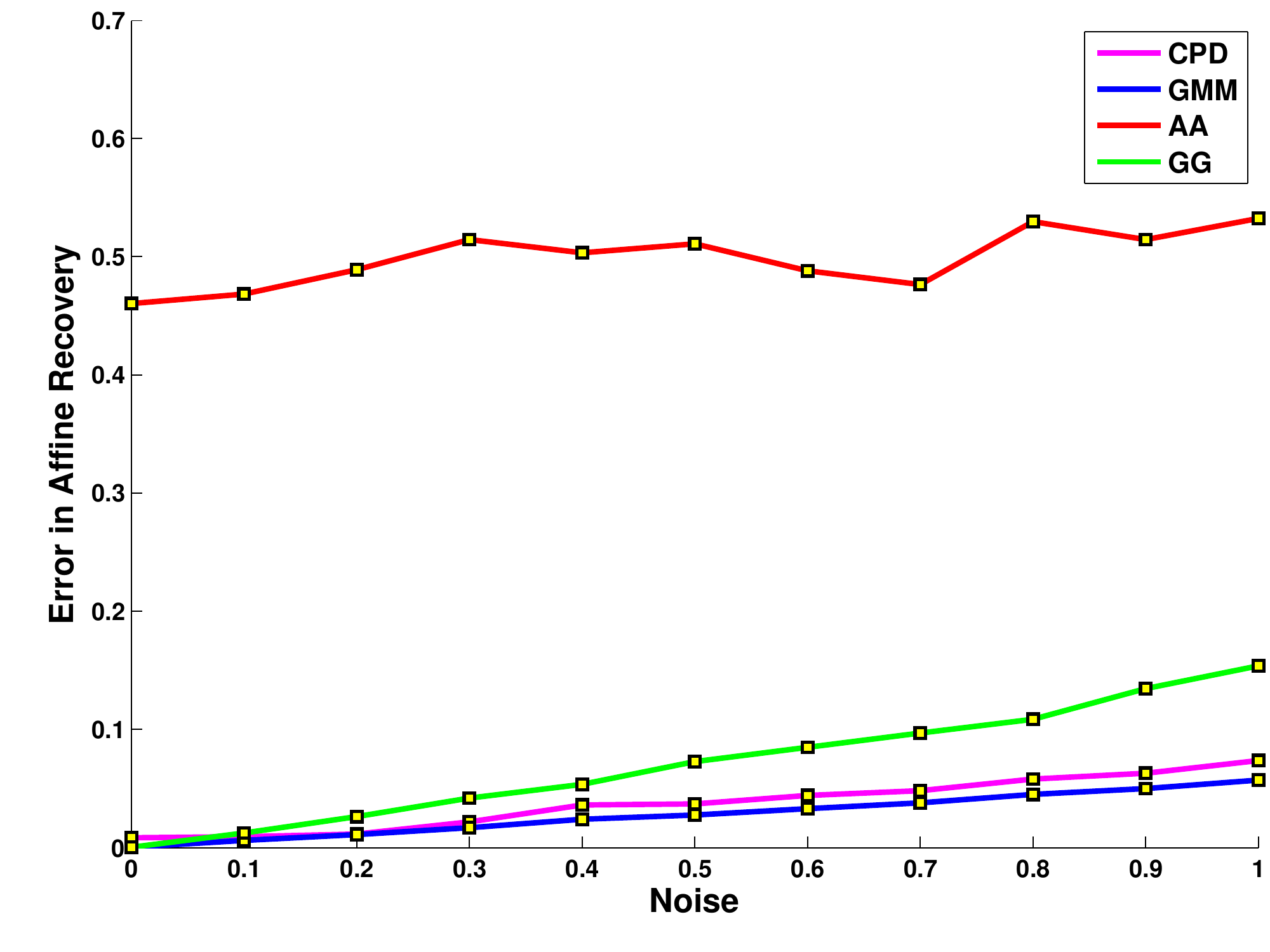}}\tabularnewline 
  {\footnotesize{}(a) Big-Ang, Big-Sc, No-Trans} &
  \hspace{-12pt}{\footnotesize{}(b) Big-Ang, Sm-Sc, Sm-Trans} & {\footnotesize{}e) Sm-Ang,
    Sm-Sc, Sm-Trans}\tabularnewline 
  \hspace{-8pt}
	\includegraphics[width=3.5cm,height=2.8cm]{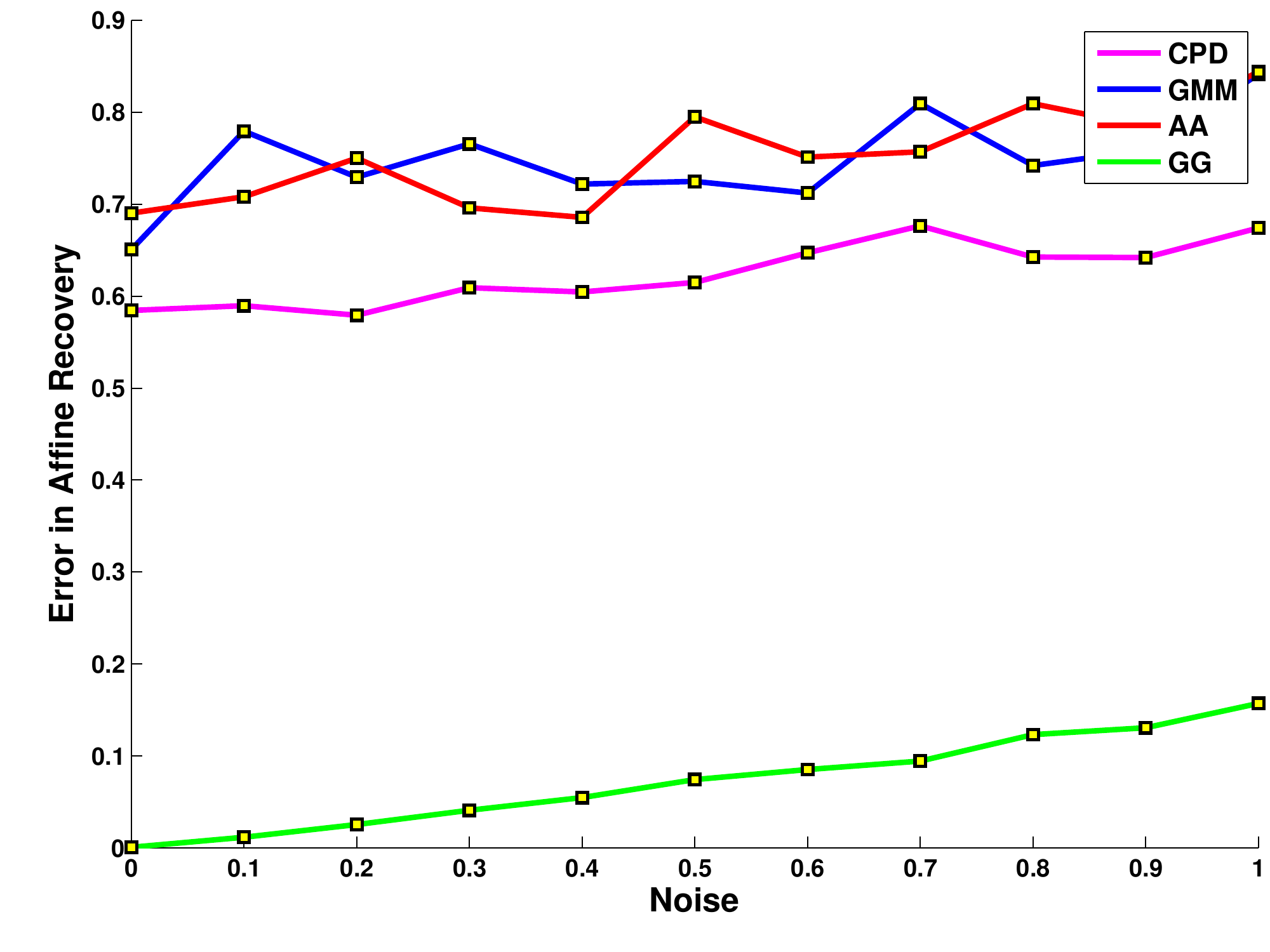}
  &
  
	\includegraphics[width=3.5cm,height=2.8cm]{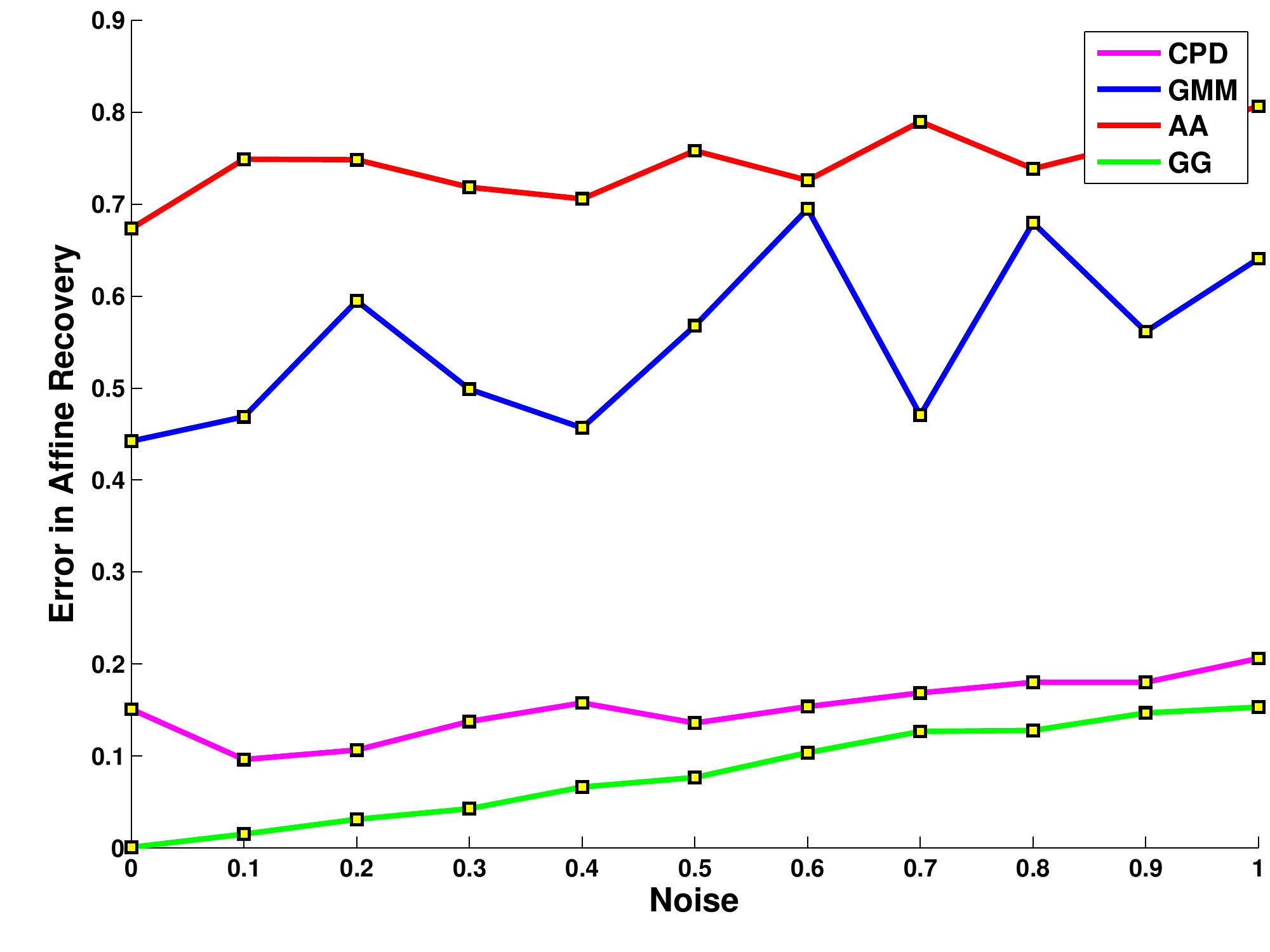}\vspace{-5pt}
  & \hspace{-12pt}\tabularnewline 
  \hspace{-6pt}{\footnotesize{}(c) Big-Ang, Med-Sc, Sm-Trans} &
  \hspace{-12pt}{\footnotesize{}(d) Med-Ang, Sm-Sc, Sm-Trans} & \tabularnewline 
\end{tabular}
\par\end{centering}

\protect\caption{Noise 2D. We outperform the competitors in (a)-(d) as shown above
by the green curves and are slightly outperformed in (e) by CPD and
GMM. Due to the affine invariance of the proposed GrassGraph (GG) approach, our error curves
are consistent across all cases versus the other methods that fluctuate
depending on the size of the affine transformation. \label{fig: Noise-2D}}
\end{figure}

\begin{table}
\centering{}%
\begin{tabular}{|c|c|c|c|}
\cline{2-4} 
\multicolumn{1}{c|}{} & {\footnotesize{}Angle} & {\footnotesize{}Scale} & {\footnotesize{}Translation}\tabularnewline
\hline 
{\footnotesize{}Small} & {\footnotesize{}$\pi/6:\pi/6$} & {\footnotesize{}$-1:1$} & {\footnotesize{}$-10:10$}\tabularnewline
\hline 
{\footnotesize{}Medium } & {\footnotesize{}$\pi/4:\pi/4$} & {\footnotesize{}$-2:2$} & {\footnotesize{}$-20:20$}\tabularnewline
\hline 
{\footnotesize{}Large} & {\footnotesize{}$\pi/2:\pi/2$} & {\footnotesize{}$-3:3$} & {\footnotesize{}$-30:30$}\tabularnewline
\hline 
\end{tabular}\vspace{2pt}\protect\caption{Experimental parameters used to simulate different sizes of affine
transformations.\label{tab: Affine parameter classification}}
\end{table}

The translations potentially ranged from $\{-30\leq t_{x},t_{y},t_{z}\leq 30\}$, depending on the severity the transformation.  The bounds for the three different rotation angles and scale parameters were ${-\pi/2\leq\theta_{x},\theta_{y},\theta_{z},\sigma_{x},\sigma_{y},\sigma_{z}\leq\pi/2}$
and ${-3\leq s_{x},s_{y},s_{z}\leq3}$. For the experiments,
the bounds on the angles, scaling values and translation values were
varied to produce different sizes of affine transformations. Table
\ref{tab: Affine parameter classification} shows the classification
of the affine parameters used to simulate the transformations for
the experiments.

\begin{figure}[t]
\begin{centering}
\begin{tabular}{ccc}
\hspace{-8pt}\includegraphics[width=3.5cm,height=2.8cm]{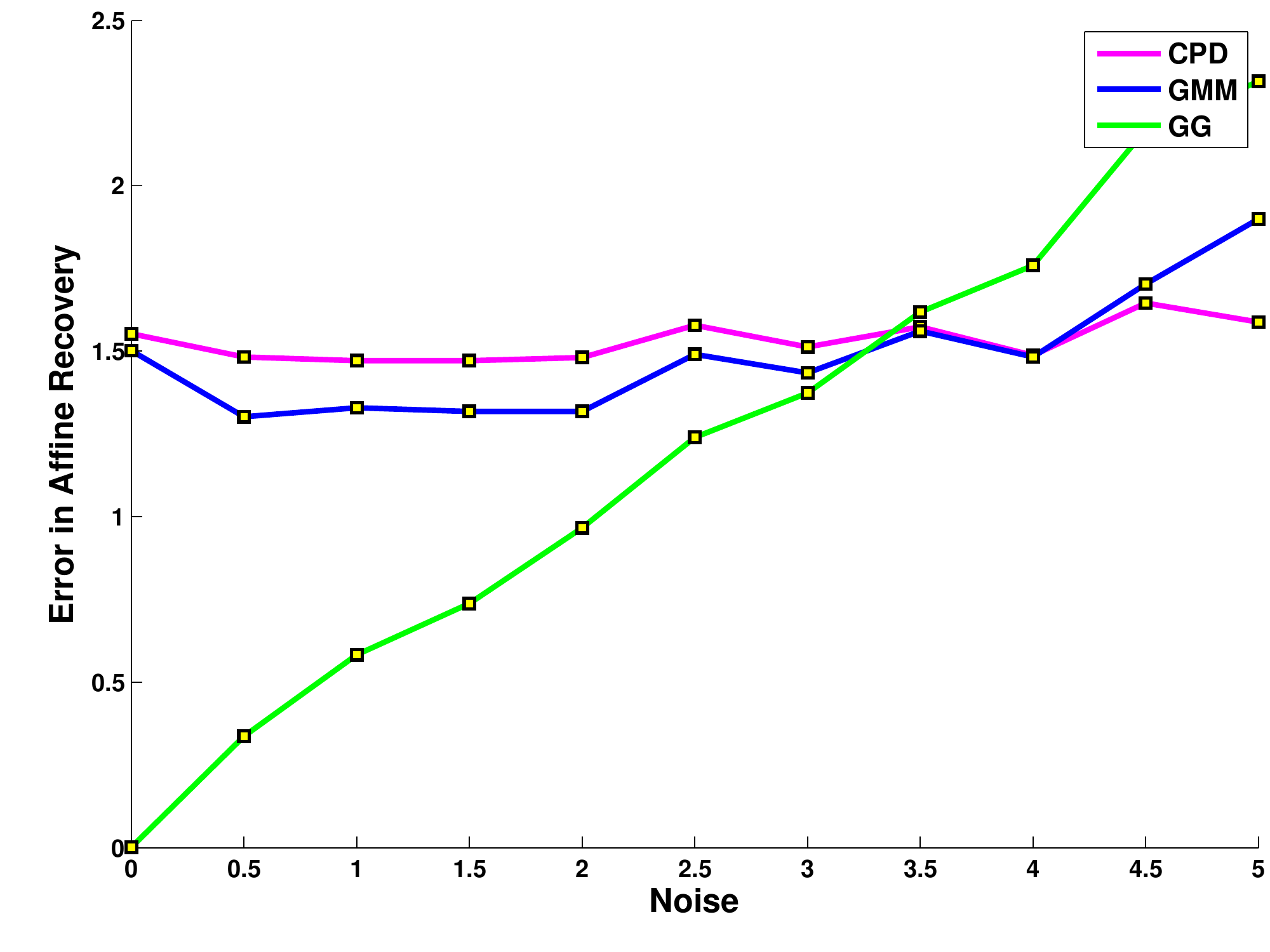} & 
\hspace{-12pt}\includegraphics[width=3.5cm,height=2.8cm]{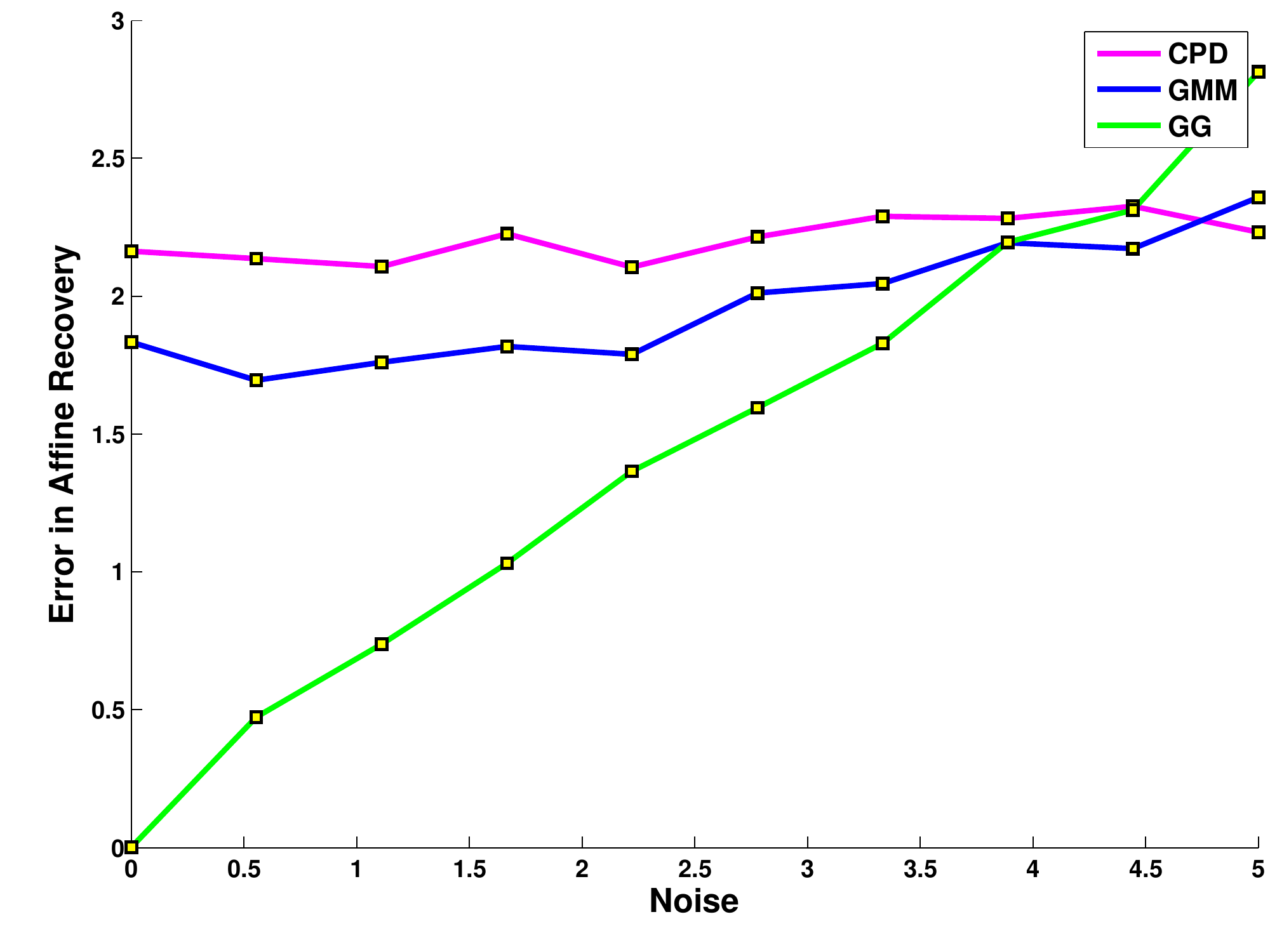}\vspace{-5pt} & 
{\hspace{-12pt}\includegraphics[width=3.5cm,height=2.8cm]{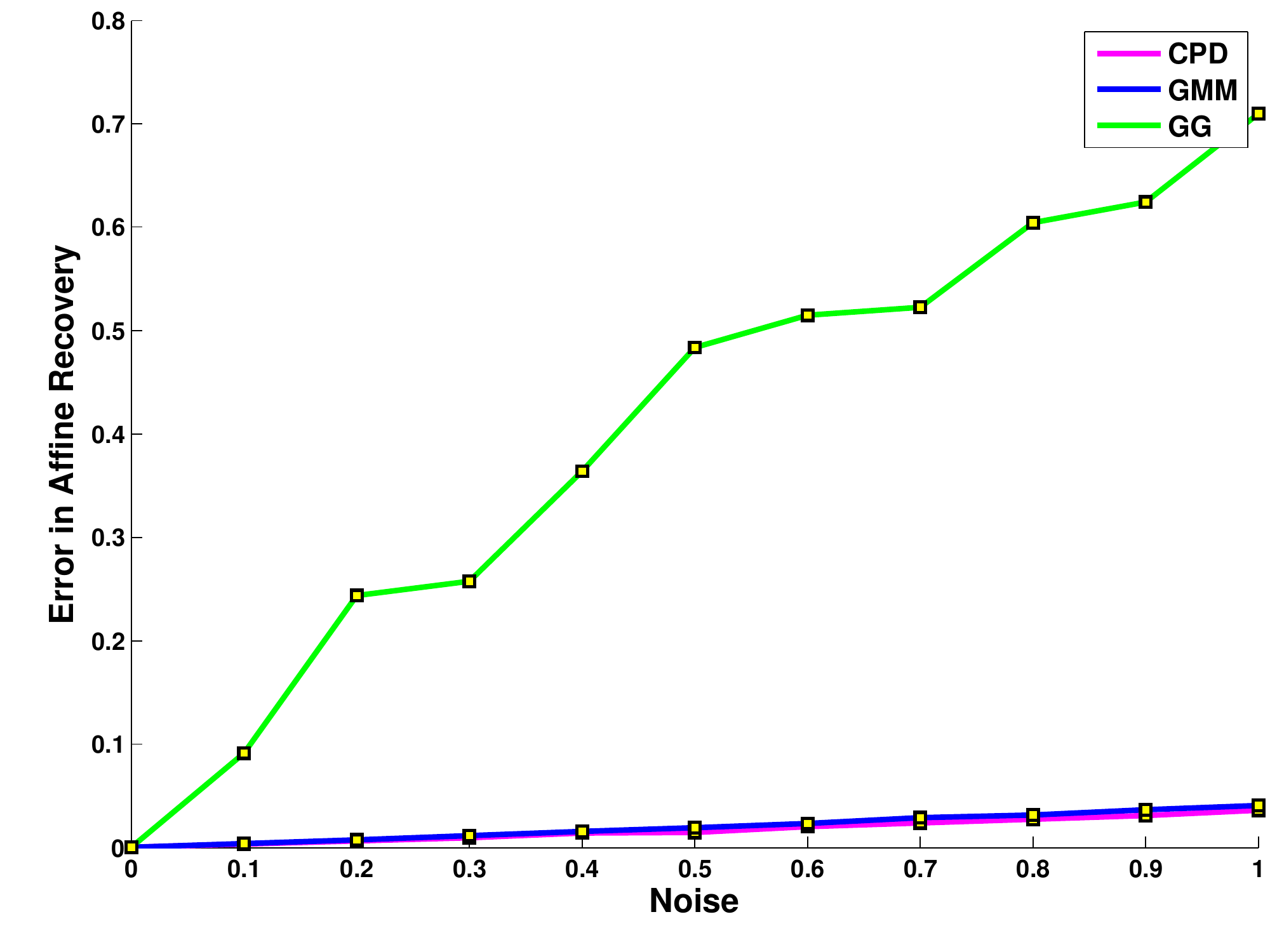}}\tabularnewline
\hspace{-6pt}{\footnotesize{}(a) Med-Ang, Sm-Sc, No-Trans} & {\footnotesize{}(b) Sm-Ang, Med-Sc, No-Trans} & {\footnotesize{}(e) Sm-Ang, Sm-Sc, No-Trans}\tabularnewline
\hspace{-8pt}\includegraphics[width=3.5cm,height=2.8cm]{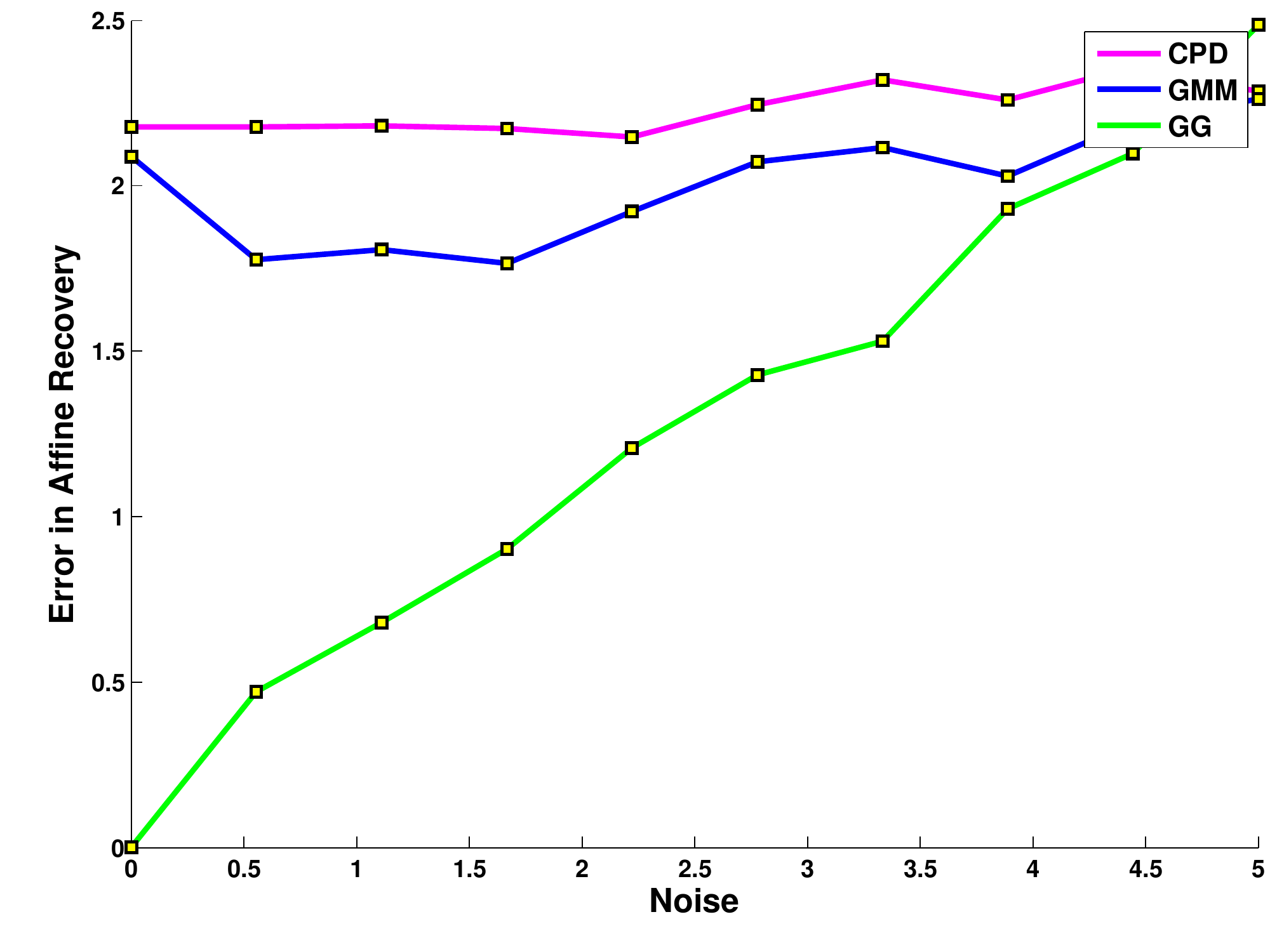} & \hspace{-12pt}\includegraphics[width=3.5cm,height=2.8cm]{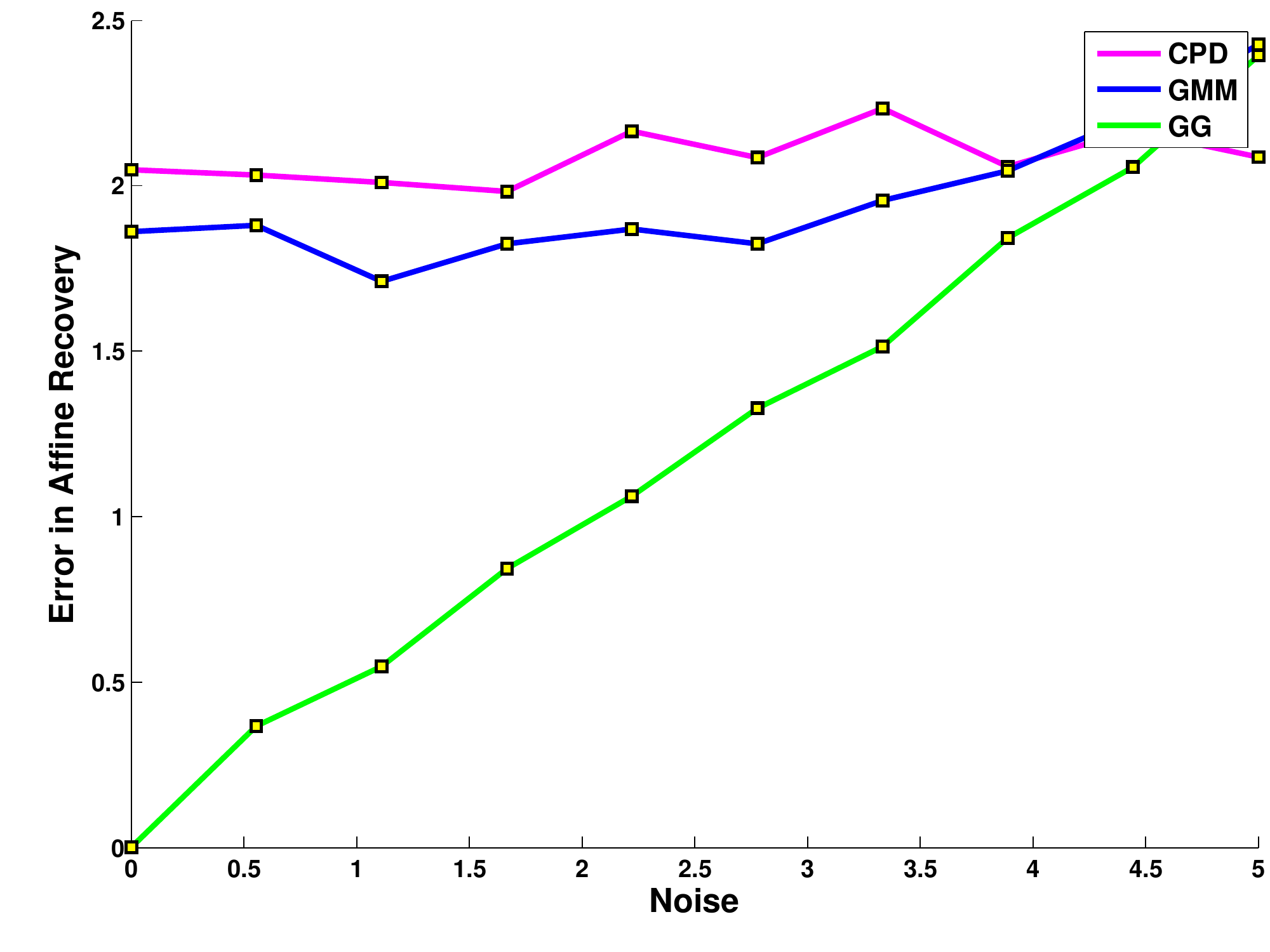}\vspace{-5pt} & \hspace{-12pt}\tabularnewline
\hspace{-6pt}{\footnotesize{}(c) Med-Ang, Med-Sc, No-Trans} & {\footnotesize{}(d) Big-Ang, Sm-Sc, No-Trans} & \tabularnewline
\end{tabular}
\par\end{centering}

\protect\caption{Noise 3D. We outperform the competing methods in (a)-(d) at noise
levels below 3.5 in (a), 4 in (b), 4.6 in (c) and 4.7 in (d). We trail in (e)
in the case of small angles and scale without translation.
However, as the sizes of the affines increase, the competing
methods are struggle to recover the affine transformation when noise
is present whereas our invariant method performs well. \label{fig: Noise 3D}}
\end{figure}

To perform the experiments, each base shape was transformed with 30 different
transformations to form affine shapes. The noise protocol was applied to each of these
affine shapes for the eleven values of noise ranging from 0:1 for
2D and 0:5 for 3D. Figure \ref{fig: Registered Noisy shapes} provides
examples of noisy shapes in 2D and 3D. All of the affine shapes are
evaluated at a single noise level for a particular method---the errors
across all the affine shapes are averaged and this error value is
assigned to the noise level for that method. For example in Figure
\ref{fig: Noise-2D}(a), each marker along a curve represents the
average error of the 600 (20 base shapes x 30 affines) affine shapes
evaluated at that noise level for the method corresponding to the
curve's color. The results of the noise experiments in 2D and 3D are
shown in Figures \ref{fig: Noise-2D} and \ref{fig: Noise 3D}, respectively.
For 2D, across the eleven noise levels with the four methods (GrassGraph
and three competitors) we get a total of 132,000 experimental trials
and in 3D for the three methods we ran a total of 99,000 trials. 

For the 2D noise experiments, the GrassGraph (green) method performed
the best across (a)-(d). However, when there was no translation, small angles
and small scaling, our method was slightly outperformed by CPD and
GMM in (e). Once the affine parameters are increased however, the true
utility of our invariant method was highlighted as we outperformed
the competing methods. CPD seems to be more susceptible to larger
angles and scale whereas GMM is affected more by translations. In
case (a), for large affines we see that the competing methods have
almost flat curves, this suggests that the correspondences retrieved
across all the noise levels are relatively the same, which means that
the methods themselves were failing. The Algebraic Affine (AA) method
seems to perform the worst across all of the cases, while CPD and
GMM fluctuate with the change in affine transformation. Our curve
increases gracefully with noise because we are invariant to the affine transformation.
We emphasize that only nearest neighbor correspondences were used on the affine invariant
coordinates whereas the competing methods
required numerical optimization.

\begin{figure}[t]
\begin{centering}
\begin{tabular}{ccc}
\hspace{-8pt}\includegraphics[width=3.5cm,height=2.8cm]{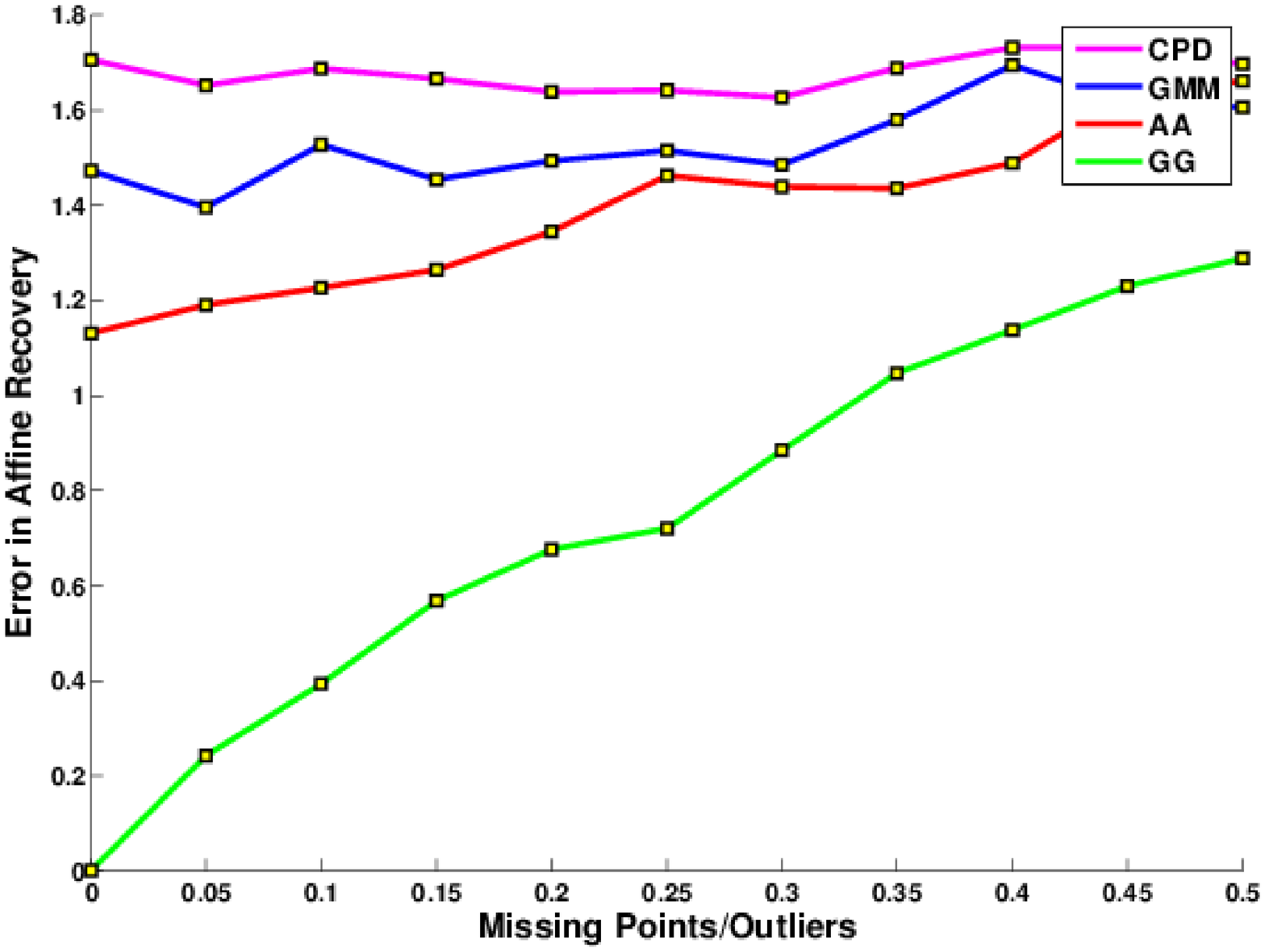} & 
\hspace{-12pt}\includegraphics[width=3.5cm,height=2.8cm]{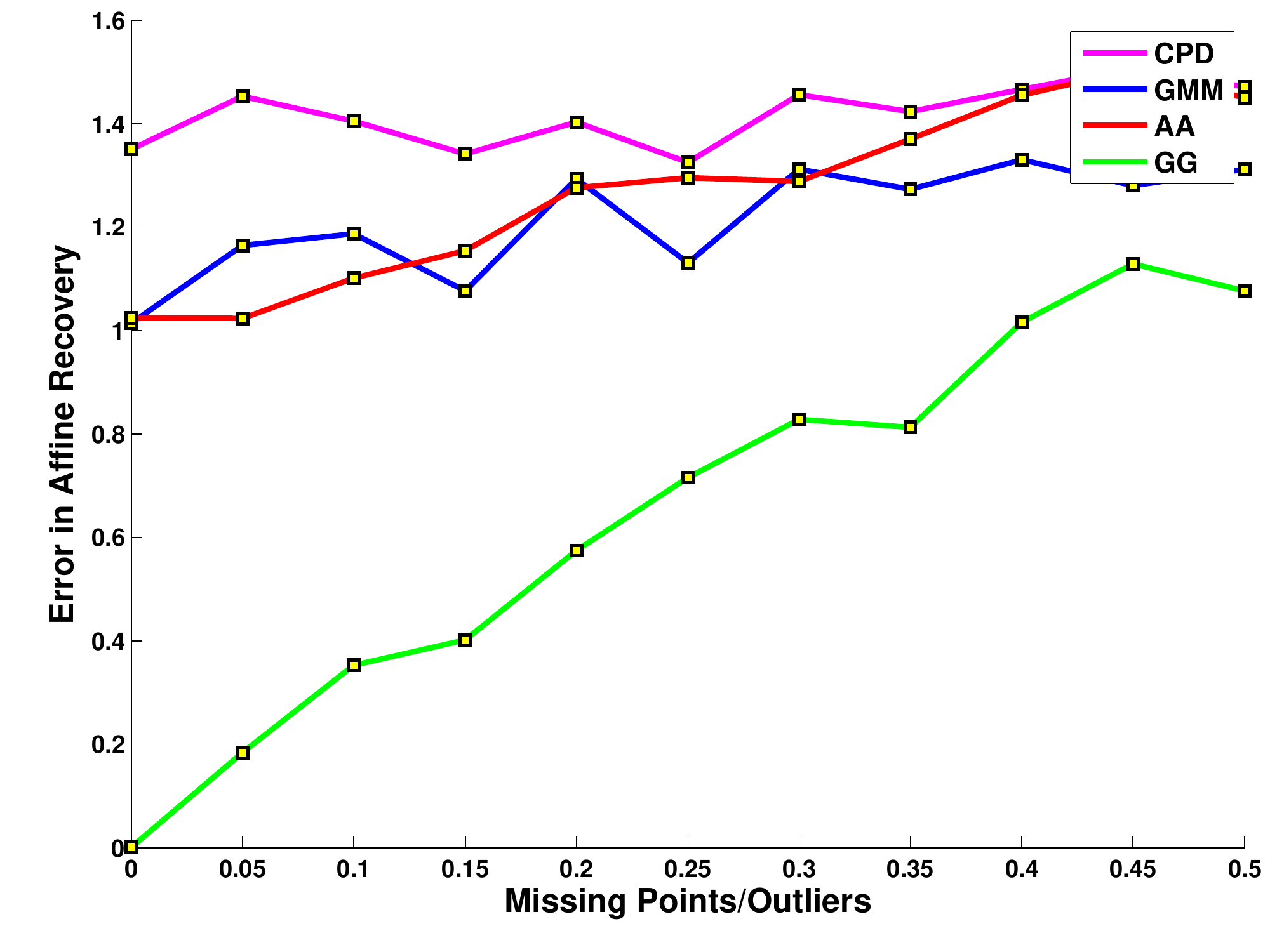}\vspace{-5pt} & 
{\hspace{-12pt}\includegraphics[width=3.5cm,height=2.8cm]{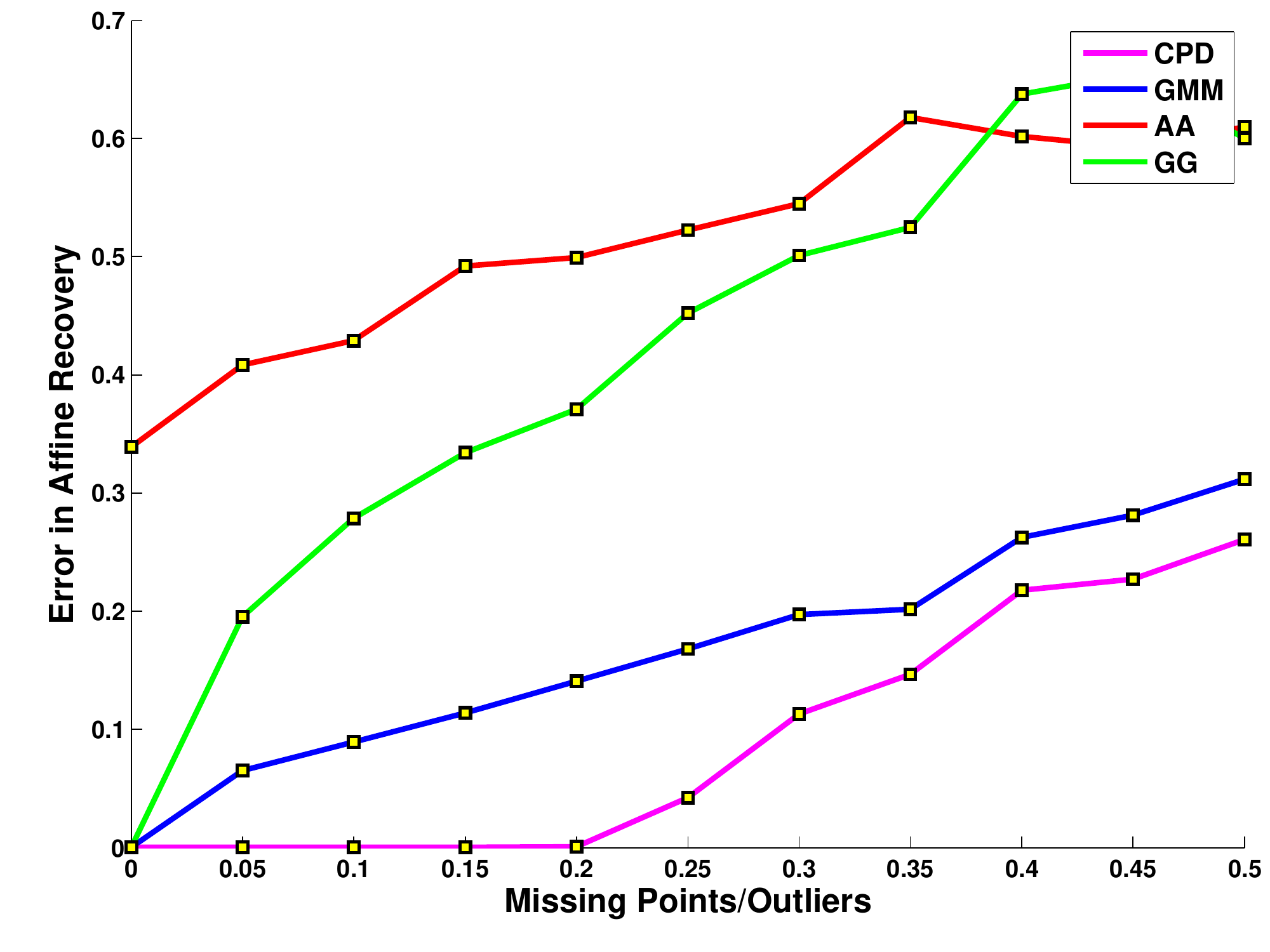}}\tabularnewline
\hspace{-6pt}{\footnotesize{}(a) Big-Ang, Med-Sc, Med-Trans} & 
{\footnotesize{}(b) Big-Ang, Med-Sc, Sm-Trans} & 
{\footnotesize{}(e) Sm-Ang, No-Sc, No-Trans}\tabularnewline
\hspace{-8pt}\includegraphics[width=3.5cm,height=2.8cm]{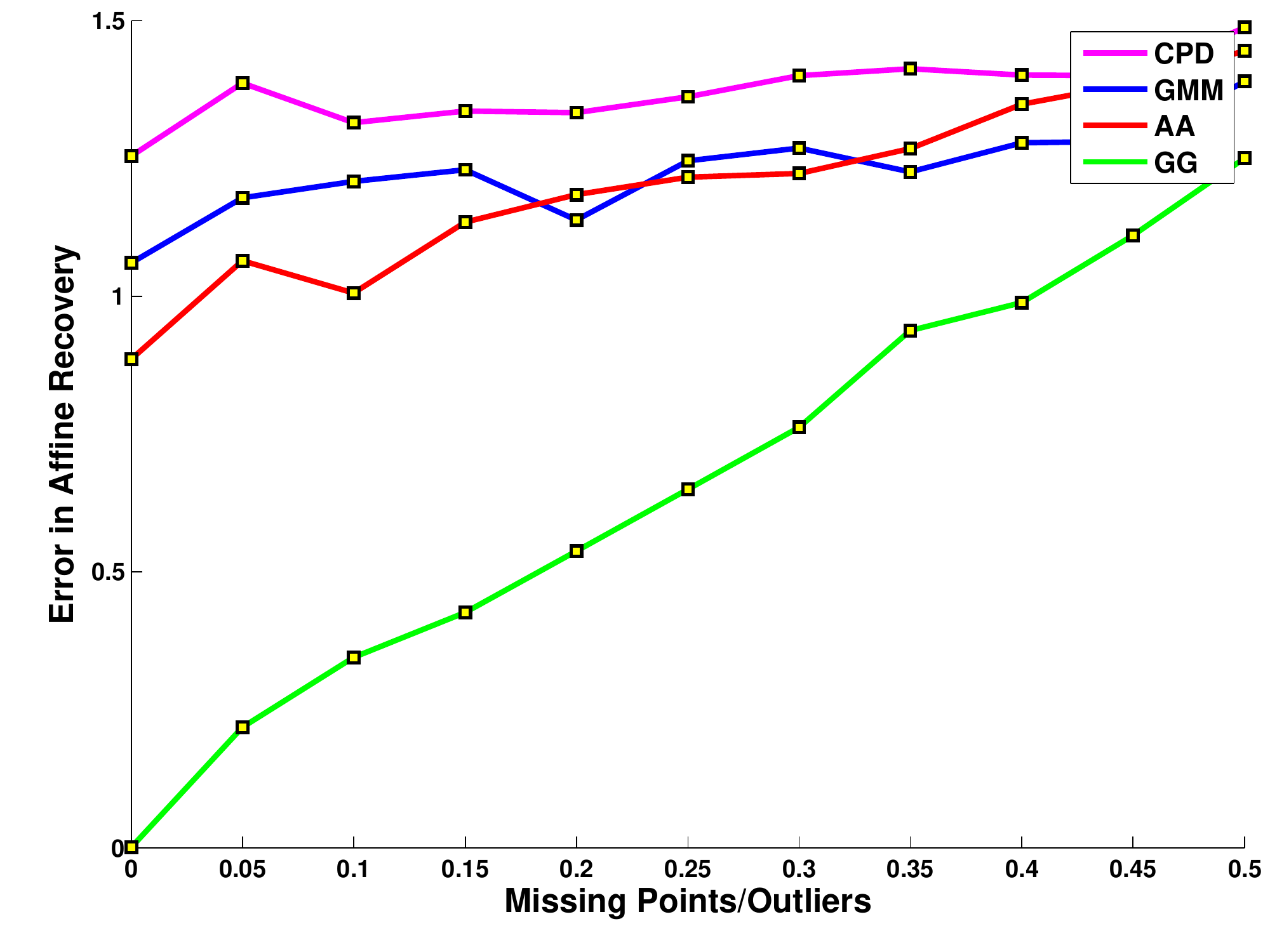} & 
\hspace{-12pt}\includegraphics[width=3.5cm,height=2.8cm]{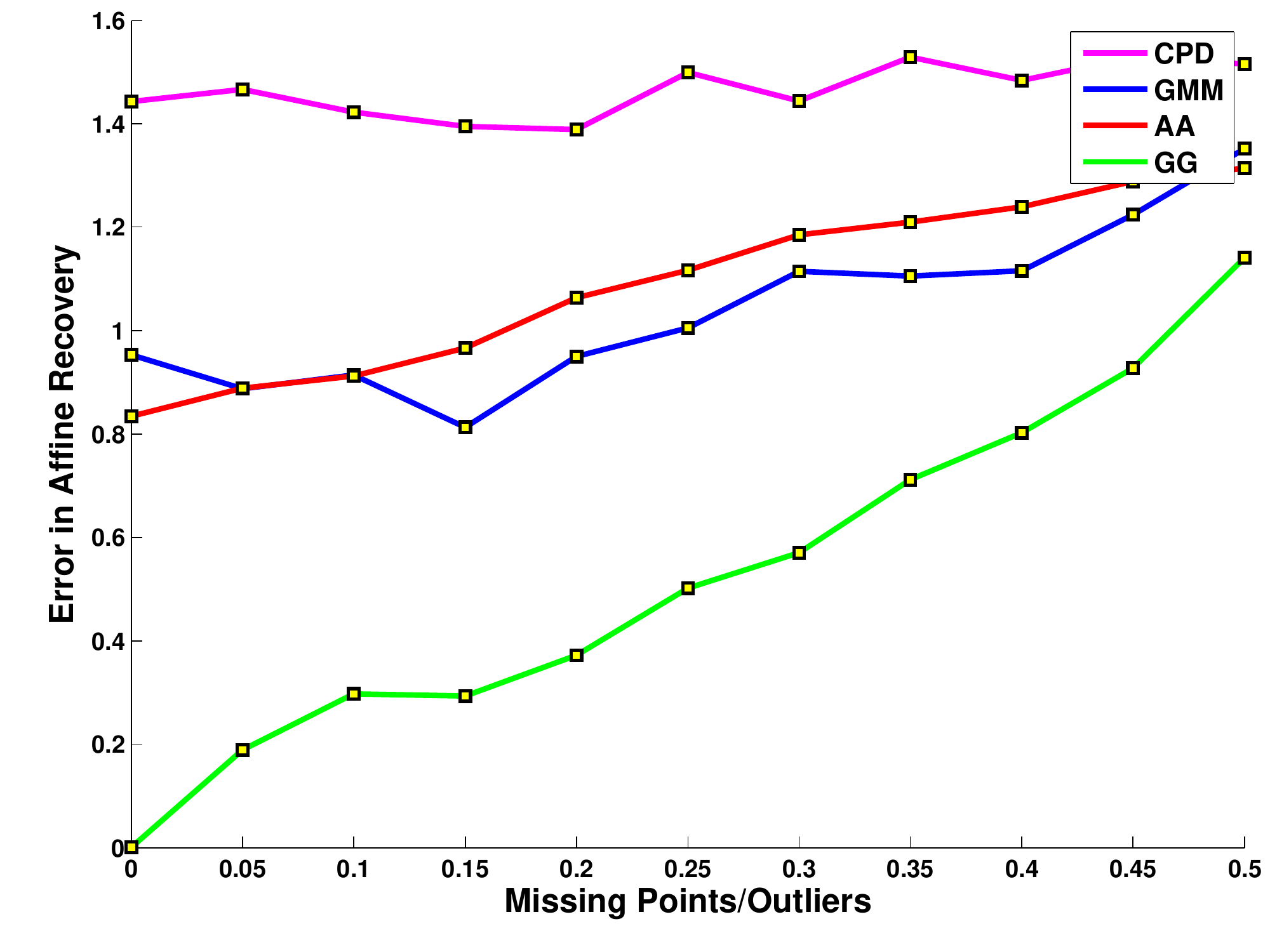}\vspace{-5pt} & 
\hspace{-12pt}\tabularnewline
\hspace{-6pt}{\footnotesize{}(c) Big-Ang, Big-Sc, No-Trans} & 
\hspace{-12pt}{\footnotesize{}(d) Big-Ang, Sm-Sc, No-Trans} & \tabularnewline
\end{tabular}
\par\end{centering}

\protect\caption{MPO 2D. In cases (a)-(d) the GrassGraph approach (green curve) outperforms
the competing methods. In case (e) CPD and GMM outperform us but the
affine transformation is only a rotation matrix. Once scale and translation
are included CPD performs the worst, GMM and AA perform differently
depending on the size of the angle and the translation.\label{fig: MPO 2D}}
\end{figure}

In Figure \ref{fig: Noise 3D}, we see that our method outperforms
the competing methods at lower noise levels except in case (e) where
the affine transformation was very small. The nearest neighbor approach
that we employ holds up decently for larger affine transformations
versus smaller ones. GMM performs better than CPD in all cases except
in (e). The error in our method increases with increasing noise levels
due to the affine invariance. This is not the case for the non-affine
invariant methods CPD and GMM, which do not fluctuate as much. As we
moved up one dimension from 2D to 3D our method still performs well
using our simple nearest-neighbor correspondence finding method. This shows that
the GrassGraph approach should indeed be the first method considered
when recovering correspondences under noisy affine conditions. Now
we assess our method's performance in the presence of missing points
and outliers.

\begin{figure}[t]
\begin{centering}

\par\end{centering}

\begin{centering}
\begin{tabular}{ccc}
\hspace{-8pt}\includegraphics[width=3.5cm,height=2.8cm]{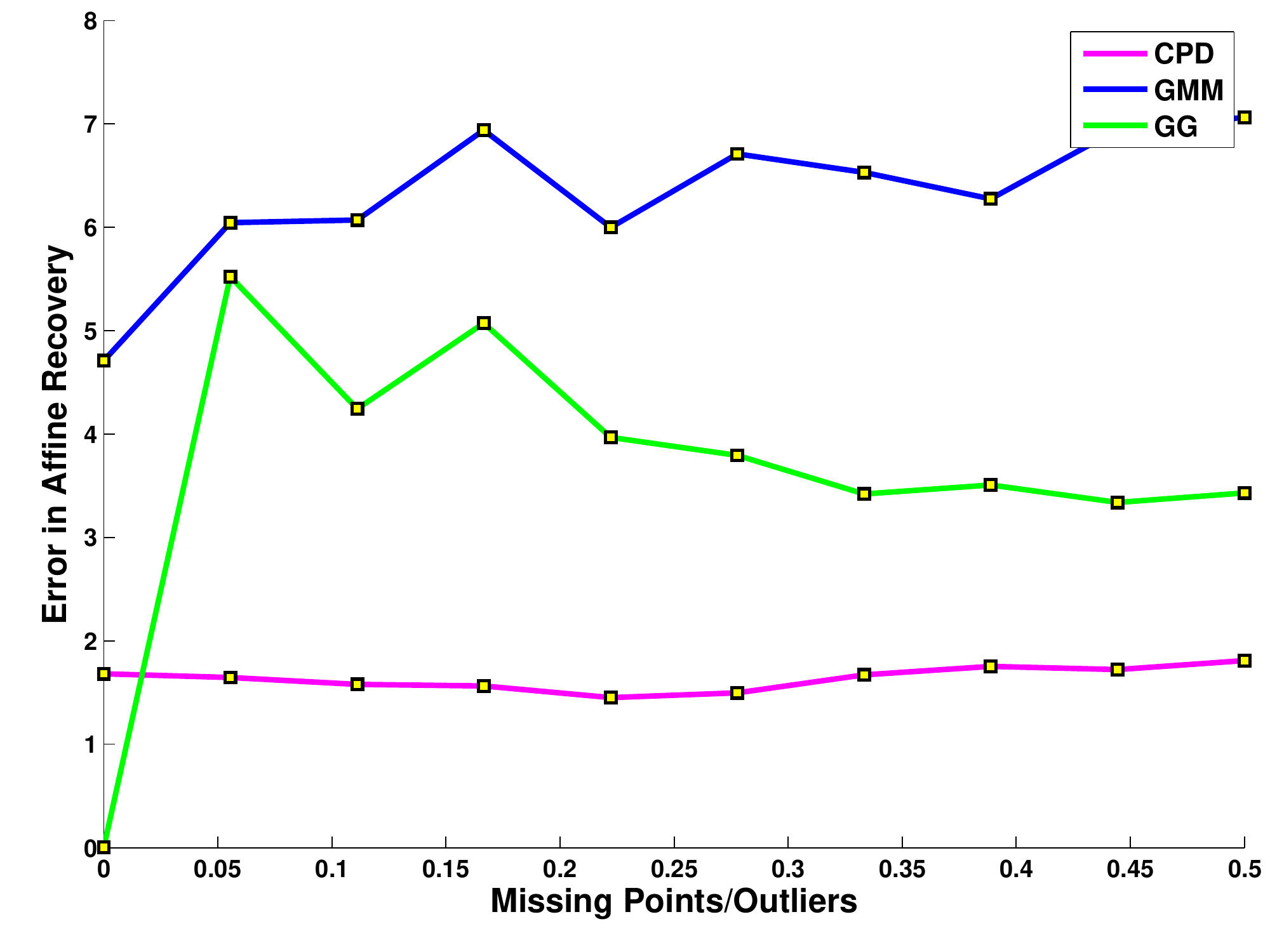} & 
\hspace{-12pt}\includegraphics[width=3.5cm,height=2.8cm]{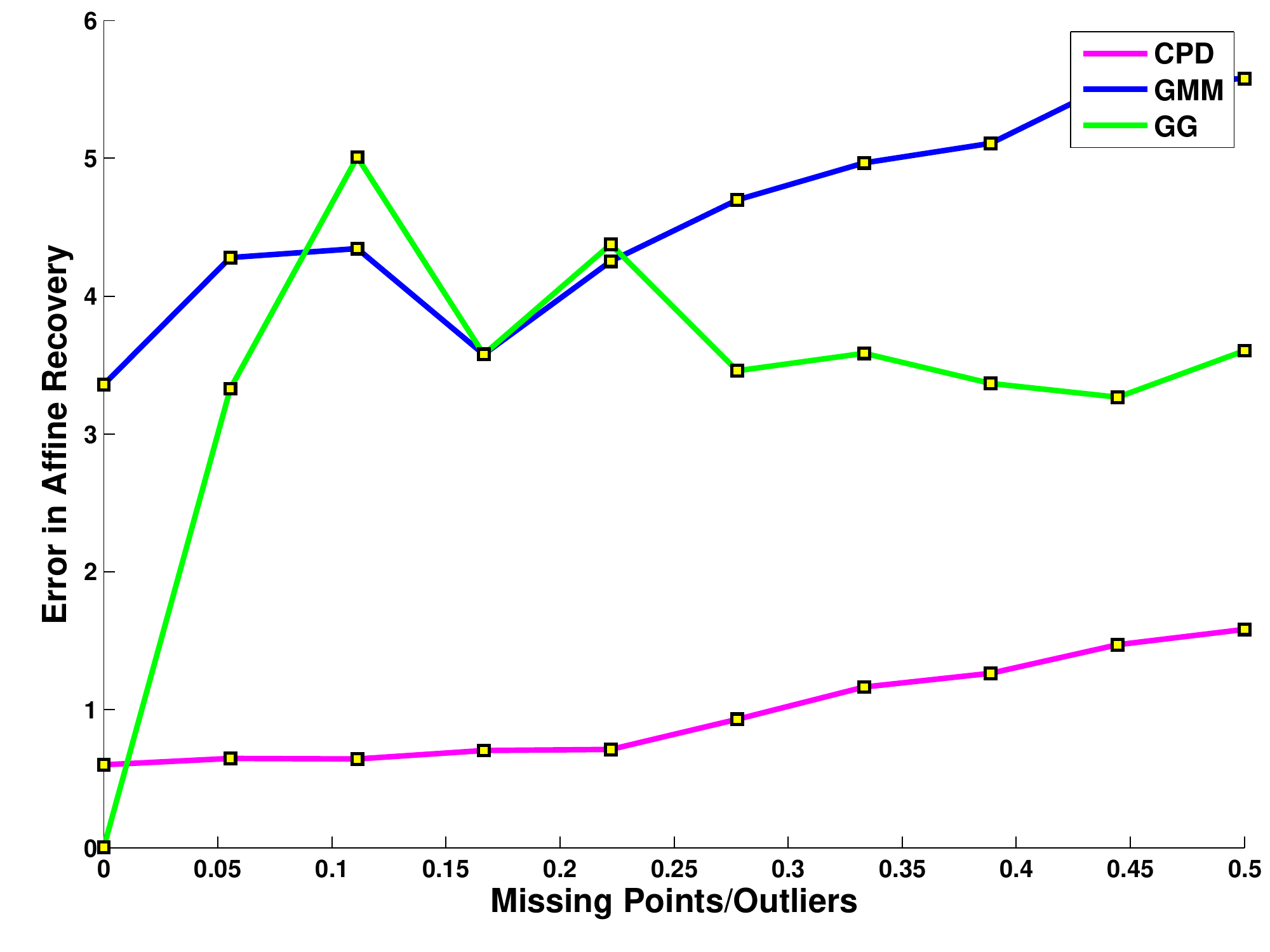}\vspace{-5pt} & 
{\hspace{-12pt}\includegraphics[width=3.5cm,height=2.8cm]{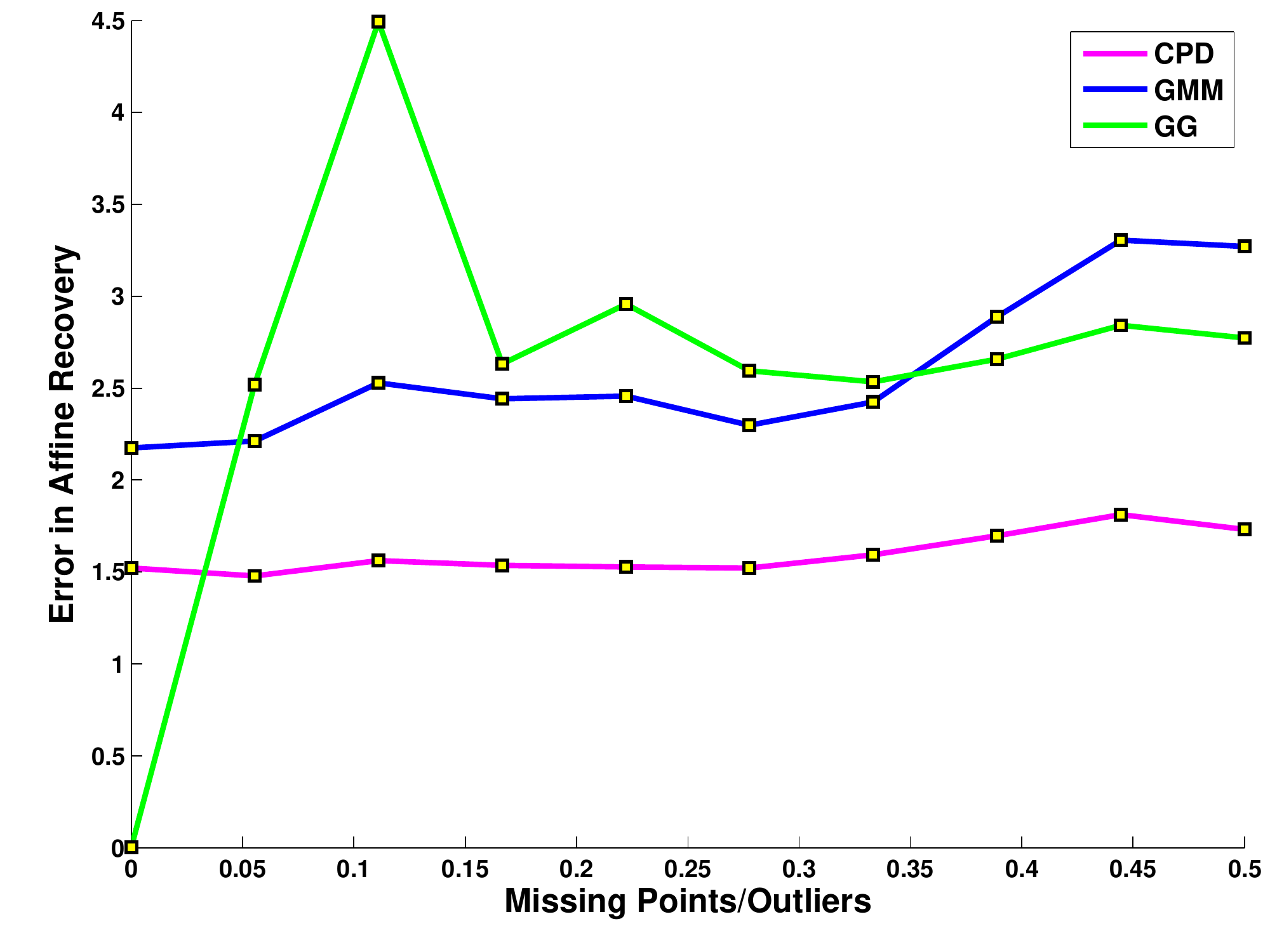}}\tabularnewline
\hspace{-6pt}{\footnotesize{}(a) Big-Ang, Sm-Sc, No-Trans} & 
{\footnotesize{}(b) Med-Ang, Sm-Sc, Sm-Trans} & 
{\footnotesize{}(e) Sm-Ang, Sm-Sc, No-Trans}\tabularnewline
\hspace{-8pt}\includegraphics[width=3.5cm,height=2.8cm]{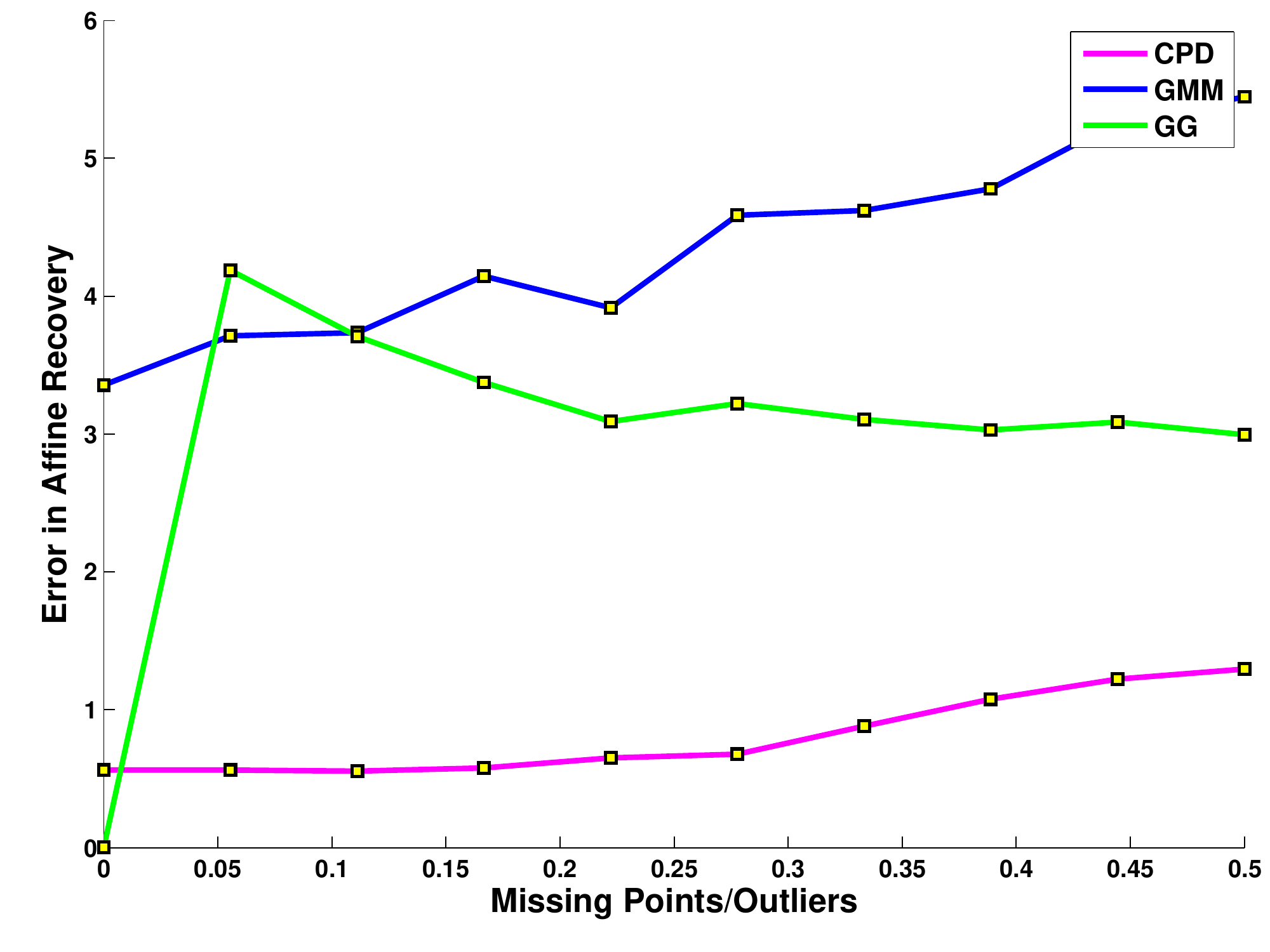} & 
\hspace{-12pt}\includegraphics[width=3.5cm,height=2.8cm]{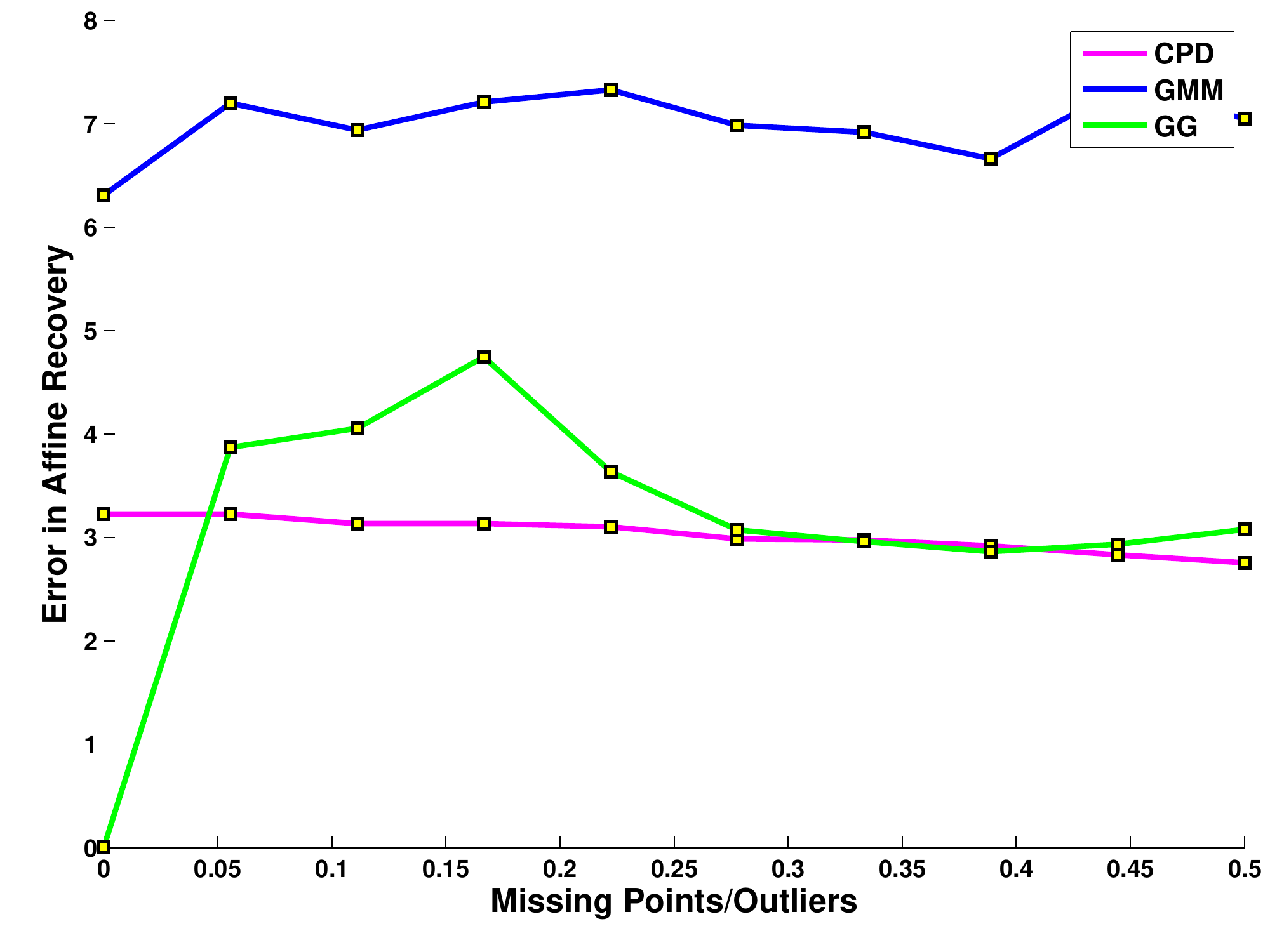}\vspace{-5pt} & 
\hspace{-12pt}\tabularnewline
\hspace{-6pt}{\footnotesize{}(c) Sm-Ang, Sm-Sc, Sm-Trans} & 
\hspace{-12pt}{\footnotesize{}(d) Big-Ang, Sm-Sc, Sm-Trans} & \tabularnewline
\end{tabular}
\par\end{centering}
\protect\caption{MPO 3D. Notice that the GrassGraph approach has a distinct peak at
low levels of MPO and it tapers off as the occlusions and outliers
increase. This lower performance can be attributed to our simple nearest-neighbor correspondence recovery approach. In cases (a)-(d) we are able to outperform GMM
but not CPD because of its coherent motion constraint. \label{fig: MPO 3D}}
\end{figure}

The MPO protocol was followed here with the same affine shapes used
in the noise experiments, yielding 120,000 trials for 2D and 90,000
for 3D with the results shown in Figures \ref{fig: MPO 2D} and \ref{fig: MPO 3D},
respectively. In 2D we outperform the competing methods in (a)-(d)
and are outperformed by CPD and GMM in (e). Note that the case in
(e) has no scale and translation, so it is essentially solving for
the rotation. In cases (a)-(d) CPD performs the worst, AA and GMM
alternate depending on the scale and translation. When the scale increases
between (c) and (d), GMM performs worse than AA. In cases (a) and
(b) as the translation changes from small to medium, there is no noticable
effect in the competing methods. In 2D, our method proves to be
a viable first choice algorithm for correspondence
recovery.

Our performance on 3D MPO is somewhat different---the outlier rejection
schemes built into the competing methods outperform us in some cases. In cases
(a)-(d) we outperform GMM but not CPD. The motion coherence constraint in CPD
is able to withstand the increasing outliers and occlusion. As the amount of
occlusions and outliers increase, our method has a distinct spike at lower
levels and tapers off at higher levels. We attribute this
to our simple correspondence finding scheme.  In the presence of
outliers, the chances of two points being mutual nearest neighbors decreases,
hence our error curves flatten out. Although we do not perform better than all
the competing methods on 3D MPO, our performance in 2D and 3D still
serves as strong evidence that our extremely simple method should be an
algorithm of first choice for correspondence recovery.

\section{Conclusions\label{sec:Conclusions}}

Feature matching is at the heart of many applications in computer
vision. Image registration, object recognition,
and shape analysis all rely heavily on robust methods for recovery
of correspondences and estimation of geometric transformations between
domains. As a core need, myriad pioneering efforts and mathematically
sophisticated formulations have resulted in a multitude of approaches.
However, very few offer the triadic balance of sound theoretical development,
ease of implementation, and robust performance as the proposed GrassGraph
framework. 

GrassGraph develops a true affine invariant through a two-stage process.
First, a Grassmannian representation (GR) is achieved through the
use of standard SVD. Secondly, we approximate the Laplace-Beltrami
operator (LBO) on the GR domain, whose eigenspace coordinates then
free us from an inherent ambiguity present in the coordinates extracted
from the GR. Within this true affine invariant setting, establishing
correspondences reduces to simple nearest-neighbor selection (though
more complex linear assignment solvers can also be used). Correspondences
in this new space are bijectively related to the original pair of
feature points; hence, we are able to directly recover the affine
transformation between them. Our method was evaluated on a broad spectrum
of test cases, parameter settings, noise corruption levels, occlusions,
and included comparisons to other  competing methods. In all, we executed
441,000 experimental trials. Both the number and diversity of test
scenarios far exceed other recent efforts. Under such rigorous validation,
GrassGraph demonstrated state-of-the-art performance, providing credence
to the efficacy of the approach. In the future, we plan to leverage
the lessons learned in the present affine matching situation and investigate
extensions to the non-rigid case. 

%Code is available to run the matching algorithm `out of the box.'

\bibliographystyle{ieee}
\bibliography{ReferencesMerged_V1}

\begin{thebibliography}{10}\itemsep=-1pt

\bibitem{Ball81}
D.~H. Ballard.
\newblock Generalized {H}ough transform to detect arbitrary patterns.
\newblock {\em IEEE Transactions on Pattern Analysis Machine Intelligence},
  13:111--122, 1981.

\bibitem{begelfor2006affine}
E.~Begelfor and M.~Werman.
\newblock Affine invariance revisited.
\newblock In {\em IEEE Conference on Computer Vision and Pattern Recognition},
  2006.

\bibitem{Boothby02}
W.~M. Boothby.
\newblock {\em An Introduction to Differentiable Manifolds and Riemannian
  Geometry}.
\newblock Academic Press, 2002.

\bibitem{Caelli04}
T.~Caelli and S.~Kosinov.
\newblock An eigenspace projection clustering method for inexact graph
  matching.
\newblock {\em IEEE Transactions on Pattern Analysis and Machine Intelligence},
  26:515--519, 2004.

\bibitem{Carcassoni03b}
M.~Carcassoni and E.~Hancock.
\newblock Correspondence matching with modal clusters.
\newblock {\em IEEE Transactions on Pattern Analysis and Machine Intelligence},
  25:1609--1615, 2003.

\bibitem{Carcassoni03}
M.~Carcassoni and E.~R. Hancock.
\newblock Spectral correspondence for point pattern matching.
\newblock {\em Pattern Recognition}, 36:193--204, 2003.

\bibitem{Chui03b}
H.~Chui and A.~Rangarajan.
\newblock A new point matching algorithm for non-rigid registration.
\newblock {\em Computer Vision and Image Understanding}, 89:114--141, 2003.

\bibitem{Dalal07}
P.~Dalal, B.~Munsell, S.~Wang, J.~Tang, K.~Oliver, H.~Ninomiya, X.~Zhou, and
  H.~Fujita.
\newblock A fast 3{D} correspondence method for statistical shape modeling.
\newblock In {\em IEEE Conference on Computer Vision and Pattern Recognition},
  pages 1--8, 2007.

\bibitem{Chung97}
C.~R. Fan.
\newblock {\em Spectral Graph Theory}.
\newblock American Mathematical Society, 1997.

\bibitem{Fischler73}
M.~A. Fischler and R.~A. Elschlager.
\newblock The representation and matching of pictorial structures.
\newblock {\em IEEE Transactions on Computers}, 22:67--92, 1973.

\bibitem{Ha05}
V.~Ha and J.~Moura.
\newblock Affine-permutation invariance of 2{D} shapes.
\newblock {\em IEEE Transactions on Image Processing}, 14:1687--1700, 2005.

\bibitem{Ho07}
J.~Ho, M.~Yang, A.~Rangarajan, and B.~Vemuri.
\newblock A new affine registration algorithm for matching 2{D} point sets.
\newblock In {\em IEEE Workshop on Applications of Computer Vision}, page~25,
  2007.

\bibitem{Isaacs11}
J.~Isaacs and R.~Roberts.
\newblock Metrics of the {Laplace-Beltrami} eigenfunctions for 2{D} shape
  matching.
\newblock In {\em IEEE International Conference on Systems, Man and
  Cybernetics}, pages 3347--3352, 2011.

\bibitem{Jain06}
V.~Jain and H.~Zhang.
\newblock Robust {3D} shape correspondence in the spectral domain.
\newblock In {\em IEEE International Conference on Shape Modeling and
  Applications}, pages 19--31, 2006.

\bibitem{jian2011robust}
B.~Jian and B.~C. Vemuri.
\newblock Robust point set registration using gaussian mixture models.
\newblock {\em IEEE Transactions on Pattern Analysis and Machine Intelligence},
  33:1633--1645, 2011.

\bibitem{Jones08}
P.~W. Jones, M.~Maggioni, and R.~Schul.
\newblock Manifold parametrizations by eigenfunctions of the {L}aplacian and
  heat kernels.
\newblock {\em Proceedings of the National Academy of Sciences},
  105:1803--1808, 2008.

\bibitem{Lamdan90}
Y.~Lamdan, J.~Schwartz, and H.~Wolfson.
\newblock Affine invariant model-based object recognition.
\newblock {\em IEEE Transactions on Robotics and Automation}, 6:578--589, 1990.

\bibitem{SHREC12a}
B.~Li, T.~Schreck, A.~Godil, and et. al.
\newblock {{SHREC12} Track: Sketch-Based 3{D} Shape Retrieval}.
\newblock {\em Eurographics Workshop on 3{D} Object Retrieval}, 2012.

\bibitem{Li14b}
R.~Li, P.~Turaga, A.~Srivastava, and R.~Chellappa.
\newblock Differential geometric representations and algorithms for some
  pattern recognition and computer vision problems.
\newblock {\em Pattern Recognition Letters}, 43:3 -- 16, 2014.

\bibitem{Luo00}
B.~Luo and E.~R. Hancock.
\newblock Alignment and correspondence using singular value decomposition.
\newblock In {\em Structural, Syntactic, and Statistical Pattern Recognition},
  volume 1876, pages 226--235, 2000.

\bibitem{Mateus07}
D.~Mateus, F.~Cuzzolin, R.~Horaud, and E.~Boyer.
\newblock Articulated shape matching using locally linear embedding and
  orthogonal alignment.
\newblock In {\em IEEE International Conference on Computer Vision (ICCV)},
  2007.

\bibitem{Mateus08}
D.~Mateus, R.~Horaud, D.~Knossow, F.~Cuzzolin, and E.~Boyer.
\newblock Articulated shape matching using {L}aplacian eigenfunctions and
  unsupervised point registration.
\newblock In {\em IEEE Conference on Computer Vision and Pattern Recognition},
  2008.

\bibitem{Moyou14}
M.~Moyou, K.~E. Ihou, and A.~M. Peter.
\newblock {LBO Shape Densities}: Efficient 3{D} shape retrieval using wavelet
  densities.
\newblock In {\em IEEE International Conference on Pattern Recognition}, 2014.

\bibitem{Myronenko06}
A.~Myronenko, X.~Song, and .~Carreira-Perpiñán.
\newblock Non-rigid point set registration: Coherent point drift ({CPD}).
\newblock In {\em In Advances in Neural Information Processing Systems 19}. MIT
  Press, 2006.

\bibitem{GatorBait100}
A.~Peter and A.~Rangarajan.
\newblock The {G}ator{B}ait 100 shape database.
\newblock \url{http://www.cise.ufl.edu/~anand/publications.html}, 2007.

\bibitem{Raviv14}
D.~Raviv, A.~M. Bronstein, M.~M. Bronstein, D.~Waisman, N.~A. Sochen, and
  R.~Kimmel.
\newblock Equi-affine invariant geometry for shape analysis.
\newblock {\em Journal of Mathematical Imaging and Vision}, 50:144--163, 2014.

\bibitem{Reuter06}
M.~Reuter, F.-E. Wolter, and N.~Peinecke.
\newblock Laplace-{B}eltrami spectra as {'Shape-DNA'}of surfaces and solids.
\newblock {\em Computer-Aided Design}, 38:342--366, 2006.

\bibitem{rustamov2007laplace}
R.~M. Rustamov.
\newblock Laplace-{B}eltrami eigenfunctions for deformation invariant shape
  representation.
\newblock In {\em Proceedings of the fifth Eurographics symposium on Geometry
  processing}, pages 225--233, 2007.

\bibitem{Scott91}
G.~L. Scott and H.~C. Longuet-Higgins.
\newblock An algorithm for associating the features of two images.
\newblock {\em Proceedings of the Royal Society of London Biological Sciences},
  244:21--26, 1991.

\bibitem{Shapiro92}
L.~S. Shapiro and J.~M. Brady.
\newblock Feature-based correspondence: an eigenvector approach.
\newblock {\em Image and Vision Computing}, 10:283 -- 288, 1992.

\bibitem{Siddiqi99}
K.~Siddiqi, A.~Shokoufandeh, S.~J. Dickinson, and S.~W. Zucker.
\newblock Shock graphs and shape matching.
\newblock {\em International Journal of Computer Vision}, 35:13--32, 1999.

\bibitem{Chellappa08a}
P.~Turaga, A.~Veeraraghavan, and R.~Chellappa.
\newblock Statistical analysis on stiefel and grassmann manifolds with
  applications in computer vision.
\newblock In {\em IEEE Conference on Computer Vision and Pattern Recognition},
  pages 1--8, 2008.

\bibitem{Umey88}
S.~Umeyama.
\newblock An eigendecomposition approach to weighted graph matching problems.
\newblock {\em IEEE Transactions on Pattern Analysis Machine Intelligence},
  10:695--703, 1988.

\bibitem{Wang09b}
Z.~Wang and H.~Xiao.
\newblock Dimension-free affine shape matching through subspace invariance.
\newblock In {\em IEEE Conference on Computer Vision and Pattern Recognition},
  pages 2482--2487, June 2009.

\bibitem{Zhang04}
J.~Zhang and A.~Rangarajan.
\newblock Affine image registration using a new information metric.
\newblock In {\em IEEE Conference on Computer Vision and Pattern Recognition},
  pages 848--855, 2004.

\bibitem{Zhang94}
Z.~Zhang.
\newblock Iterative point matching for registration of free-form curves and
  surfaces.
\newblock {\em International Journal of Computer Vision}, 13:119--152, 1994.

\bibitem{Zhou12}
F.~Zhou and F.~{de la Torre}.
\newblock Factorized graph matching.
\newblock In {\em IEEE Conference on Computer Vision and Pattern Recognition},
  pages 127--134, 2012.

\bibitem{Zhou2013}
F.~Zhou and F.~{de la Torre}.
\newblock Deformable graph matching.
\newblock In {\em {IEEE} Conference on Computer Vision and Pattern
  Recognition}, pages 2922--2929, 2013.

\end{thebibliography}

\end{document}